%% file: iclr2027_conference.tex
\pdfoutput=1  
\documentclass{article}
\usepackage{iclr2027_conference,times}
\usepackage{amsmath,amssymb}
\usepackage{graphicx,booktabs,multirow,tabularx}
\usepackage{algorithm,algpseudocode}
\usepackage{enumitem,alltt,wrapfig,needspace,fontawesome5}
\usepackage{placeins,xcolor}
\usepackage{hyperref,url}
\usepackage{paperlayout}

\input{macros}

\title{\methodfull{}: Hypothesis-Based Neuro-Symbolic Agents through Ontological Abstraction}
\author{Donghyeok Lee$^{1}$, Bong Gyun Kang$^{1}$, Junyong Ahn$^{1}$ \& Sungroh Yoon$^{1,2}$ \\
{\small $^{1}$Interdisciplinary Program in Artificial Intelligence, Seoul National University, Seoul, South Korea} \\
{\small $^{2}$Department of Electrical and Computer Engineering, Seoul National University, Seoul, South Korea} \\
{\small Correspondence to: Sungroh Yoon \texttt{<sryoon@snu.ac.kr>}}}
\iclrfinalcopy

\begin{document}
\maketitle
\lhead{Preprint. Under review.}

\input{figures/fig_abstract_float}

\begin{abstract}
\input{sections/abstract}
\end{abstract}

\input{sections/introduction}
\input{sections/benchmark}
\input{sections/method}
\input{sections/experiments}
\input{sections/related_work}
\input{sections/conclusion}
\FloatBarrier
\input{sections/statements}

\bibliographystyle{iclr2027_conference}
\bibliography{references}

\clearpage
\appendix
\setlength{\abovetopsep}{3pt}
\setcounter{equation}{0}
\renewcommand{\theequation}{A\arabic{equation}}
\renewcommand{\theHequation}{appendix.\arabic{equation}}
\setcounter{figure}{0}
\renewcommand{\thefigure}{A\arabic{figure}}
\renewcommand{\theHfigure}{appendix.\arabic{figure}}
\setcounter{table}{0}
\renewcommand{\thetable}{A\arabic{table}}
\renewcommand{\theHtable}{appendix.\arabic{table}}
\input{sections/appendix_environment}

\input{sections/appendix_method}
\input{sections/appendix_bindings}
\input{sections/appendix_results}

\input{sections/appendix_reproducibility}

\end{document}

%% file: macros.tex
\newcommand{\method}{OntoAct}                        
\newcommand{\methodfull}{\method}                    

\newcommand{\benchmark}{Tacit-FLE}                   

\newcommand{\fle}{FLE}
\newcommand{\flefull}{Factorio Learning Environment}
\newcommand{\mars}{Mars}
\newcommand{\alfworld}{ALFWorld}
\newcommand{\walle}{WALL-E}
\newcommand{\nesyc}{NeSyC}
\newcommand{\ifr}{IfR}
\newcommand{\react}{ReAct}
\newcommand{\reflexion}{Reflexion}

\newcommand{\reactsearch}{{\react{}\textsuperscript{*}}} 
\newcommand{\llmluna}{{Luna}}                  
\newcommand{\llmterra}{{Terra}}              
\newcommand{\llmlunareasoning}{{Luna\textsuperscript{$R$}}}         
\newcommand{\llmlunafull}{GPT-5.6 Luna}
\newcommand{\llmterrafull}{GPT-5.6 Terra}

\newcommand{\chainicon}{\faLink\enspace}
\newcommand{\conjunctionicon}{\faPuzzlePiece\enspace}
\newcommand{\eliminationicon}{\faFilter\enspace}

\newcommand{\schema}{T}
\newcommand{\worldknowledge}{K_W}
\newcommand{\episodestate}{A_t}

\newcommand{\corpus}{D}
\newcommand{\cause}{c^{\ast}}

\newcommand{\sem}[1]{[\![#1]\!]}
\newcommand{\unk}{\mathord{?}}
\newcommand{\op}[1]{\operatorname{#1}}

\newcommand{\epigraphline}[2]{{\par\vspace{-2pt}\begin{flushright}\small\makebox[\linewidth][c]{\itshape ``#1''}\\[1pt]--- #2\end{flushright}\vspace{2pt}}}

\definecolor{draftcolor}{rgb}{0.40,0.20,0.48}
\newcommand{\rev}[1]{\textcolor{black}{#1}}

\newcommand{\promptbox}[2]{%
  \par\medskip\noindent
  \fcolorbox{black!35}{black!3}{%
    \begin{minipage}{\dimexpr\linewidth-2\fboxsep-2\fboxrule\relax}
      \small\textbf{#1}\par\smallskip
      \small\ttfamily\raggedright #2
    \end{minipage}}\par\medskip}

\newcommand{\methodexample}[2][Example]{%
  \par\smallskip
  {\setlength{\fboxsep}{5pt}\setlength{\fboxrule}{0.3pt}%
    \noindent\fcolorbox{black!18}{black!2}{%
      \begin{minipage}{\dimexpr\linewidth-2\fboxsep-2\fboxrule\relax}
        \textbf{#1.} #2
      \end{minipage}}\par}%
  \smallskip}

%% file: figures/fig_abstract_float.tex
\begin{figure}[H]
\centering
\includegraphics[width=\linewidth]{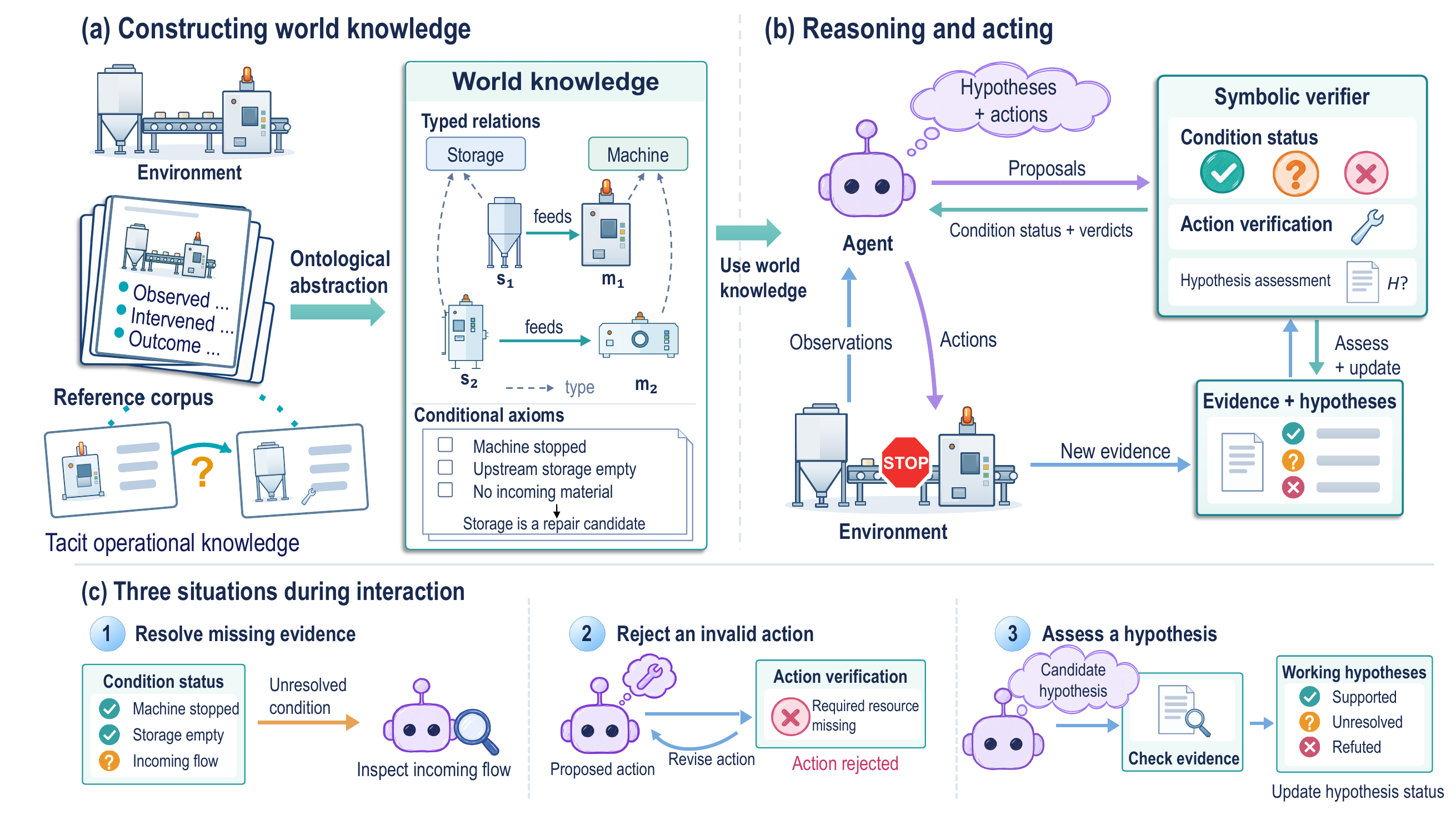}
\caption{{\textbf{The LLM proposes, the ontology verifies.}
(a) \method{} abstracts recorded experience into world knowledge over typed entities, relations, and conditions.
(b) Symbolic checks assess proposed hypotheses and actions against current evidence.
(c) Feedback guides observation, action revision, and hypothesis assessment.}}
\label{fig:abstract}
\end{figure}

%% file: sections/abstract.tex
{
Learning from experience requires recovering reusable knowledge and determining when it applies to a new situation.
We introduce \method{}, a hypothesis-based neuro-symbolic framework built on \emph{ontological abstraction}.
\method{} turns individual experiences into axioms over entity types, relations, and conditions, making their underlying patterns reusable across situations.
A large language model (LLM) uses this world knowledge to propose hypotheses and actions, while a symbolic verifier assesses their supporting conditions against current observations.
Verification identifies missing evidence, and new observations and intervention outcomes update hypothesis assessments and guide subsequent actions.
We also introduce \benchmark{}, a benchmark built on the \flefull{} (\fle{}) for diagnosing and repairing hidden faults in existing factories using inherited operational records.
Its synthetic maintenance corpus distributes evidence across records, and an executed repair may leave the fault unresolved, requiring agents to reassess their diagnoses through interaction.
{Across \benchmark{}, \mars{}, and \alfworld{}, \method{} achieves leading task performance with efficient interaction and inference.}
}

%% file: sections/introduction.tex
\section{Introduction}
\label{sec:introduction}
\epigraphline{We can know more than we can tell.}{Michael Polanyi, \emph{The Tacit Dimension} \citep{polanyi1966tacit}}

{
{Agents often need to act in environments whose operating principles are only partially specified.
Inherited logs, notes, and interaction trajectories record what happened and what was done.
They may leave unstated the dependencies and conditions that made an action useful, or when that experience applies elsewhere.}
We call this \emph{tacit operational knowledge}: knowledge expressed through recorded observations, interventions, and outcomes but left implicit as reusable principles.
}

{
Using this experience requires bridging two gaps.
{First, the agent must recover patterns that generalize beyond the entities named in individual records.
A particular machine--storage connection, for example, instantiates a supplier--recipient relationship; reusable rules express conditions over the entity types and these relational roles.
Second, it must determine whether those relationships and conditions hold in its current situation.
Both gaps matter whether experience is retrieved as text or encoded as executable or symbolic rules \citep{lewis2020rag,zhou2025walle,choi2025nesyc}.}
The agent needs explicit hypotheses that it can test through observation and intervention, retain when supported, and revise when contradicted.
}

{
We introduce \method{}, a neuro-symbolic framework that connects these tasks through \emph{ontological abstraction}.
It retains incident-specific entities and relations while expressing reusable axioms over types, relational roles, and operating conditions, so the same pattern can be instantiated for different entities.
A large language model (LLM) proposes hypotheses and actions from this world knowledge; a symbolic verifier checks their support against current observations (Figure~\ref{fig:abstract}).
Feedback identifies supported hypotheses and conditions requiring further observation.
New observations and intervention outcomes guide hypothesis revision and subsequent actions.
The guiding principle is simple: \emph{the LLM proposes, the ontology verifies}.
}

{
{To evaluate knowledge recovery and use, we also introduce \benchmark{}, a maintenance benchmark built on the \flefull{} \citep{hopkins2025fle}.}
It asks agents to diagnose hidden faults in existing plants from a finite corpus of synthetic maintenance logs and operating documents, together with live observations (Figure~\ref{fig:task}).
Evidence is distributed across records, and an executed repair may leave the fault unresolved.
Success therefore requires selecting useful evidence and translating it into an effective repair.
The benchmark evaluates how agents combine inherited knowledge with feedback from their own observations and interventions.
}
Our contributions are:
\begin{itemize}[leftmargin=*]
    \item {\textbf{\benchmark{}} (Section~\ref{sec:benchmark}), a benchmark for diagnosis and repair from inherited operational records, designed to test the combination of distributed evidence with observations and intervention outcomes;}
    \item {\textbf{\method{}} (Section~\ref{sec:method}), a framework that turns individual experiences into reusable ontological world knowledge and uses it to form, test, and revise hypotheses that guide interaction;}
    \item {\textbf{Performance and efficiency} (Section~\ref{sec:experiments}), demonstrating leading task performance and efficient interaction across \benchmark{}, \mars{}, and \alfworld{}.
    We further support these results with a range of analyses across experimental settings.}
\end{itemize}

{We introduce \benchmark{} in Section~\ref{sec:benchmark} and present the \method{} framework in Section~\ref{sec:method}.
Section~\ref{sec:experiments} evaluates the framework on \benchmark{} and existing benchmarks, followed by related work in Section~\ref{sec:related_work} and conclusions in Section~\ref{sec:conclusion}.}

%% file: sections/benchmark.tex
\section[\benchmark{}: A Benchmark for Tacit Knowledge]{{\benchmark{}: A Benchmark for Tacit Knowledge}}
\label{sec:benchmark}

{\benchmark{} evaluates how agents recover dependencies and repair conditions from past incidents and apply them to new faults.
Built on the factory simulation of \fle{} \citep{hopkins2025fle}, it pairs existing plants with inherited maintenance records and supports diagnosis through observation and intervention {(Appendix~\ref{app:environment}, Figure~\ref{fig:task})}.}

\Needspace{6\baselineskip}
\subsection[Task setting]{{Task setting}}
\label{sec:setting}

{\textbf{Existing plant.}
{An \emph{episode} starts with one hidden fault in an existing plant.
The agent receives unit kinds, positions, and operating readings.}
Inspections reveal further details, while production evolves as actions advance simulation time (Appendix~\ref{app:task_definition}).}

{\textbf{Reference corpus.}
{Each plant comes with a fixed, finite corpus $\corpus$ of synthetic \emph{maintenance logs} and \emph{operating documents}, provided before evaluation.}
The logs describe local observations, checks, repairs, and outcomes; operating documents state selected expectations, such as planned outputs.
{These sources support inference of component dependencies and repair conditions.}}

{\textbf{Diagnosis and repair.}
{Within a fixed turn budget, the agent inspects units, waits, or attempts replacements or refills.
A replacement restores the unit's original specification, including configuration.}
Success requires replacing an accepted repair target or completing a valid refill of an accepted storage target with the required material.
{Failed attempts consume turns and provide feedback; episodes end upon success or budget exhaustion.}}

\subsection[Faults and diagnostic evidence]{{Faults and diagnostic evidence}}
\label{sec:generation}
\label{sec:types}

{Faults include supply interruptions and off-plan configurations, motivated by component failures and departures from intended operation in process diagnosis \citep{venkatasubramanian2003faults}.
Drawing on model-based diagnosis \citep{dekleer1987gde,dekleer1989sherlock}, we construct three task types that differ in how inherited records support a new diagnosis.
Appendix~\ref{app:types} provides examples of the records and observations in each setting.}

{\chainicon\textbf{Chain.}
Local incidents provide evidence of dependencies between components.
For example, one log pairs a stoppage at $B$ with a failure at $A$; another pairs a stoppage at $C$ with a failure at $B$.
Combining the incidents suggests a dependency path $A\!\rightarrow\!B\!\rightarrow\!C$.
When $C$ stalls in a new episode, this path guides upstream inspections, where current readings help locate a supply interruption.}

{\conjunctionicon\textbf{Conjunction.}
A repair precedent, an operating plan, and a current observation play complementary roles.
For example, a past log records that replacing machine $A$ resolved a mismatch between its configured and planned output.
In a new episode, the plan specifies product $X$ for machine $B$, while inspection shows $B$ configured for $Y$.
Recognizing the shared mismatch condition connects a repair demonstrated on $A$ to a candidate intervention on $B$.}

{\eliminationicon\textbf{Elimination.}
Records describe destroyed units, supply interruptions, and power failures, but omit the off-plan configuration encountered in evaluation.
The agent uses recorded symptoms and checks to rule out explanations inconsistent with current observations.
Further observations or interventions help distinguish the remaining possibilities.}

{\textbf{Construction and evaluation.}
Plants, fault cases, and reference corpora are generated from a common specification (Appendix~\ref{app:generation}), with the evaluation population and scale-specific settings detailed in Appendix~\ref{app:benchmark_population} and construction checks in Appendix~\ref{app:validation}.
Records exclude repairs of evaluation targets; consistent plant-specific identifiers reduce shortcuts from Factorio vocabulary.
\rev{We report repair success, mean turns, and spurious interventions (executed repairs that do not meet the success criterion); Appendix~\ref{app:scoring} gives their definitions and aggregation rules.}}

%% file: sections/method.tex
\input{algorithms/alg_stages}
\begin{methodlayout}
\section[\method{}: Ontological Abstraction for Action]{\method{}: Ontological Abstraction for Action}
\label{sec:method}

{
{\method{} turns inherited experience into reusable world knowledge for assessing proposals during interaction.
Following \emph{the LLM proposes, the ontology verifies}, the LLM proposes axioms, hypotheses, and actions, while a symbolic verifier checks them against supporting evidence using the ontology's types, relations, and conditions.
This division governs both knowledge construction and action (Figure~\ref{fig:overview}).}
}
\input{figures/fig_overview_float}

\subsection{Ontological abstraction}
\label{sec:representation}

{
{Reusing an incident requires identifying the relations and conditions that make it relevant to another situation.
\emph{Ontological abstraction} expresses these requirements over entity types and relational roles while retaining the particular facts from which they were drawn.}

\textbf{World knowledge.}
\rev{An ontology schema $\schema$ defines entity types, relations, observable attributes, and admissible axiom forms.}
The corpus-derived ontology separates ground assertions $A_W$ from conditional axioms $R_W$, which express action requirements and diagnostic patterns:
{\begin{equation}
\worldknowledge=(\schema,A_W,R_W).
\label{eq:method_knowledge}
\end{equation}\par}
\textbf{Episode evidence.}
{Applying these reusable conditions requires evidence from the current situation.
\rev{At step $t$, $f_{\mathrm{ground}}$ converts observations into episode assertions $\episodestate$.}
These assertions augment the corpus-derived facts, yielding the knowledge used to assess hypotheses and actions:
\begin{equation}
K_t=(\schema,A_W\cup\episodestate,R_W).
\label{eq:method_current_knowledge}
\end{equation}}
{This combines reusable knowledge with current evidence; see Appendix~\ref{app:worked_example} for a detailed example.\par}
}

\subsection{Constructing world knowledge}
\label{sec:induction}
{
\setlength{\intextsep}{0pt}
\wrapinductionalgorithm

{To build this world knowledge, \method{} processes the reference corpus $\corpus$ before interaction.
The LLM proposes candidate axioms, and the ontology checks which ones enter $\worldknowledge$.\par}
\textbf{The LLM proposes axioms.}
For a record batch $b\subseteq\corpus$, the LLM proposes ground assertions and conditional axioms in the forms allowed by $\schema$:
{\begin{equation}
\widetilde X_b=f_{\mathrm{abs}}(\schema,b)=\{(x_i,\pi_i)\}_{i=1}^{n_b},
\label{eq:method_extraction}
\end{equation}
\par}
{\rev{Each candidate conditional axiom $x_i$ is paired with source records $\pi_i$; $n_b$ counts all candidates.}\par}

\FinishWrap
\textbf{The ontology constrains admission.}
{\rev{Admission requires schema and source validity ($\beta_i$) and a set $\mathcal C_{x_i}$ of tests for its axiom type:}}%
{\setlength{\abovedisplayskip}{2pt}%
\setlength{\belowdisplayskip}{0pt}%
\begin{subequations}
\label{eq:method_admission_test}
\begin{align}
\beta_i&=\op{Typed}(x_i;\schema)\,
\mathbf1[\varnothing\ne\pi_i\subseteq\corpus],
\label{eq:method_source_validity}\\
\alpha_i&=\op{Admissible}(x_i,\pi_i;\worldknowledge,\corpus)=\beta_i\prod_{c\in\mathcal C_{x_i}}
\mathbf1[c(x_i,\pi_i;\worldknowledge,\corpus)].
\label{eq:method_axiom_admission}
\end{align}
\end{subequations}}
\rev{Here $\op{Typed}$ checks schema conformance and $\mathbf1[P]$ indicates whether $P$ holds.
Every test $c\in\mathcal C_{x_i}$ must pass; Appendix~\ref{app:admission} specifies the tests for each environment.}
{\rev{Admitted axioms ($\alpha_i=1$) enter $A_W$ or $R_W$ with their sources via $f_{\mathrm{store}}$; $f_{\mathrm{corpus}}$ checks the assembled knowledge (Algorithm~\ref{alg:induction}).}\par}
}

\subsection{Reasoning and acting with world knowledge}
\label{sec:inference}

{
\setlength{\intextsep}{0pt}
\wrapinferencealgorithm
\textbf{The LLM proposes claims and actions.}
{Let $A$, $K$, and $H$ denote current assertions, knowledge, and hypotheses; primes mark updates.
\rev{Given goal $g$ and feedback $\eta$ (initially empty each turn), the LLM proposes claims $\Delta A$, hypotheses $\Delta H$, and action $a$:}
\begin{equation}
(\Delta A,\Delta H,a)=f_{\mathrm{propose}}(g,K,H,\rev{\eta}).
\label{eq:method_proposal}
\end{equation}}
{\rev{Structural claims specify relations.
Each hypothesis $h$ specifies required relations and nonempty expected conditions $\Phi(h)$, instantiating an induced pattern or proposing a new explanation.}\par}

\textbf{Checking claims against current evidence.}
{\rev{For episode claims, $f_{\mathrm{admit}}$ checks types, structural evidence, and correspondence with cited patterns, including every required condition.}
Admitted structural claims update knowledge:}%
{\setlength{\abovedisplayskip}{2pt}%
\setlength{\belowdisplayskip}{2pt}%
\begin{subequations}
\label{eq:method_claim_update}
\begin{align}
A'&=A\cup f_{\mathrm{admit}}(\Delta A;K),
\label{eq:method_assertion_update}\\
K'&=(\schema,A_W\cup A',R_W).
\label{eq:method_knowledge_update}
\end{align}
\end{subequations}\par}
\WFclear
{\rev{When $h$ instantiates pattern $r$ with conditions $\Phi_r$, $\Phi(h)$ retains every condition, substituting current entities.
Each atom $\varphi=(u,p,\mathrm{op},z)$ tests entity $u$'s observable attribute $p$ against expected value $z$ with comparison $\mathrm{op}$; current readings yield:}}
{\setlength{\abovedisplayskip}{2pt}%
\begin{equation}
\sem{\varphi}_{K'}=
\begin{cases}
\top,&\text{resolved evidence satisfies }p(u)\mathrel{\mathrm{op}}z,\\
\bot,&\text{resolved evidence contradicts }p(u)\mathrel{\mathrm{op}}z,\\
\unk,&\text{otherwise}.
\end{cases}
\label{eq:method_atom_evaluation}
\end{equation}}
{Missing readings remain unknown ($\unk$), and the same evaluations test action conditions.
An open, admitted hypothesis receives the status:}
\begin{equation}
\op{Status}(h)\gets
\begin{cases}
\text{refuted},&\exists\varphi\in\Phi(h):\sem{\varphi}_{K'}=\bot,\\
\text{confirmed},&\forall\varphi\in\Phi(h):\sem{\varphi}_{K'}=\top,\\
\text{open},&\text{otherwise}.
\end{cases}
\label{eq:method_hypothesis_status}
\end{equation}
\rev{Confirmation records observational support, subject to refutation by later readings or repair outcomes; updated hypotheses form $H'$.}

\textbf{The ontology verifies the action.}
{\rev{The verifier tests $R_W$: violated conditions and prohibitions block actions; missing readings alone do not.
An action requiring explanatory support needs a confirmed hypothesis whose conditions still hold; observations need no confirmed hypothesis.}
It returns:
\begin{equation}
(v,\eta)=f_{\mathrm{verify}}(a;K',H'),
\label{eq:method_proposal_verification}
\end{equation}}
{\rev{A blocked verdict $v=\bot$ requests revision while budget remains; $\eta$ reports failed and unresolved checks.
Algorithm~\ref{alg:inference} summarizes this interaction loop.
Appendix~\ref{app:diagnostic} details hypothesis admission, condition evaluation, repair verification, and refutation by observations and outcomes.}}
\label{sec:bindings}
}
\FloatBarrier
\end{methodlayout}

%% file: algorithms/alg_stages.tex
\newcommand{\inductionalgorithm}{%
\begin{algorithm}[H]
{
\caption{{Constructing world knowledge}}
\label{alg:induction}
\footnotesize
\begin{algorithmic}[1]
\Require \rev{ontology schema $\schema$, corpus $\corpus$}
\State $\worldknowledge\gets(\schema,\varnothing,\varnothing)$
\For{each batch $b\subseteq\corpus$}
    \State $\widetilde X_b\gets f_{\mathrm{abs}}(\schema,b)$ \Comment{LLM; Eq.~\ref{eq:method_extraction}}
    \For{each candidate $(x_i,\pi_i)\in\widetilde X_b$}
        \State $\alpha_i\gets\op{Admissible}(x_i,\pi_i;\worldknowledge,\corpus)$ \Comment{Eq.~\ref{eq:method_axiom_admission}}
        \If{$\alpha_i=1$}
            \State $\worldknowledge\gets f_{\mathrm{store}}(\worldknowledge,x_i,\pi_i)$ \Comment{Store pair}
        \EndIf
    \EndFor
\EndFor
\State $\worldknowledge\gets f_{\mathrm{corpus}}(\worldknowledge,\corpus)$ \Comment{Corpus checks}
\State \Return $\worldknowledge$
\end{algorithmic}
}
\end{algorithm}%
}

\newcommand{\inferencealgorithm}{%
\begin{algorithm}[H]
{
\caption{{Reasoning and acting}}
\label{alg:inference}
\footnotesize
\begin{algorithmic}[1]
\Require \rev{$\worldknowledge$, goal $g$, initial observation $o_1$}
\Require {\rev{$B$ turns; up to $J$ proposals per turn}}
\State $A,H,a_{\mathrm{prev}}\gets\varnothing$
\For{$t=1,\ldots,B$}
    \State $\rev{A\gets f_{\mathrm{ground}}(A,o_t)}$ \Comment{Observations}
    \State $K\gets(\schema,A_W\cup A,R_W)$
    \State $H\gets f_{\mathrm{eval}}(H;K,a_{\mathrm{prev}})$ \Comment{\rev{Reassess hypotheses}}
    \State $\eta\gets\varnothing$
    \For{$j=1,\ldots,J$}
        \State $(\Delta A,\Delta H,a)\gets f_{\mathrm{propose}}(g,K,H,\eta)$ \Comment{Eq.~\ref{eq:method_proposal}}
        \State $A\gets A\cup f_{\mathrm{admit}}(\Delta A;K)$ \Comment{Eq.~\ref{eq:method_assertion_update}}
        \State $K\gets(\schema,A_W\cup A,R_W)$ \Comment{Eq.~\ref{eq:method_knowledge_update}}
        \State $H\gets f_{\mathrm{eval}}(H\cup f_{\mathrm{admit}}(\Delta H;K);K)$ \Comment{Eq.~\ref{eq:method_hypothesis_status}}
        \State $(v,\eta)\gets f_{\mathrm{verify}}(a;K,H)$ \Comment{Eq.~\ref{eq:method_proposal_verification}}
        \If{$v\ne\bot$} \textbf{break} \EndIf
    \EndFor
    \State {$(a_{\mathrm{prev}},o_{t+1})\gets f_{\mathrm{execute}}(a,v)$} \Comment{Act; observe}
    \If{terminated} \textbf{break} \EndIf
\EndFor
\end{algorithmic}
}
\end{algorithm}%
}

\newcommand{\wrapinductionalgorithm}{%
\par
\begin{wraptable}{r}{0.475\linewidth}
\vspace{-18pt}
\begin{minipage}[t]{\linewidth}
\vspace{0pt}
\algrenewcommand{\algorithmiccomment}[1]{\nobreak\hfill\mbox{$\triangleright$ ##1}}
\algrenewcommand{\algorithmicindent}{0.7em}
\inductionalgorithm
\end{minipage}
\end{wraptable}%
}

\newcommand{\wrapinferencealgorithm}{%
\par
\begin{wraptable}{r}{0.5\linewidth}
\begin{minipage}[t]{\linewidth}
\vspace{0pt}
\algrenewcommand{\algorithmiccomment}[1]{\nobreak\hfill\mbox{$\triangleright$ ##1}}
\algrenewcommand{\algorithmicindent}{0.7em}
\inferencealgorithm
\end{minipage}
\end{wraptable}%
}

%% file: figures/fig_overview_float.tex
\begin{figure}[!t]
\centering
\includegraphics[width=\linewidth]{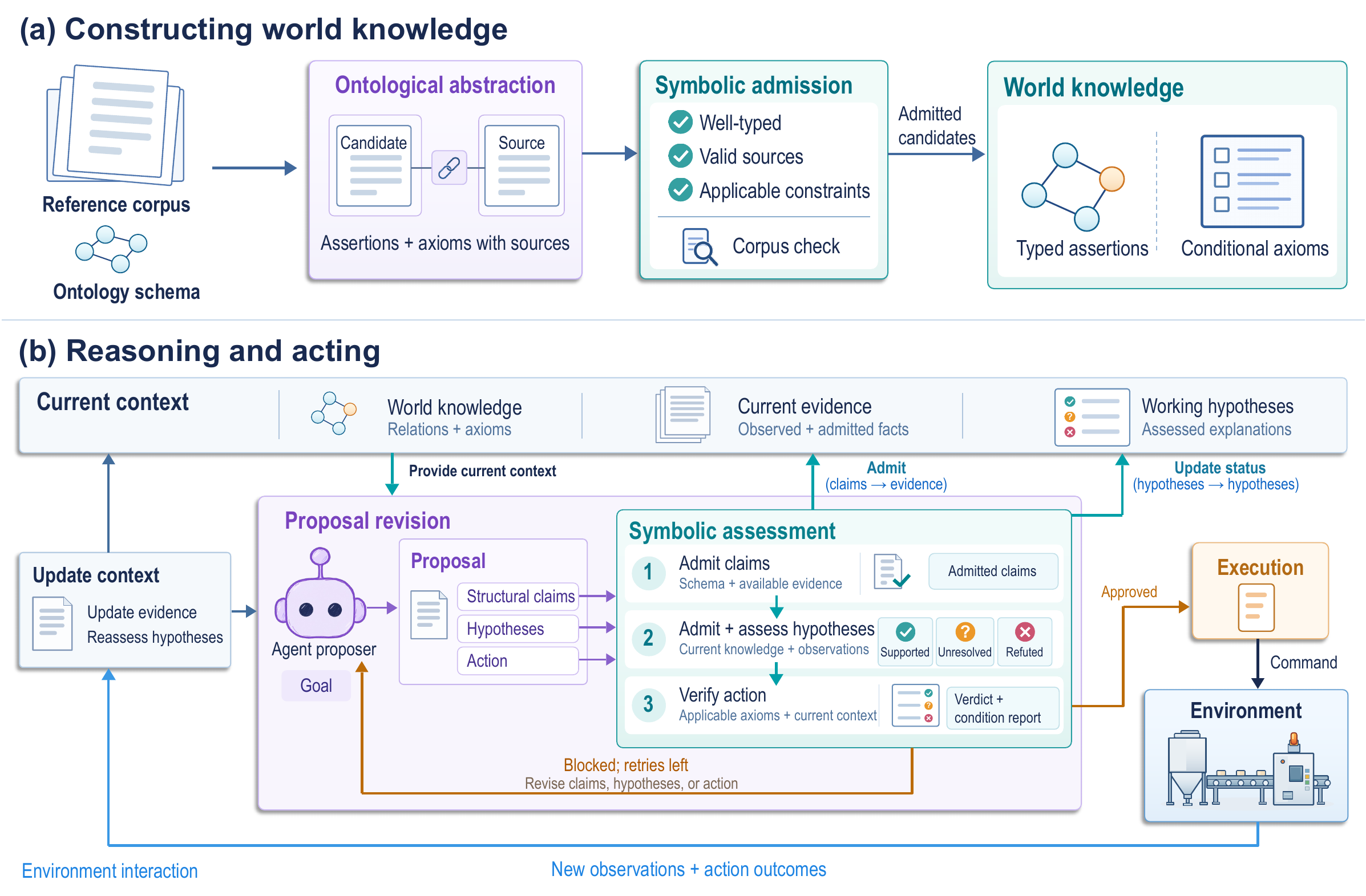}
\caption{{\textbf{The \method{} framework.}
(a) The LLM proposes axioms from a reference corpus; symbolic admission checks their conformance to the {schema} and applicable evidence constraints.
(b) The LLM proposes claims, hypotheses, and actions using the resulting world knowledge; ontology-based verification supplies feedback for revision, while observations and action outcomes update the reasoning context.}}
\label{fig:overview}
\vspace{-6pt}
\end{figure}

%% file: sections/experiments.tex
\begin{experimentlayout}
\section{Experiments}
\label{sec:experiments}
\suppressfloats[t] 

{
\subsection{Experimental Setup}
\label{sec:setup}

\paragraph{Environments and records.}
\textbf{\benchmark{}} tests hidden-fault repair across three task types and plant scales (Section~\ref{sec:benchmark}).
\textbf{\mars{}} \citep{tang2024mars} tests survival and crafting under altered world mechanics; we use the eight-world configuration of \citet{zhou2025walle}.
\textbf{\alfworld{}} \citep{shridhar2021alfworld} tests household instruction following on 134 unseen tasks.
\rev{Corpus-based methods share plant maintenance and operating records, \mars{} exploration records, or \alfworld{} training-game trajectories; Appendix~\ref{app:corpus_protocol} details these corpora and evaluation tasks.}

\paragraph{Baselines.}
We compare \textbf{\method{}} with \textbf{\react{}} \citep{yao2023react}, retrieval-augmented \textbf{\reactsearch{}} \citep{lewis2020rag},
corpus-adapted \textbf{\reflexion{}} \citep{shinn2023reflexion}, and \textbf{\ifr{}} \citep{tang2024mars}, which uses natural-language rules.
\rev{\textbf{\nesyc{}} \citep{choi2025nesyc} and \textbf{\walle{}} \citep{zhou2025walle} check actions against induced answer-set programs and executable rules, respectively.}
Implementations and prompts are detailed in Appendix~\ref{app:baselines}.

\paragraph{Models and protocol.}
\rev{We use \llmlunafull{} \citep{openai2026luna} with reasoning effort \texttt{none}.}
\rev{Medium \benchmark{} and \mars{} also use Luna with \texttt{medium} effort (\llmlunareasoning{}) and \llmterrafull{} (\llmterra{}; \texttt{none}) \citep{openai2026terra}.}
\rev{Methods share tasks and interaction budgets; Appendix~\ref{app:experimental_setup} details settings, budgets, and metrics.}
}

\begingroup
\input{sections/experiments_tacit}

\input{sections/experiments_mars}

\input{sections/experiments_alfworld}
\par\endgroup
\end{experimentlayout}

%% file: sections/experiments_tacit.tex
\begingroup
\input{macros_tacit_names}
\input{tables/tab_main}
{
\subsection{\benchmark{}}
\label{sec:main_results}

{
\paragraph{How well does \method{} repair hidden faults?}
{\rev{Failed repairs consume turns without ending an episode, so eventual success can mask repeated unsuccessful interventions.}
\rev{\method{} achieves the highest overall success at every scale with 70--91\% fewer spurious interventions than \ifr{}, the strongest baseline by overall success (Table~\ref{tab:main}).}
\rev{It also uses the fewest turns among methods with successful repairs.}
Its largest advantage is on chain tasks requiring dependency inference across records; \ifr{} matches or exceeds it on elimination.\par}
}
}
{
\input{tables/tab_tacit_candidate_wrap}
{
\paragraph{Does diagnosis narrow toward the true fault?}
{\rev{\method{} proposes fewer replacement targets than \walle{} and \ifr{}, yet covers the true fault more often, overall and within its first three candidates (Table~\ref{tab:tacit_candidate_summary}).}
\rev{Across scales, 71.4--90.0\% of target-aligned hypotheses remain and 95.6--99.6\% of off-target hypotheses are refuted (Figure~\ref{fig:tacit_verification}; Table~\ref{tab:tacit_hypothesis_diagnostics}).
Appendix~\ref{app:tacit_diagnostics} details this refinement from observations and unsuccessful repairs.}
\par}
}
\FinishWrap
}

{
\paragraph{Does \method{} benefit from reasoning and stronger backbones?}
\label{sec:model_analysis}
{\rev{\method{} benefits from both: reasoning and \llmterra{} raise success from 83.3\% to 97.2\% and 100.0\%, while SI falls from 2.39 to 0.75 and 0.44, respectively (69\% and 81\% reductions; Figure~\ref{fig:tacit_model_conditions}).
\react{} and \reactsearch{} achieve higher success with \llmterra{} than with added reasoning, whereas \reflexion{} and \ifr{} benefit more from reasoning.
Yet higher success can mask repeated failed repairs: with reasoning or \llmterra{}, these baselines still incur 7.9--13.7 spurious interventions per episode.
\method{} also reduces mean turns from 19.4 to 8.4 in both conditions (Table~\ref{tab:tacit_models}).}
\par}
}
\input{figures/fig_tacit_analysis}
\endgroup

%% file: macros_tacit_names.tex
\renewcommand{\reactsearch}{\react{}\textsuperscript{*}}
\renewcommand{\llmluna}{{Luna}}
\renewcommand{\llmlunareasoning}{{Luna\textsuperscript{$R$}}}
\renewcommand{\llmterra}{{Terra}}

%% file: tables/tab_main.tex
\begin{table}[!b]
\vspace{-24pt}
\centering
\begingroup
\input{macros_tacit_names}
\caption{\textbf{Repair performance on \benchmark{}.}
{Success (\%) by type and overall (All); SI: mean spurious interventions; Turns: mean turns.
Failed episodes use the full budget (Appendix~\ref{app:scoring}).
Bold: best; underlined: second-best overall success, interventions, and turns.
Intervention and turn rankings require at least one repair.}}
\label{tab:main}
\small
\setlength{\tabcolsep}{2pt}
\medmuskip=1mu
\begin{tabular*}{\linewidth}{@{\extracolsep{\fill}}cc@{\hspace{4pt}}|@{\hspace{4pt}}cccc@{\hspace{4pt}}|@{\hspace{4pt}}cc@{}}
\toprule
\multicolumn{2}{c}{} & \multicolumn{4}{c}{{Repair success (\%) $\uparrow$}} & \multicolumn{2}{c}{} \\
\cmidrule(lr){3-6}
Scale & Method & {\chainicon Chain} & {\conjunctionicon Conjunction} & {\eliminationicon Elimination} & All & {SI $\downarrow$} & {Turns $\downarrow$} \\
\midrule
\multirow{7}{*}{Small} & \react{} & 0.0 & 71.4 & \textbf{75.0} & 39.3 & \underline{4.6} & 22.2 \\
 & \reactsearch{} & 15.4 & 85.7 & 37.5 & 39.3 & 7.2 & 22.9 \\
 & \reflexion{} & 7.7 & 71.4 & \textbf{75.0} & 42.9 & 5.3 & 23.2 \\
 & \ifr{} & 38.5 & \textbf{100.0} & \textbf{75.0} & \underline{64.3} & 6.4 & \underline{17.4} \\
 & \nesyc{} & 0.0 & 0.0 & 0.0 & 0.0 & 0.0 & 30.0 \\
 & \walle{} & 7.7 & 14.3 & 12.5 & 10.7 & 7.1 & 29.0 \\
 & \method{} (ours) & \textbf{69.2} & \textbf{100.0} & \textbf{75.0} & \textbf{78.6} & \textbf{1.9} & \textbf{11.8} \\
\midrule
\multirow{7}{*}{Medium} & \react{} & 28.6 & 64.3 & 75.0 & 52.8 & 11.3 & 41.8 \\
 & \reactsearch{} & 21.4 & 64.3 & 87.5 & 52.8 & 13.4 & 38.7 \\
 & \reflexion{} & 21.4 & 42.9 & 37.5 & 33.3 & 15.4 & 46.4 \\
 & \ifr{} & 28.6 & 85.7 & \textbf{100.0} & \underline{66.7} & 11.9 & \underline{31.7} \\
 & \nesyc{} & 0.0 & 0.0 & 0.0 & 0.0 & 0.0 & 60.0 \\
 & \walle{} & 14.3 & 28.6 & 87.5 & 36.1 & \underline{10.1} & 48.0 \\
 & \method{} (ours) & \textbf{57.1} & \textbf{100.0} & \textbf{100.0} & \textbf{83.3} & \textbf{2.4} & \textbf{19.4} \\
\midrule
\multirow{7}{*}{Large} & \react{} & 5.9 & 72.7 & 75.0 & 45.0 & 16.1 & 49.9 \\
 & \reactsearch{} & 23.5 & 36.4 & 66.7 & 40.0 & 19.6 & 56.4 \\
 & \reflexion{} & 11.8 & 9.1 & 50.0 & 22.5 & 23.2 & 63.1 \\
 & \ifr{} & 23.5 & 90.9 & \textbf{100.0} & \underline{65.0} & 17.8 & \underline{42.6} \\
 & \nesyc{} & 0.0 & 0.0 & 0.0 & 0.0 & 0.0 & 75.0 \\
 & \walle{} & 5.9 & 45.5 & 75.0 & 37.5 & \underline{14.9} & 56.6 \\
 & \method{} (ours) & \textbf{82.4} & \textbf{100.0} & 91.7 & \textbf{90.0} & \textbf{1.6} & \textbf{17.2} \\
\bottomrule
\end{tabular*}
\par
\endgroup
\end{table}

%% file: tables/tab_tacit_candidate_wrap.tex
\begin{wraptable}{r}{0.48\linewidth}
\vspace{\dimexpr-\intextsep-\abovecaptionskip+\baselineskip\relax}
\centering
\small
\caption{{\textbf{\rev{Candidate count and fault coverage.}}
\rev{Medium \benchmark{} (36 tasks); targets include unexecuted proposals.} \rev{Coverage counts tasks with the true fault among all candidates or the first three.}}}
\label{tab:tacit_candidate_summary}
\input{tables/tab_tacit_candidate_summary}
\end{wraptable}

%% file: tables/tab_tacit_candidate_summary.tex
\begingroup
\fontsize{8}{9.5}\selectfont
\setlength{\tabcolsep}{1.5pt}
\begin{tabular*}{\linewidth}{@{\extracolsep{\fill}}cccc@{}}
\toprule
 & \method{} & \walle{} & \ifr{} \\
\midrule
Distinct candidates/episode & 3.36 & 5.14 & 9.00 \\
Fault in candidate set & 32/36 & 17/36 & 24/36 \\
Fault in first 3 & 25/36 & 10/36 & 4/36 \\
\bottomrule
\end{tabular*}
\par
\endgroup

%% file: figures/fig_tacit_analysis.tex
\begin{figure}[!tbp]
\centering
\begin{minipage}[t]{0.48\linewidth}
\vspace{0pt}
\input{figures/fig_tacit_verification}
\end{minipage}\hfill
\begin{minipage}[t]{0.48\linewidth}
\vspace{0pt}
\input{figures/fig_tacit_model_conditions_float}
\end{minipage}\par
\vspace{-6pt}
\end{figure}

%% file: figures/fig_tacit_verification.tex
\centering
{
\includegraphics[width=\linewidth]{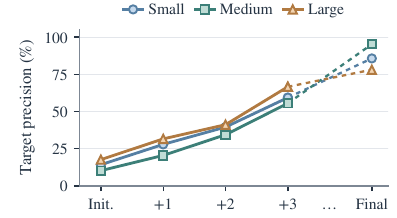}\par
\vspace{12.24pt} 
\small
\caption{{\textbf{\rev{Hypothesis precision over turns.}} Target precision for all tracked hypotheses before selection (Init.) and those retained +1/+2/+3 turns after admission or at the final episode turn.}}
\label{fig:tacit_verification}
\par
}

%% file: figures/fig_tacit_model_conditions_float.tex
\centering
{
\includegraphics[width=\linewidth]{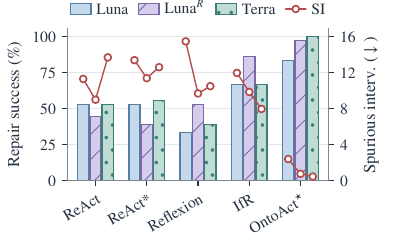}
\small
\caption{{\textbf{\rev{Repair success and SI.}}
\rev{Medium \benchmark{}, 36 tasks. Bars: success (left); markers: mean SI (right, lower is better). $R$: \texttt{medium} reasoning.}
Full metrics are in Table~\ref{tab:tacit_models}.}}
\label{fig:tacit_model_conditions}
\par
}

%% file: sections/experiments_mars.tex
\begingroup
\input{macros_mars_names}
\input{tables/tab_mars}
{
\subsection{\mars{}}
\label{sec:mars_results}

\begingroup
\setlength{\intextsep}{0pt}
\begin{wrapfigure}{r}{0.49\linewidth}
\input{figures/fig_mars_efficiency_float}
\end{wrapfigure}
{
\paragraph{How well does \method{} adapt to altered mechanics?}
{\method{} supports action under changed prerequisites, extending the framework's utility beyond hidden-fault diagnosis.
\rev{It achieves the highest mean Reward and Score across the seven modified worlds (Table~\ref{tab:mars}), although combined changes remain difficult.}
\rev{Inspecting \nesyc{}'s low-scoring runs reveals overly restrictive learned conditions that block planning and trigger repeated waiting; Appendix~\ref{app:mars_models} details these failures.}
\par}
}
{
{\rev{Compared with \reactsearch{}, the strongest baseline by mean Score, \method{} achieves about 50\% higher Score with 45\% fewer evaluation tokens per environment step (0.84k versus 1.54k; Figure~\ref{fig:mars_efficiency}).
Compact learned conditions shorten input contexts at similar LLM call rates.
Appendix~\ref{app:usage} analyzes context usage and reports construction costs separately.}
\par}
}
\WFclear 
\endgroup
}

\newpage 
\input{figures/fig_mars_analysis}

{
\input{tables/tab_mars_knowledge_use}
{
\paragraph{Do learned conditions guide actions before execution?}
{\rev{\method{} checks candidate actions against learned rules before execution.}
For resource conditions learned correctly by both methods, \method{} redirects all 70 violating initial candidates before execution, whereas \ifr{} executes 20 violating commands and replans after their failures (Table~\ref{tab:mars_knowledge_use}).
The distinction is when knowledge changes behavior: \method{} checks represented requirements against the current state before spending an environment interaction.
None of its 136 issued commands violates these shared conditions.
\rev{These traces cover matched knowledge on different trajectories; broader resource-rule coverage remains incomplete.}
\rev{Appendix~\ref{app:mars_knowledge_use} details the shared-condition matching, trajectory analysis, and coverage limits.}
\par}
}
\FinishWrap
}
\par
\endgroup

%% file: macros_mars_names.tex
\renewcommand{\reactsearch}{\react{}\textsuperscript{*}}
\renewcommand{\llmluna}{{Luna}}
\renewcommand{\llmlunareasoning}{{Luna\textsuperscript{$R$}}}
\renewcommand{\llmterra}{{Terra}}

%% file: tables/tab_mars.tex
\begin{table}[!t]
\centering
\begingroup
\input{macros_mars_names}
\caption{\textbf{\rev{\mars{} Reward and Score by world.}}
\llmluna{}; five evaluation episodes per world.
{Reward combines achievement rewards and health changes (mean $\pm$ sample SD); Score geometrically aggregates achievement success rates.}
Average excludes Default; its Reward SD is computed across matched-seed world averages.
Bold/underlined: best/second-best. Metric definitions are in Appendix~\ref{app:metric_definitions}.}
\label{tab:mars}
\fontsize{8pt}{9.5pt}\selectfont
\setlength{\tabcolsep}{2pt}
\medmuskip=1mu
\begin{tabular*}{\linewidth}{@{\extracolsep{\fill}}c*{7}{c}@{}}
\toprule
World & \react{} & \reactsearch{} & \reflexion{} & \ifr{} & \nesyc{} & \walle{} & \method{} \\
\midrule
\multicolumn{8}{@{}c}{\textit{Reward $\uparrow$}} \\
Default & \underline{$10.10\pm4.80$} & $9.30\pm1.64$ & $7.90\pm2.95$ & $8.10\pm2.35$ & $0.70\pm0.55$ & $6.30\pm6.06$ & $\boldsymbol{12.10\pm1.22}$ \\
Terrain & $8.70\pm2.30$ & \underline{$12.16\pm4.03$} & $8.68\pm5.66$ & $9.30\pm3.96$ & $-0.90\pm0.00$ & $7.50\pm3.78$ & $\boldsymbol{12.30\pm1.92}$ \\
Survival & $7.10\pm4.00$ & \underline{$9.50\pm3.44$} & $6.30\pm5.07$ & $7.30\pm4.82$ & $0.98\pm0.99$ & $6.30\pm4.97$ & $\boldsymbol{10.46\pm6.39}$ \\
Task dep. & $4.10\pm1.00$ & $2.90\pm1.30$ & $4.10\pm2.92$ & \underline{$7.30\pm1.10$} & $0.90\pm0.45$ & $3.90\pm0.84$ & $\boldsymbol{11.30\pm1.10}$ \\
Terr.\ surv. & $6.10\pm2.35$ & $\boldsymbol{8.10\pm2.35}$ & $5.10\pm2.92$ & $6.10\pm4.30$ & $2.70\pm1.82$ & $3.90\pm3.11$ & \underline{$6.30\pm3.03$} \\
Terr.\ task & $3.50\pm1.67$ & \underline{$6.90\pm2.17$} & $6.48\pm2.56$ & $3.90\pm2.05$ & $1.10\pm1.00$ & $2.90\pm1.64$ & $\boldsymbol{7.70\pm2.88}$ \\
Surv.\ task & $\boldsymbol{6.50\pm6.11}$ & $2.10\pm2.74$ & $3.70\pm2.41$ & $\boldsymbol{6.50\pm4.04}$ & $2.30\pm1.30$ & \underline{$5.70\pm5.18$} & $5.50\pm6.02$ \\
All three & $0.90\pm2.68$ & $\boldsymbol{1.50\pm2.61}$ & $0.50\pm1.14$ & $-0.10\pm1.30$ & $-0.90\pm0.00$ & $0.70\pm1.52$ & \underline{$1.10\pm2.12$} \\
\cmidrule(lr){1-8}
Average & $5.27\pm1.10$ & \underline{$6.17\pm1.31$} & $4.98\pm1.40$ & $5.76\pm2.14$ & $0.88\pm0.37$ & $4.41\pm1.91$ & $\boldsymbol{7.81\pm1.75}$ \\
\midrule
\multicolumn{8}{@{}c}{\textit{Score (\%) $\uparrow$}} \\
Default & $\boldsymbol{31.85}$ & $13.54$ & $13.67$ & $11.74$ & $0.69$ & $15.26$ & \underline{$29.70$} \\
Terrain & $12.71$ & \underline{$32.88$} & $19.29$ & $22.61$ & $0.00$ & $13.32$ & $\boldsymbol{34.34}$ \\
Survival & $11.10$ & \underline{$18.29$} & $11.58$ & $14.79$ & $1.15$ & $11.94$ & $\boldsymbol{25.83}$ \\
Task dep. & $3.26$ & $2.01$ & $4.46$ & \underline{$8.44$} & $0.51$ & $2.69$ & $\boldsymbol{20.73}$ \\
Terr.\ surv. & $6.49$ & $\boldsymbol{10.52}$ & $5.79$ & \underline{$10.12$} & $2.30$ & $5.10$ & $7.63$ \\
Terr.\ task & $2.65$ & \underline{$6.93$} & $6.31$ & $3.21$ & $0.76$ & $2.36$ & $\boldsymbol{10.17}$ \\
Surv.\ task & $\boldsymbol{17.65}$ & $2.18$ & $3.19$ & $7.09$ & $1.49$ & $8.73$ & \underline{$11.50$} \\
All three & $\boldsymbol{1.80}$ & \underline{$1.69$} & $0.67$ & $0.74$ & $0.00$ & $1.19$ & $1.27$ \\
\cmidrule(lr){1-8}
Average & $7.95$ & \underline{$10.64$} & $7.33$ & $9.57$ & $0.89$ & $6.47$ & $\boldsymbol{15.92}$ \\
\bottomrule
\end{tabular*}
\par
\endgroup
\end{table}

%% file: figures/fig_mars_efficiency_float.tex
\begin{experimentpanel}{figure}
  \centering
  \includegraphics[width=\linewidth,trim=0 6bp 0 0,clip]{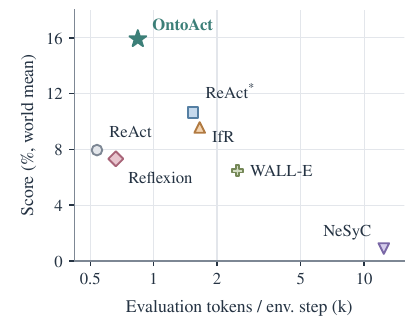}\par
  \vspace{-4pt}
  \caption{{\textbf{\rev{\mars{} Score and evaluation tokens.}}
  \rev{\llmluna{}, seven modified worlds.}
  Equal world weights; input/output tokens exclude knowledge construction.}}
  \label{fig:mars_efficiency}
\end{experimentpanel}

%% file: figures/fig_mars_analysis.tex
\begingroup
\begin{wraptable}{r}{0.48\linewidth}
\vspace{\dimexpr-\intextsep-\abovecaptionskip+\baselineskip\relax}
\input{tables/tab_mars_models}
\end{wraptable}
\paragraph{Does \method{} benefit from more capable models?}
{\method{}'s gains from reasoning and stronger backbones persist across environments, even though baseline responses change (Table~\ref{tab:mars_models}).
Unlike in \benchmark{}, all displayed \mars{} baselines benefit from both changes; \ifr{} narrows the gap under reasoning.
\method{} still retains the highest mean Score, supporting the framework's ability to build on model improvements in both diagnosis and action under altered mechanics.
Full results appear in Appendix~\ref{app:mars_models}.
\par}
\FinishWrap
\endgroup

%% file: tables/tab_mars_models.tex
\begin{experimentpanel}{table}
\centering
\begingroup
\input{macros_mars_names}
\footnotesize
\caption{{\textbf{\rev{\mars{} Score across model conditions.}}
\rev{Mean Score (\%) over seven modified worlds.}
Bold/underlined: best/second-best per condition.}}
\label{tab:mars_models}
\footnotesize
\setlength{\tabcolsep}{2pt}
\medmuskip=1mu
\begin{tabular*}{\linewidth}{@{\extracolsep{\fill}}cccc@{}}
\toprule
Method & \llmluna{} & \llmlunareasoning{} & \llmterra{} \\
\midrule
\react{} & 7.95 & 13.06 & 13.91 \\
\reactsearch{} & \underline{10.64} & 15.23 & 16.48 \\
\reflexion{} & 7.33 & 12.91 & 12.09 \\
\ifr{} & 9.57 & \underline{20.60} & \underline{18.86} \\
\walle{} & 6.47 & 9.93 & 7.79 \\
\midrule
\method{} (ours) & \textbf{15.92} & \textbf{22.99} & \textbf{21.57} \\
\bottomrule
\end{tabular*}
\par
\endgroup
\end{experimentpanel}

%% file: tables/tab_mars_knowledge_use.tex
\begin{wraptable}{r}{0.5\linewidth}
\vspace{\dimexpr-\intextsep-\abovecaptionskip+\baselineskip\relax}
\centering
\begingroup
\input{macros_mars_names}
{
\footnotesize
\caption{\textbf{\rev{Violations of shared learned conditions.}}
Resource conditions correctly stored by both methods.
\llmluna{}, seven modified worlds, 35 episodes per method.
\rev{Fractions use each method's eligible candidates or commands at the specified stage;
Appendix~\ref{app:mars_knowledge_use} details matching and denominators.}}
\label{tab:mars_knowledge_use}
\setlength{\tabcolsep}{2pt}
\renewcommand{\arraystretch}{1}
\begin{tabularx}{\linewidth}{@{}>{\centering\arraybackslash}Xcc@{}}
\toprule
 & \method{} & \ifr{} \\
\midrule
Initial candidate violations & \shortstack[c]{70/245\\(28.6\%)} & \shortstack[c]{20/107\\(18.7\%)} \\
Redirected before execution & \shortstack[c]{70/70\\(100\%)} & \shortstack[c]{0/20\\(0\%)} \\
Issued command violations & \shortstack[c]{0/136\\(0\%)} & \shortstack[c]{20/107\\(18.7\%)} \\
\bottomrule
\end{tabularx}
\par
}
\endgroup
\end{wraptable}

%% file: sections/experiments_alfworld.tex
\begingroup
\input{macros_alfworld_names}
{
\subsection{\alfworld{}}
\label{sec:alfworld_results}
}

{
\input{tables/tab_alfworld_summary}
{
\paragraph{Can \method{} maintain high success with fewer LLM calls?}
{\rev{\method{} completes 132/134 tasks versus 129/134 for \walle{}, using 52.5\% fewer LLM calls and 32.6\% fewer evaluation tokens (Table~\ref{tab:alfworld_summary}).}
\rev{Most savings arise from \method{}'s symbolic condition checks versus \walle{}'s extra LLM outcome-prediction calls (1.02 versus 1.95 calls per environment step); shorter trajectories also contribute.}
\rev{\nesyc{}'s 26.1\% uses disclosed initial object locations and inventory ($\dagger$); Appendix~\ref{app:alfworld_models} compares observation conditions and LLM call roles.}
Task-family results and separate knowledge-construction costs appear in Appendices~\ref{app:alfworld_types} and~\ref{app:usage}.
\par}
}
\par
\FinishWrap
}
\endgroup

%% file: macros_alfworld_names.tex
\renewcommand{\reactsearch}{\react{}\textsuperscript{*}}
\renewcommand{\llmluna}{{Luna}}

%% file: tables/tab_alfworld_summary.tex
\begin{wraptable}{r}{0.5\linewidth}
\vspace{\dimexpr-\intextsep-\abovecaptionskip\relax}
\centering
\begingroup
\input{macros_alfworld_names}
\footnotesize
\caption{\textbf{\rev{\alfworld{} success and evaluation cost.}}
\rev{\llmluna{}.} SR: success (\%); mean environment Steps, LLM Calls, and Tokens (k) per episode.
$^\dagger$Initial object locations and inventory provided.}
\label{tab:alfworld_summary}
\footnotesize
\setlength{\tabcolsep}{2pt}
\medmuskip=1mu
\begin{tabular*}{\linewidth}{@{\extracolsep{\fill}}ccccc@{}}
\toprule
Method & SR (\%) & Steps & Calls & Tokens (k) \\
\midrule
\nesyc{}$^\dagger$ & 26.1 & 38.6 & 9.3 & 281.32 \\
\walle{} & 96.3 & 14.8 & 28.8 & 47.70 \\
\method{} (ours) & 98.5 & 13.5 & 13.7 & 32.16 \\
\bottomrule
\end{tabular*}
\par
\endgroup
\end{wraptable}

%% file: sections/related_work.tex
\section{Related Work}
\label{sec:related_work}

\paragraph{Neuro-symbolic LLM agents.}
Neuro-symbolic agents keep what an LLM learns about an environment in a checkable form, as code world models and planning domains \citep{tang2024worldcoder,khan2026onelife,zhu2025language}, dependency graphs \citep{nottingham2023do}, or rule bases \citep{chen2024automanual}, each refined against the agent's own interaction.
\walle{} learns code rules and a state graph from mismatches between the predictions of an LLM world model and observed transitions, and lets the rules veto actions \citep{zhou2025walle}; \nesyc{} learns and continually repairs symbolic knowledge in answer-set form with a hypothesis-testing loop \citep{choi2025nesyc}; \ifr{} induces natural-language rules from experience in \mars{} \citep{tang2024mars}.
A complementary line separates proposing from checking: solvers decide what an LLM formalizes \citep{pan2023logiclm,olausson2023linc,ye2023satlm}, and external critics or runtime guards check the plans and actions it proposes \citep{kambhampati2024modulo,chen2025shieldagent,xiang2025guardagent}.

\paragraph{Acting with ontologies and knowledge graphs.}
Agents also use structured knowledge: ontologies mediate planning \citep{john2024planning}, scene and procedural graphs constrain LLM planners \citep{rana2023sayplan,zhou2025procedural}, knowledge graphs serve as memory \citep{anokhin2025arigraph} and as the reference against which LLM plans are verified \citep{petruzzellis2025hierarchical}, and such graphs are built from text, including maintenance records \citep{vancauter2024maintenance,zhang2024canonicalize}.
\method{} joins these threads by inducing ontological world knowledge from inherited records, and \benchmark{} evaluates the repairs that follow.

%% file: sections/conclusion.tex
\section[Conclusion]{{Conclusion}}
\label{sec:conclusion}

\rev{We introduced \method{}, a hypothesis-based neuro-symbolic framework that connects learning from experience to action through ontological abstraction.
It recovers reusable axioms over entity types, relations, and conditions from operational records.
An LLM uses this knowledge to propose hypotheses and actions, while symbolic verification checks their support against current evidence.
Observations and intervention outcomes continually refine hypothesis assessments and guide subsequent decisions.
We also introduced \benchmark{} to evaluate how agents combine inherited operational knowledge with observation and intervention to diagnose and repair hidden faults.}

\rev{\method{} achieves the highest overall repair success at every \benchmark{} scale while substantially reducing spurious interventions.
Reasoning and stronger backbones further increase success and reduce unsuccessful repairs.
Results on \mars{} and \alfworld{} extend these benefits to altered world mechanics and household tasks, demonstrating strong task performance alongside efficient interaction and inference.
Our evaluation uses predefined ontology schemas and world mechanics that remain fixed during interaction.
Future work will explore learning and extending these schemas from experience, and revising world knowledge as environments evolve.
Together, these results show how reusable abstractions, explicit hypotheses, and ongoing verification connect accumulated experience to more successful and efficient action.}

%% file: sections/statements.tex
\subsection*{Ethics Statement}
All experiments run in simulators on synthetic data.
The maintenance corpus of \benchmark{} is generated and contains no record of any real plant, person, or organization.
The work targets industrial maintenance, where an agent's interventions can carry physical risk; the verifier of \method{} is a step toward auditability, not a safety guarantee, and we make no claim of readiness for deployment.

\subsection*{Reproducibility Statement}
The benchmark generator, the agent, and every baseline port, with the seeds and prompts used in the experiments, will be released with the camera-ready version.
Appendix~\ref{app:environment} specifies the generator, Appendix~\ref{app:bindings} the bindings and baselines, and Appendix~\ref{app:reproducibility} the settings, budgets, and metric definitions behind each number.

%% file: sections/appendix_environment.tex
{
\section{\benchmark{}: Specification, Examples, and Validation}
\label{app:environment}

This appendix specifies the benchmark introduced in Section~\ref{sec:benchmark}.
\textbf{GitHub repository:} \emph{will be released}.

\smallskip
\noindent\begin{tabularx}{\linewidth}{@{}lX@{}}
\textbf{Section} & \textbf{Contents} \\
\ref{app:task_definition} & Agent inputs, observations, and actions \\
\ref{app:generation} & Plant generation and reference corpus construction \\
\ref{app:types} & Illustrative examples of the three task types \\
\ref{app:benchmark_population} & Evaluation population and scoring \\
\ref{app:validation} & Benchmark construction checks \\
\end{tabularx}
}
\input{figures/fig_task_float}
{

\subsection{Task and information available to the agent}
\label{app:task_definition}
\label{app:fle_relation}

A world $w=(P,\corpus)$ pairs a plant $P$ with a fixed reference corpus $\corpus$.
An episode $e=(w,\cause,B)$ introduces one hidden fault $\cause$ and allows $B$ interaction turns.
The agent receives the corpus, a task brief, a unit roster, and observations of the faulty plant.
The agent interacts with the plant through \benchmark{}'s text commands for inspection and repair.
\benchmark{} runs on the Factorio server infrastructure of \fle{} and adds a plant generator, fault injection, a maintenance corpus, a text command interface, and repair evaluation.
Whereas the troubleshooting suite of \fle{} evaluates spatial error identification in existing factories, \benchmark{} evaluates diagnosis and executed repairs from finite maintenance records.

\subsubsection{Observations and anonymization}
\label{app:anonymization}

The task brief gives the turn budget and the plant's normal production rate in final products per minute; the roster lists unit kinds and positions.
A public status board reports machine run states, storage totals and changes, elapsed game time, and shipping counts.
Power and material-transfer devices require inspection, while opening a unit returns itemized contents.
Conjunction also exposes the configured output for comparison with its operating plan.
Table~\ref{tab:tacit_information} separates available information from evaluator metadata.
Figure~\ref{fig:tacit_interface_example} illustrates the initial input and a subsequent command response.
}
\input{tables/tab_tacit_information}
\input{figures/fig_tacit_interface_example}
{

Items, entities, recipes, and fluids are anonymized with consistent world-specific labels across records and observations.
Unit kinds, positions, structural suffixes, and measured states remain visible.
This reduces reliance on familiar Factorio names while retaining spatial and operational cues.

\subsubsection{Actions and repair feedback}
\label{app:actions}

In Factorio, production evolves with simulation time: machines process inputs, materials move between components, and inventories and shipping counts change.
Table~\ref{tab:tacit_interface} lists the available commands and the game time each advances.
A replacement restores a unit to its original specification, including its configuration; a refill specifies the storage, item, and amount.
Valid actions advance the simulation by the durations in Table~\ref{tab:tacit_interface}, and each response reflects the updated plant state.
Malformed commands consume a turn without advancing the plant.
An accepted repair ends the episode; otherwise interaction continues and supplies further evidence.
}
\input{tables/tab_tacit_interface}
{

\subsection{Constructing plants and inherited records}
\label{app:generation}

A common seeded specification ties the simulated plant to its synthetic records.
Construction has four components:
\begin{enumerate}[leftmargin=*,itemsep=2pt,topsep=3pt]
\item \textbf{Generate the plant.} Select target products according to the scale profile, then trace Factorio's recipe graph backward to identify the required intermediate products and raw inputs.
Compile these dependencies into a spatial layout of machines, input and output storage, material-transfer devices, and power infrastructure.
The world seed controls product selection and layout choices, including shared production stages and spatial spread.
\item \textbf{Establish normal operation.} Build the plant in Factorio and measure positive healthy shipping.
\item \textbf{Introduce a fault.} Inject a supply cut for Chain or an off-plan machine configuration for Conjunction and Elimination.
Cap downstream inventory so it cannot mask the fault indefinitely, and require stalled shipping or a persistent telemetry deviation.
Other product lines may continue operating.
\item \textbf{Construct the reference corpus.} Render historical incidents and operating documents from the same specification, withhold evaluation incidents, and check the intended distribution of evidence.
\end{enumerate}
Appendix~\ref{app:validation} summarizes the checks applied to generated plants and corpora.

}

\paragraph{Example plant topologies.}
Figures~\ref{fig:tacit_plants_small_1}--\ref{fig:tacit_plants_small_2}, \ref{fig:tacit_plants_medium_1}--\ref{fig:tacit_plants_medium_2}, and \ref{fig:tacit_plants_large_1}--\ref{fig:tacit_plants_large_2} show small, medium, and large plants, respectively.
Four selected examples per scale illustrate variation in production depth, branching, and shared intermediate stages.
Arrows indicate material flow; dots join branches, and curved crossings are unconnected.
Small squares denote input and output buffers.
Positions are rearranged for readability, and utilities are omitted.
Anonymized unit identifiers are shown above original names in italics, which are included only as a reading aid.
\input{figures/fig_tacit_layouts_pairs}
\FloatBarrier
{

\subsubsection{Corpus contents and provenance}
\label{app:corpus}

Each maintenance log contains five fields: \texttt{OCCURRED}, \texttt{OBSERVED}, \texttt{CHECKED}, \texttt{ACTION}, and \texttt{RESULT}.
These describe local symptoms, checks, and repairs, leaving relations and conditions to be inferred across records.
The corpus also includes explicit operating information: Chain supplies a buffer-policy note, and Conjunction supplies planned outputs.

Three construction details matter for interpreting these records:
\begin{itemize}[leftmargin=*,itemsep=2pt,topsep=3pt]
\item \textbf{Synthetic provenance.} Historical incidents are rendered from templates over the same plant specification.
\item \textbf{Document counts.} Required incidents and distractors are supplemented to the configured corpus size; a recurring incident may appear under separate log identifiers.
Chain can add dependency records and its operating note afterward.
Table~\ref{tab:profiles} reports the resulting document counts, including these additions and repetitions.
\item \textbf{Held-out faults.} No registered test fault appears as a historical repair target in the shared corpus.
The same unit may still appear as a downstream observation, roster entry, or planned-output entry.
Records are shuffled, and the Conjunction plan lists applicable machines rather than singling out the faulty one.
\end{itemize}

\subsection{Three task types: illustrative examples}
\label{app:types}

Tables~\ref{tab:example_chain}--\ref{tab:example_elimination} pair excerpts from the maintenance logs and operating documents with environment responses to the agent's commands.
Each table pairs original document and response excerpts with an interpretation of their diagnostic relevance.

\begin{itemize}[leftmargin=*,itemsep=2pt,topsep=3pt]
\item \chainicon\textbf{Chain: compose dependencies.}
Local records connect successive units.
A separate buffer-policy note distinguishes normally empty internal buffers from yard-stocked intakes whose lack of incoming material indicates a supply cut.
\item \conjunctionicon\textbf{Conjunction: combine complementary evidence.}
A prior incident supplies a repair pattern, a plan supplies the expected output, and a current reading supplies the actual output.
The mismatch is unit-specific: an item label alone does not indicate a fault.
\item \eliminationicon\textbf{Elimination: test alternative explanations.}
Records describe destroyed units, supply cuts, and power failures, but omit the off-plan configuration encountered in evaluation.
Rejecting documented explanations narrows the search; further observations or interventions may still be needed to distinguish remaining candidates.
\end{itemize}
}
\input{sections/appendix_tacit_examples}
\FloatBarrier
{

\subsection{Evaluation population and scoring}
\label{app:benchmark_population}

Each combination of scale and task type has ten worlds: the first five are reserved for development, and the last five for evaluation.
All registered fault positions in the evaluation worlds are tested, giving 45 world--type corpora and 104 episodes (Table~\ref{tab:profiles}).
For each world, we rebuild the same plant and inject each registered fault in a separate episode, keeping its reference corpus fixed.
Small, medium, and large use one, two, and four target products, respectively; corpus size, realized topology, task mix, and turn budgets also vary.
}
\input{tables/tab_profiles}
{

A method may construct reusable knowledge from the corpus or retrieve records according to its evaluation procedure.
No additional exploratory training episodes are supplied.
Corpus-derived knowledge may be shared across a world's episodes, but test observations, hypotheses, and outcomes are reset each time.
The no-corpus control omits the records; method-specific access is detailed in Appendix~\ref{app:baselines}.

\subsubsection{Repair criterion and metrics}
\label{app:scoring}

The evaluator retains an accepted repair-target set that is hidden from the agent.
Replacing an accepted target completes the episode.
A refill must address an accepted target, be accepted by the environment, and use the planned supply item when specified.
Otherwise the attempt consumes a turn and the episode continues, until a repair succeeds or the budget is exhausted.
\begin{itemize}[leftmargin=*,itemsep=2pt,topsep=3pt]
\item \textbf{Repair success:} the fraction of scheduled evaluation episodes with an accepted repair within the turn budget.
\item \textbf{Turns:} mean interaction turns, including the full budget for unsuccessful episodes.
\item \textbf{Spurious interventions:} mean executed replacements or refills per episode that fail the repair criterion, including an unsuccessful refill of an otherwise relevant unit.
\end{itemize}
LLM calls and tokens are accounted for separately from environment turns (Appendix~\ref{app:usage}).
}
\FloatBarrier
\input{sections/appendix_validation}

%% file: figures/fig_task_float.tex
\begin{figure}[!t]
\centering
\includegraphics[width=\linewidth]{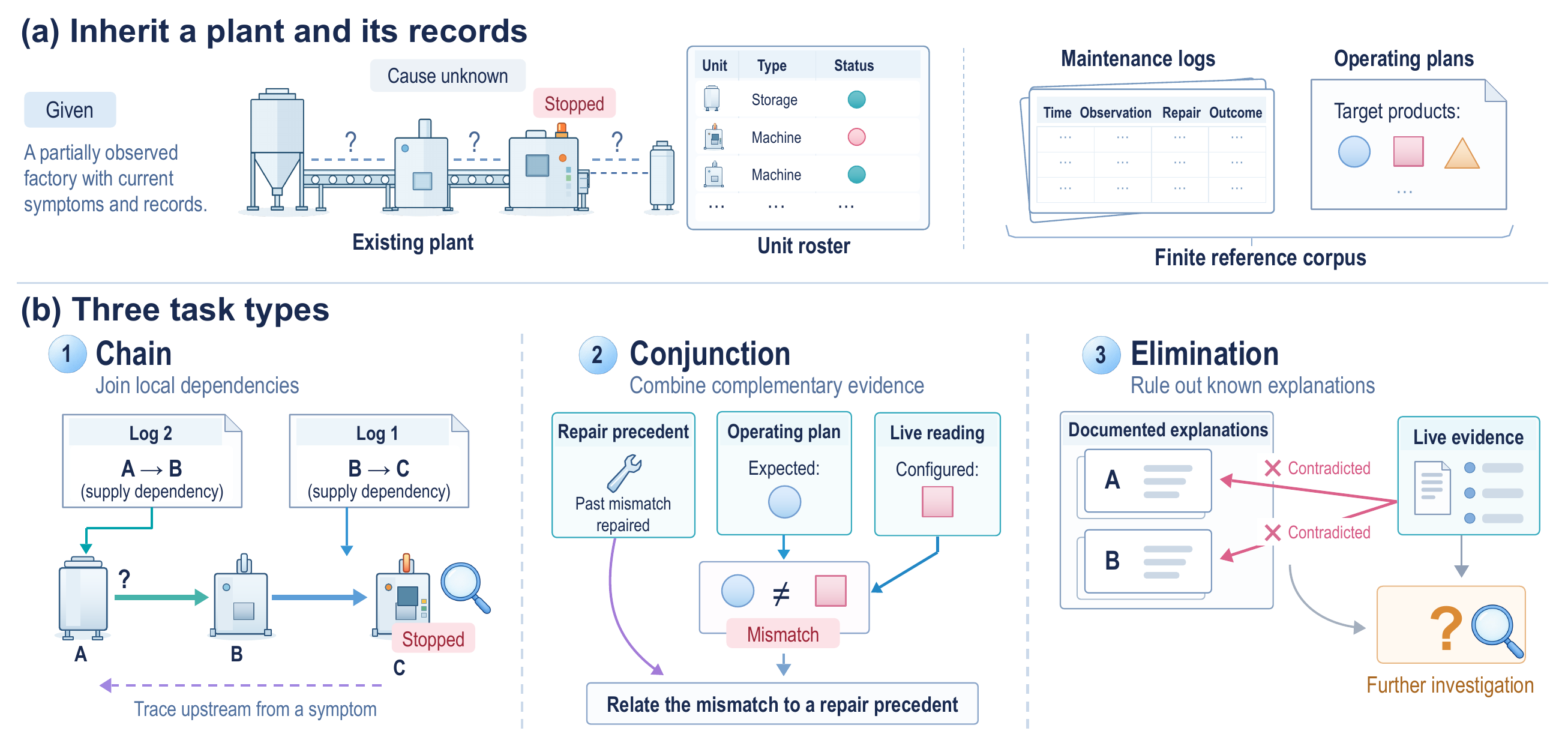}
\caption{{\textbf{\benchmark{}.}
(a) An agent inherits an existing plant, a finite reference corpus of maintenance logs and operating documents, and observations of a new fault.
(b) \chainicon Chain links dependencies across incidents; \conjunctionicon Conjunction combines repair precedents, plans, and observations; \eliminationicon Elimination rules out known explanations contradicted by current evidence.}}
\label{fig:task}
\end{figure}

%% file: tables/tab_tacit_information.tex
\begin{table}[htbp]
\centering
{
\caption{{\textbf{Information boundary.} Public records can disclose selected planned facts; the complete original specification remains evaluator metadata.}}
\label{tab:tacit_information}
{
\begin{tabularx}{\linewidth}{@{}>{\raggedright\arraybackslash}p{0.23\linewidth} >{\raggedright\arraybackslash}X@{}}
\toprule
Source & Available information \\
\midrule
Initial task & Turn budget, normal production rate (final products/minute), unit kinds, and coordinates. \\
Inherited records & Local incident observations, checks, repairs, and outcomes; buffer policy for Chain; planned outputs for Conjunction. \\
Current observations & Machine states, storage totals and changes, shipping; inspections and itemized readings on request. \\
Repair feedback & Episode termination after an accepted repair; continued interaction otherwise. \\
\midrule
Evaluator only & Injected fault, accepted repair targets, full original plant specification, and coverage annotations. \\
\bottomrule
\end{tabularx}\par
}
}
\end{table}

%% file: figures/fig_tacit_interface_example.tex
\begin{figure}[!htbp]
\centering
{
\setlength{\fboxsep}{7pt}
\fcolorbox{black!20}{black!2}{%
\begin{minipage}{\dimexpr\linewidth-2\fboxsep-2\fboxrule\relax}
\small\raggedright
\textbf{Task brief and unit roster}\par\smallskip
{\ttfamily\footnotesize Contract throughput: about 2 final products per minute.\par}
{\ttfamily\footnotesize st-04 @(150.5,192.5)\par}
{\ttfamily\footnotesize st-05 @(164.5,164.5)\par}
\par\smallskip
\textbf{Initial observation}\par\smallskip
{\ttfamily\footnotesize [STATUS] first reading\par}
{\ttfamily\footnotesize machines: mat-078 running, mat-214 running, st-00 stopped, st-01 stopped, st-02 stopped, st-03 stopped, st-04 stopped, st-05
running\par}
{\ttfamily\footnotesize Shipping buffer ship cumulative 8 (mat-136 x8). (elapsed 0s)\par}
\par\smallskip
\textbf{Agent command}\par\smallskip
{\ttfamily\footnotesize open st-05\par}
\par\smallskip
\textbf{Environment response}\par\smallskip
{\ttfamily\footnotesize st-05 opened: input buffer empty. configured output: mat-232.\par}
{\ttfamily\footnotesize [STATUS] change over the last 10s\par}
{\ttfamily\footnotesize Shipping buffer ship cumulative 10 (mat-136 x10). (elapsed 60s)\par}
{\ttfamily\footnotesize [SHIFT] not fixed yet; the shift goes on.\par}
\end{minipage}}
\caption{\textbf{Example interaction in Tacit-FLE.} A medium-scale Conjunction episode:
the task brief, roster, initial status, and a later \texttt{open} command with its response. Only the selected lines are shown; identifiers and wording are preserved.}
\label{fig:tacit_interface_example}
}
\end{figure}

%% file: tables/tab_tacit_interface.tex
\begin{table}[htbp]
\centering
{
\caption{{\textbf{Agent actions in the evaluated protocol.} Each action uses one turn. The plant advances by the indicated game time; LLM deliberation itself is not charged as game time. The status board is included in observations at no additional turn cost.}}
\label{tab:tacit_interface}
{
\begin{tabularx}{\linewidth}{@{}>{\raggedright\arraybackslash}p{0.28\linewidth} c >{\raggedright\arraybackslash}X@{}}
\toprule
Command & Game seconds & Effect or returned information \\
\midrule
\texttt{inspect $<$unit$>$} & 5 & External state; power and material-transfer device readings require this probe. \\
\texttt{open $<$unit$>$} & 10 & Itemized storage or input contents; configured output in Conjunction. \\
\texttt{wait $<$seconds$>$} & specified & Advance the plant and receive an updated status board. \\
\texttt{replace $<$unit$>$} & 15 & Restore the addressed unit to its original specification. \\
\texttt{refill $<$storage$>$ $<$item$>$ $<$count$>$} & 10 & Insert the selected item into storage; success also depends on the repair target and item. \\
\bottomrule
\end{tabularx}\par
}
}
\end{table}

%% file: figures/fig_tacit_layouts_pairs.tex
\begin{figure}[p]
\centering
\includegraphics[width=0.78\linewidth]{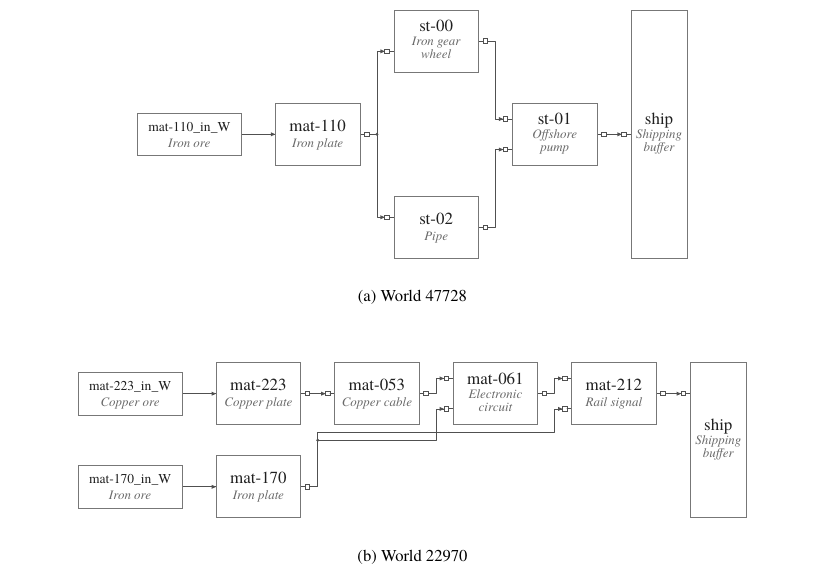}
\caption{\textbf{Small-scale plant examples (a)--(b).} Each plant has one target product.
One input stream splits and reconverges in (a), while (b) combines iron and copper through a circuit stage.}
\label{fig:tacit_plants_small_1}
\end{figure}

\begin{figure}[p]
\centering
\includegraphics[width=0.78\linewidth]{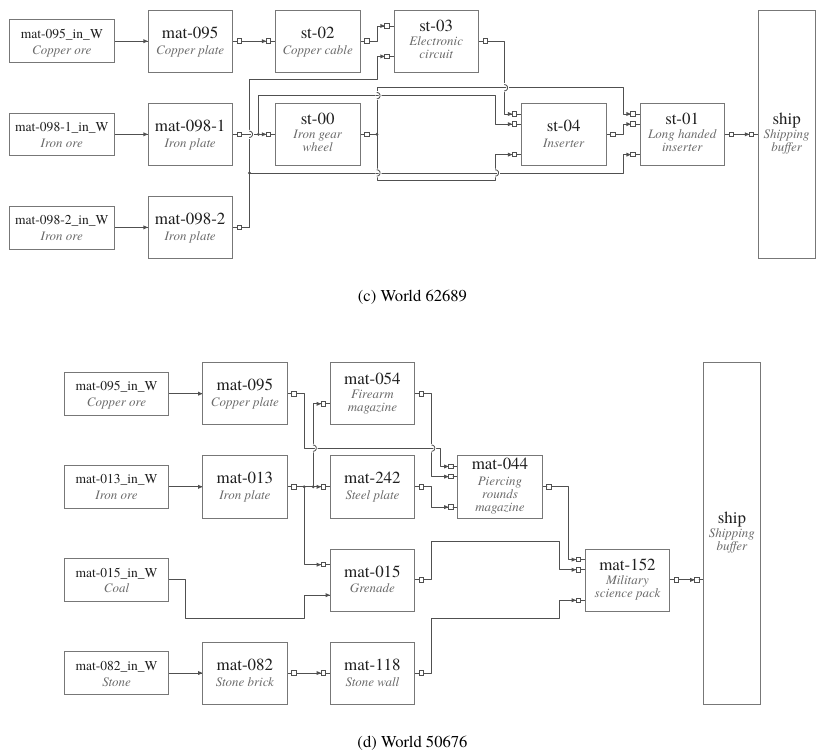}
\caption{\textbf{Small-scale plant examples (c)--(d).} Each plant has one target product.
A deeper assembly sequence in (c) contrasts with several material branches converging on one science pack in (d).}
\label{fig:tacit_plants_small_2}
\end{figure}

\begin{figure}[p]
\centering
\includegraphics[width=0.78\linewidth]{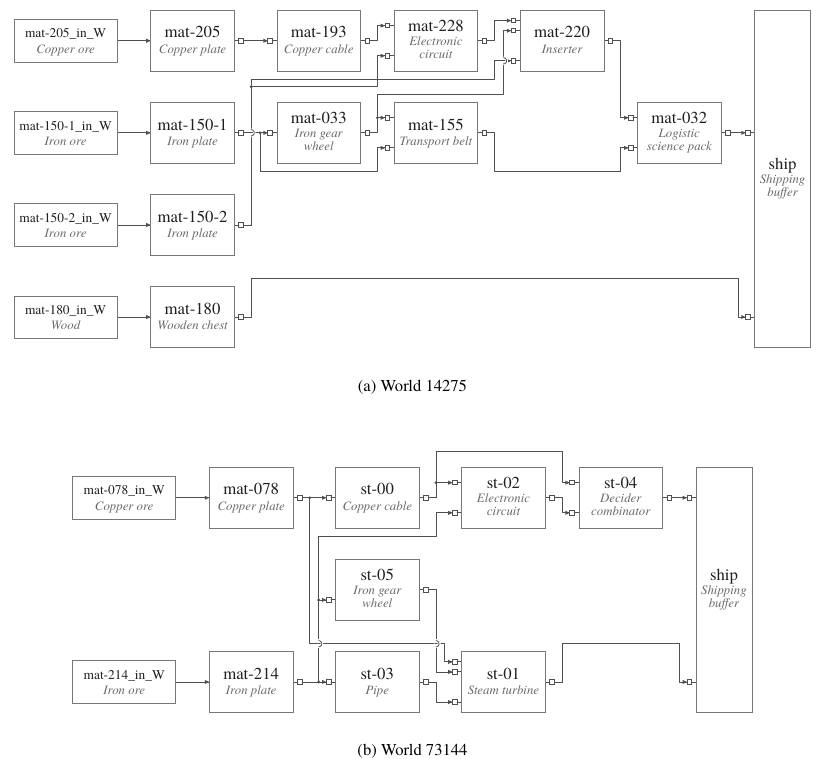}
\caption{\textbf{Medium-scale plant examples (a)--(b).} Each plant has two target products.
An independent short production line accompanies the deeper chain in (a); the two targets reuse upstream products in (b).}
\label{fig:tacit_plants_medium_1}
\end{figure}

\begin{figure}[p]
\centering
\includegraphics[width=0.78\linewidth]{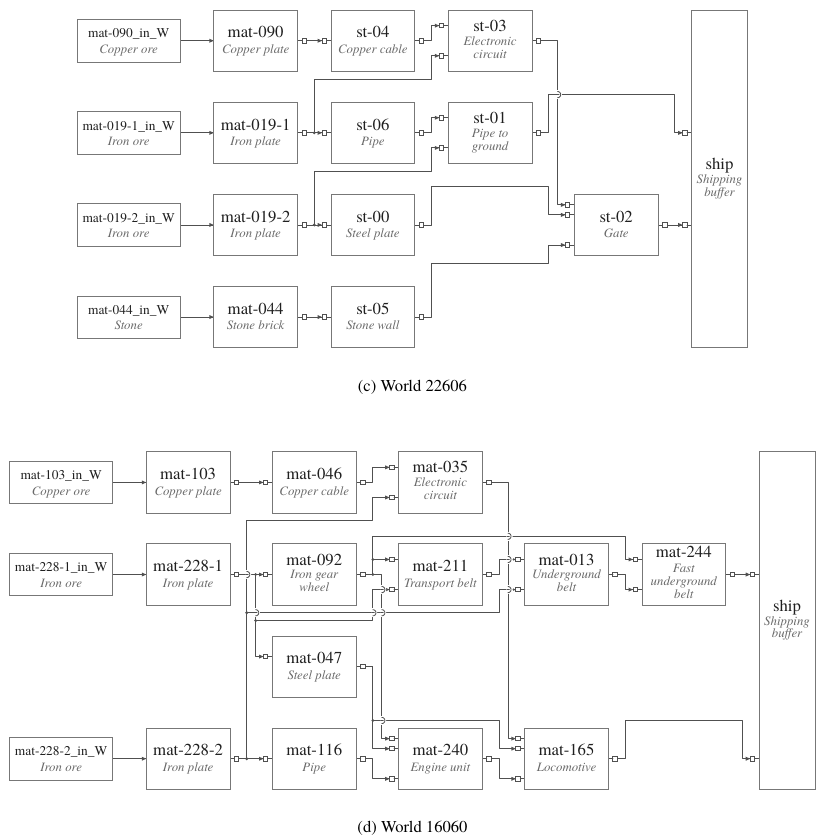}
\caption{\textbf{Medium-scale plant examples (c)--(d).} Each plant has two target products.
Parallel material branches in (c) contrast with the longer transport and engine chains in (d).}
\label{fig:tacit_plants_medium_2}
\end{figure}

\begin{figure}[p]
\centering
\includegraphics[width=0.78\linewidth]{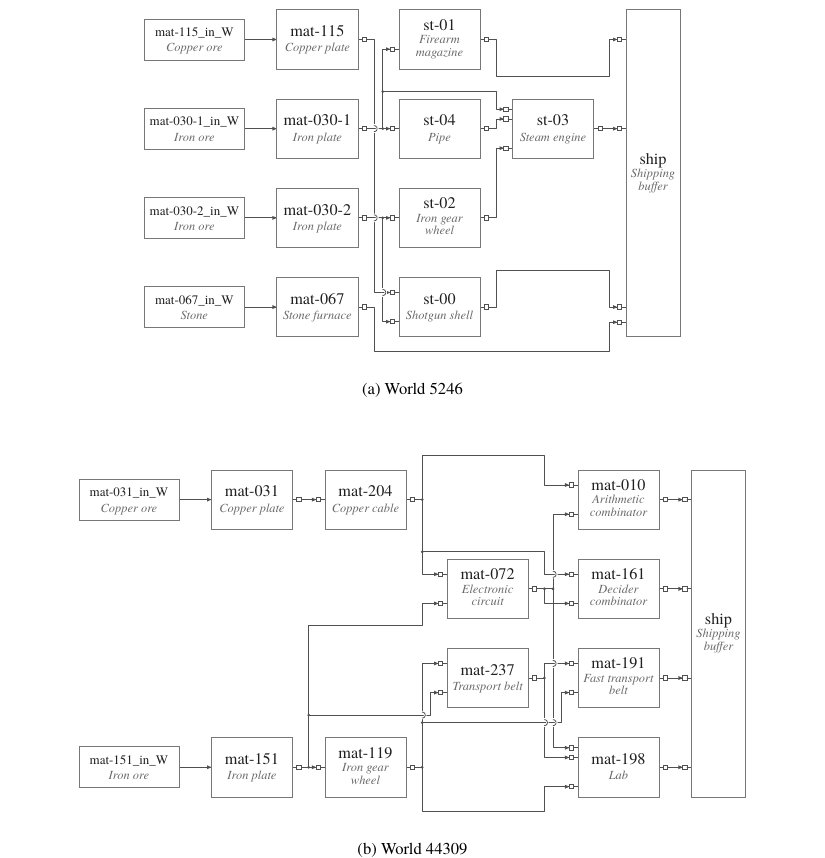}
\caption{\textbf{Large-scale plant examples (a)--(b).} Each plant has four target products.
Shorter parallel paths in (a) contrast with the reuse of a compact upstream circuit and belt network in (b).}
\label{fig:tacit_plants_large_1}
\end{figure}

\begin{figure}[p]
\centering
\includegraphics[width=0.78\linewidth]{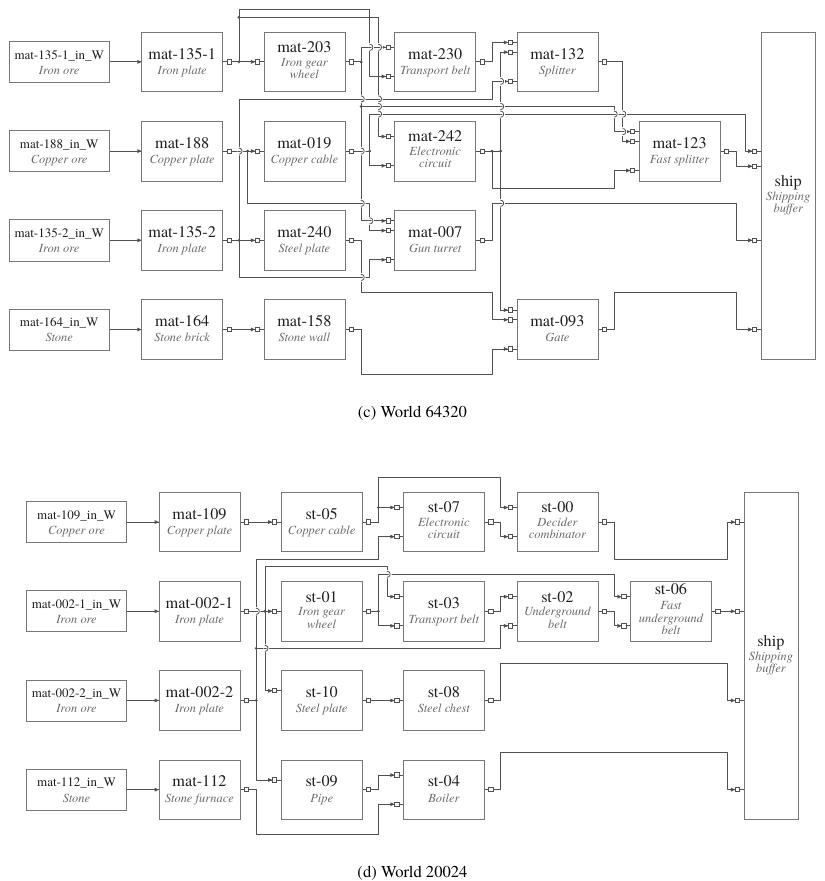}
\caption{\textbf{Large-scale plant examples (c)--(d).} Each plant has four target products.
Direct shipping accompanies further assembly in (c), while (d) contains fifteen processing units across several intermediate stages.}
\label{fig:tacit_plants_large_2}
\end{figure}

%% file: sections/appendix_tacit_examples.tex
\begin{table}[!htbp]
\centering
{
\caption{\textbf{\chainicon Chain: an illustrative example.} Record excerpts and command responses from a medium-scale evaluation case.}
\label{tab:example_chain}
\small
\begin{tabularx}{\linewidth}{@{}>{\raggedright\arraybackslash}p{0.22\linewidth} >{\raggedright\arraybackslash}X@{}}
\toprule
Evidence source & Exact excerpt \\
\midrule
Log log\_6056 & {\ttfamily [OBSERVED] mat-193 stopped, input buffer empty.}\par {\ttfamily [CHECKED] site visit {\normalfont\textemdash{}} mat-205 found destroyed.} \\
\addlinespace
Log log\_7299 & {\ttfamily [OBSERVED] mat-228 stopped, input buffer empty.}\par {\ttfamily [CHECKED] site visit {\normalfont\textemdash{}} mat-193 found destroyed.} \\
\addlinespace
Ops note buffer-policy & {\ttfamily [NORM] Line-internal feed buffers are pulled just-in-time and normally sit empty; an empty internal buffer during a stall is expected, not
evidence.}\par {\ttfamily [NORM] Raw-material intake buffers are kept stocked by the yard at all times. An intake buffer found empty with nothing arriving means its supply feed is
severed {\normalfont\textemdash{}} replace that storage.} \\
\addlinespace
Command responses & {\ttfamily mat-205\_in\_W opened: empty. flat at zero.} \\
\midrule
Interpretation & The two logs connect mat-205 to mat-193 and mat-193 to mat-228. The operating note distinguishes an intake supply cut from normal internal-buffer emptiness. \\
\bottomrule
\end{tabularx}\par
}
\end{table}

\begin{table}[!htbp]
\centering
{
\caption{\textbf{\conjunctionicon Conjunction: an illustrative example.} Record excerpts and command responses from a medium-scale evaluation case.}
\label{tab:example_conjunction}
\small
\begin{tabularx}{\linewidth}{@{}>{\raggedright\arraybackslash}p{0.22\linewidth} >{\raggedright\arraybackslash}X@{}}
\toprule
Evidence source & Exact excerpt \\
\midrule
Log log\_5841 & {\ttfamily [CHECKED] opened st-04 {\normalfont\textemdash{}} configured output does not match the plan.}\par {\ttfamily [ACTION] replaced st-04 with a fresh unit (as-built configuration restored).} \\
\addlinespace
Plan norm\_3982 & {\ttfamily - unit st-05: planned output mat-231} \\
\addlinespace
Command responses & {\ttfamily st-05 opened: input buffer empty. configured output: mat-232.} \\
\midrule
Interpretation & A repair precedent on st-04 applies to a different unit: st-05 is planned to output mat-231 but currently outputs mat-232. \\
\bottomrule
\end{tabularx}\par
}
\end{table}

\begin{table}[!htbp]
\centering
{
\caption{\textbf{\eliminationicon Elimination: an illustrative example.} Record excerpts and command responses from a medium-scale evaluation case.}
\label{tab:example_elimination}
\small
\begin{tabularx}{\linewidth}{@{}>{\raggedright\arraybackslash}p{0.22\linewidth} >{\raggedright\arraybackslash}X@{}}
\toprule
Evidence source & Exact excerpt \\
\midrule
Log log\_7360 & {\ttfamily [CHECKED] site visit {\normalfont\textemdash{}} mat-027 found destroyed.} \\
\addlinespace
Log log\_8680 & {\ttfamily [CHECKED] opened storage mat-070\_in\_W {\normalfont\textemdash{}} found empty, no inbound material.} \\
\addlinespace
Log log\_3569 & {\ttfamily [OBSERVED] multiple machines in the zone stopped at once.}\par {\ttfamily [CHECKED] zone-wide simultaneous stoppage {\normalfont\textemdash{}} power distribution loss confirmed.} \\
\addlinespace
Command responses & {\ttfamily mat-027 opened: input buffer mat-089 x5, mat-074 x3.}\par {\ttfamily mat-027: old unit removed, replacement installed to plan.}\par {\ttfamily
[SHIFT] not fixed yet; the shift goes on.}\par {\ttfamily st-00 opened: input buffer empty.}\par {\ttfamily mat-070: running.} \\
\midrule
Interpretation & Replacing mat-027 did not end the episode, which rejects the destroyed-unit explanation of log\_7360. A running machine (mat-070) is incompatible with the
zone-wide power loss of log\_3569. The stopped st-00 matches none of the recorded incidents and becomes the unit to investigate. \\
\bottomrule
\end{tabularx}\par
}
\end{table}

%% file: tables/tab_profiles.tex
\begin{table}[htbp]
\centering
\caption{{{\textbf{Evaluated Tacit-FLE profiles.} Document counts include maintenance logs and operating documents in each corpus. Episodes correspond to the specified fault
positions; $B$ is the turn budget. Scale conditions vary jointly in target products, corpus sizes, and interaction budgets.}}}
\label{tab:profiles}
{
\begin{tabular*}{\linewidth}{@{\extracolsep{\fill}}l r l r r r r@{}}
\toprule
Scale & Products & Type & Worlds & Documents/world & Episodes & $B$ \\
\midrule
Small & 1 & Chain & 5 & 7--16 & 13 & 30 \\
 &  & Conjunction & 5 & 5 & 7 & 30 \\
 &  & Elimination & 5 & 4--5 & 8 & 30 \\
\addlinespace
Medium & 2 & Chain & 5 & 32--40 & 14 & 60 \\
 &  & Conjunction & 5 & 30 & 14 & 60 \\
 &  & Elimination & 5 & 30 & 8 & 60 \\
\addlinespace
Large & 4 & Chain & 5 & 63--72 & 17 & 75 \\
 &  & Conjunction & 5 & 60 & 11 & 75 \\
 &  & Elimination & 5 & 60 & 12 & 75 \\
\bottomrule
\end{tabular*}\par
}
\end{table}

%% file: sections/appendix_validation.tex
{
\subsection{Benchmark construction checks}
\label{app:validation}
\label{app:construction_checks}
\label{app:benchmark_controls}

These checks assess whether generated cases realize the intended task type: an operating plant develops an observable fault, and its corpus provides the specified evidence.
Table~\ref{tab:benchmark_validation} summarizes the checks on plant behavior, corpus construction, and reproducibility.
}
\input{tables/tab_benchmark_validation}
{

}

%% file: tables/tab_benchmark_validation.tex
\begin{table}[htbp]
\centering
{
\caption{{\textbf{Benchmark construction checks.} The checks cover simulated plant behavior, corpus conditions, and reproducible evaluation inputs.}}
\label{tab:benchmark_validation}
{
\begin{tabularx}{\linewidth}{@{}>{\raggedright\arraybackslash}p{0.20\linewidth} >{\raggedright\arraybackslash}X >{\raggedright\arraybackslash}X@{}}
\toprule
Check & Evidence & What it establishes \\
\midrule
Plant behavior & Healthy production, followed by stalled shipping or a persistent change in operating readings after fault injection. & The injected fault has an observable effect on an operating plant. \\
Corpus construction & Required document fields and vocabulary, declared evidence coverage, and exclusion of test-fault repair records. & Compliance with the corpus construction conditions. \\
Reproducibility & Exact regeneration of all 45 evaluation corpora. & Consistent corpus inputs for the 104 evaluation episodes. \\
\bottomrule
\end{tabularx}\par
}
}
\end{table}

%% file: sections/appendix_method.tex
\section[Method Details]{{Method Details}}
\label{app:method}

\subsection[Representation and information accounting]{{Representation and information accounting}}
\label{app:accounting}

{
\rev{The world knowledge is $\worldknowledge=(\schema,A_W,R_W)$, with an ontology schema $\schema=(\mathcal{C},\mathcal{R},\mathcal{P},\Theta)$.}
Here $\mathcal{C}$ is the set of entity types, $\mathcal{R}$ the relations, $\mathcal{P}$ the observable attributes, and $\Theta$ the admissible axiom templates with their evaluation procedures.
{The induced axioms comprise ground assertions in $A_W$, which preserve particular entities and their connections, and conditional axioms in $R_W$.}
{Equation~\ref{eq:method_current_knowledge} combines these with the episode assertions $\episodestate$; Algorithm~\ref{alg:inference} uses the same tuple with the within-turn variables $A$ and $K$.}
The templates differ by environment, as listed in Table~\ref{tab:instantiation}.
{The predicate $\op{Typed}$ in Equation~\ref{eq:method_source_validity} checks conformance to this vocabulary and its admissible axiom forms; empirical support is evaluated separately.}
}

{
An induced element retains a provenance set $\pi\subseteq\corpus$ identifying its source records.
Some sources are attached automatically by the induction procedure, while other templates require the LLM to cite record identifiers.
Source association provides traceability; the additional support and compatibility checks are described below.
\rev{Table~\ref{tab:accounting} distinguishes fixed representation choices from induced knowledge and current observations.}
The experimental corpus and observation conditions, including the separately reported observation control, are specified in Appendix~\ref{app:corpus_protocol}.
}
\input{tables/tab_accounting}

{
\textbf{Knowledge lifetime.}
The corpus-derived world knowledge is reused without learning from test episodes.
Each episode starts from that knowledge and obtains a fresh state $\episodestate$ from observations.
Historical status readings are not copied into the live state; roster information can correct an entity's type in the episode's working representation.
The distinction between world knowledge and episode state concerns lifetime and evidential scope.
Both may contain assertions about individuals.
}

\subsection[Knowledge construction and admission]{{Knowledge construction and admission}}
\label{app:admission}

{
\rev{\textbf{Shared procedure.}
This section expands Algorithm~\ref{alg:induction} and Equation~\ref{eq:method_axiom_admission} into the checks used to admit corpus-derived knowledge in each environment.}
{$f_{\mathrm{abs}}$ in Equation~\ref{eq:method_extraction} denotes the proposal procedure, including the association of each candidate with its source records.}
{The admission indicator $\op{Admissible}$ in Equation~\ref{eq:method_axiom_admission} requires schema and source validity ($\beta_i$), then applies the constraints $\mathcal C_x$ supported by the knowledge type.}
{The indicator $\mathbf1[P]$ converts a satisfied condition to one and a failed condition to zero; the product in Equation~\ref{eq:method_axiom_admission} is one exactly when every applicable criterion holds.}
The membership and semantics of $\mathcal C_x$ differ by knowledge type: source-link requirements, structural consistency, empirical support, and contradiction tolerances are specified below.
{$f_{\mathrm{corpus}}$ denotes validation against the assembled records: some checks run during admission, while others require a later pass over the assembled knowledge.}
Bounded correction requests within knowledge construction are included in the proposal procedure.
The deterministic checks run without LLM calls.
}

{
\textbf{\benchmark{}: structure and diagnostic patterns.}
The LLM processes each reference document using the supplied vocabulary and proposes entities, structural relations, diagnostic patterns, and non-evidence axioms.
Entity types and relation names are closed; relation domain and range constraints and incompatible component types are checked when proposals enter the store.
The current document identifier is attached to its extracted facts and patterns.
A diagnostic pattern specifies the symptom type and status, the cause's relation to that symptom, optional cause type and readings, and a repair action when the documented intervention has a corresponding command.
For example, an incoming-flow condition is retained only when the source record supplies it.
Relations preserve particular component identities, whereas the diagnostic pattern refers to types and roles.
}

{
When repair observations can be extracted from the cited records, a support check tests whether at least one cited incident instantiates the pattern's symptom--cause relationship.
An unsupported pattern is returned to the LLM once with the discrepancy; a revised proposal that still fails is discarded.
Operating documents without a repair incident are not evaluated by this repair-specific test.
After assembling the corpus, a further check removes self-cause patterns that would confirm themselves from the symptom alone when a recorded incident shows that symptom but documents a repair of another component.
}

{
\textbf{\benchmark{}: operating information.}
Operating practices and service-count samples are elicited from the joined corpus because their evidence can span documents.
Practice templates express a forbidden action under a stated condition, a check before an intervention, or a preference between actions.
They require a valid typed form and nonempty citations drawn from the presented corpus, but do not undergo the repair-pattern support test.
Non-evidence axioms encode explicit operating notes that make a status non-diagnostic for a component class, with any stated exceptions.
Service samples are checked against their cited records and aggregated by production duty into an empirical prior over usage counts.
}

{
\textbf{\mars{}.}
The LLM reads the reference transitions and proposes typed action conditions and effects for collection, crafting, placement, movement, drinking, creatures, and material neighborhoods, together with blocked-action conditions.
Admission checks the schema and requires at least one cited transition concerning the subject of the proposed axiom.
Applicable recorded transitions are then checked for outcomes or effects that contradict it.
The experiment configuration allows at most one contradictory transition; blocked-action rules additionally require recorded unsuccessful attempts without a recorded success under the claimed condition.
This compatibility check is performed during admission.
}

{
\textbf{\alfworld{}.}
Records are grouped by action and processed in batches of at most 24 transitions.
The LLM proposes necessary action conditions over five supplied predicates: being at the target, holding the object, having empty hands, the object's presence at the target, and the target's openness.
Every cited index must identify a record of the corresponding action.
At least one cited successful transition must observe and satisfy all proposed conditions, and no successful transition in the corpus may contradict them.
These checks establish positive support and compatibility with recorded successes.
}

\subsection[Diagnostic hypotheses and verification]{{Diagnostic hypotheses and verification}}
\label{app:diagnostic}

{
\textbf{Hypothesis representation.}
In the diagnostic variant, a hypothesis is written as $h=(c,e,r,\Phi,\sigma,\pi,\alpha)$.
Its fields are a candidate cause $c$, the anomaly $e$ it explains, a referenced pattern $r$, expected observational atoms $\Phi$, required structural relations $\sigma$, cited evidence $\pi$, and a prescribed repair $\alpha$.
The pattern can refer to an induced diagnostic pattern or aggregate prior, or describe an explanation proposed by the LLM beyond those patterns.
An atom names an entity, an observable attribute, a comparison, and an expected value.
}

{
\textbf{Admission of episode claims.}
{$f_{\mathrm{admit}}$ checks the structured claims proposed during interaction.}
Named entities and anomalies must exist in the current representation, the expected attributes must be readable, and required relations must already be established.
A newly proposed relation must first be supported by a permitted basis, such as documented structure, layout connectivity, or observed contents.
For a pattern-backed diagnosis, admission also checks the correspondence between the proposed cause, the symptom, and the conditions of that pattern.
A hypothesis cannot cite a pattern requiring no incoming material while omitting the corresponding expected reading.
An operating note that makes a status non-evidence can invalidate a claim relying on that status.
}

{
\textbf{Three-valued evaluation.}
The observation map $\rho$ grounds readings in the current state, and an atom evaluates as
\[
\sem{\varphi}_{K_t}\in\{\top,\bot,\unk\}.
\]
The values denote support, contradiction, and an unresolved reading, respectively.
Missing readings remain unknown.
For a current status that disagrees with the claim, the implementation also considers recent readings and can leave an isolated fluctuation unresolved.
{An open, admitted hypothesis is refuted when an atom evaluates to $\bot$, confirmed when all its atoms evaluate to $\top$, and remains open otherwise.}
For open hypotheses, unresolved atoms are mapped to available probes and included in the context supplied to the LLM.
Confirmation records agreement with the specified observations.
}

{
\textbf{Hypotheses beyond recorded patterns.}
The LLM may propose an explanation for a cause not covered by an induced pattern, provided that its claim meets the structural and observational admission requirements.
The reasoning context also lists recorded explanations that still owe a reading, open rival hypotheses, and the stopped units that no recorded pattern explains; an explanation outside the recorded patterns is marked as supported when its cause is one of these unexplained units.
}

{
\textbf{Checking repairs and incorporating outcomes.}
{A repair is backed when an admitted LLM-proposed hypothesis is confirmed, its nonempty set of atoms still holds, it prescribes the repair verb, and the target is named by the hypothesis's cause or one of its atoms.}
Observation actions do not require a confirmed repair hypothesis; operating practices and repetition checks can nevertheless provide feedback on proposed commands.
If an executed repair targets a hypothesis's candidate cause and the episode continues, that outcome refutes the hypothesis on the next observation.
A command refused by the environment is not counted as an executed repair.
Confirmed hypotheses can also be refuted by subsequent contradictory readings or by the disappearance of the anomaly they explained.
The LLM remains responsible for selecting a cause and an action from this evidence.
When the proposal limit is reached, the final proposal and its verification verdict are passed to the executor.
}

\input{sections/appendix_method_example}

\Needspace{6\baselineskip}
\subsection[Learning scope and prompt contracts]{{Learning scope and prompt contracts}}
\label{app:scope}
\label{app:prompts}

{
{The separation between reusable knowledge and episode state follows Appendix~\ref{app:accounting}.}
\rev{The ontology schemas fix the vocabulary and rule forms.}
{$f_{\mathrm{abs}}$ and $f_{\mathrm{propose}}$ are implemented through LLM calls; their inputs and output contracts are given below.}
Verification consists of the typed constraints, relation checks, and observational predicates of Equations~\ref{eq:method_admission_test} and \ref{eq:method_atom_evaluation}--\ref{eq:method_proposal_verification}.
}

{
\textbf{Inputs and outputs.}
Knowledge-construction prompts provide the schema and the corresponding record batch, requesting structured facts or axioms.
The diagnostic action prompt provides world knowledge with its sources, layout facts, current readings, previous commands, hypothesis status, and available actions.
It requests optional structural claims and hypotheses followed by one executable command.
In \mars{} and \alfworld{}, the actor instead receives the learned action conditions, task context, observations, and feedback from failed condition checks.
The following blocks summarize the two \benchmark{} prompt contracts; wording is condensed and bracketed fields stand for the corresponding inputs.
}

{
\promptbox{Knowledge construction: record-level extraction}{
Input: [typed schema] and [reference document].\\
Extract the entities and structural relations stated or implied by the record.\\
Generalize the incident using component types, the symptom, the cause's relation to it, any stated observational conditions, and the documented repair.\\
Do not copy historical status readings into persistent structural facts.\\
Return structured entities, relations, diagnostic patterns, and explicit non-evidence policies. The current record identifier is attached as provenance.
}
}

{
\promptbox{Diagnostic interaction: explanation and action}{
Input: [world knowledge and sources], [layout], [observations], [command history], [hypotheses and their standing], [verification feedback].\\
Propose which component could explain an observed anomaly and what readings would support that explanation.\\
Name the pattern used, its required relations, expected readings, evidence, and a repair. Claim new structure with a permitted basis.\\
Use unresolved conditions and contradictory evidence to decide what to observe or revise.\\
Return optional structured claims followed by one command: inspect, open, wait, replace, or refill.
}
}

%% file: tables/tab_accounting.tex
\begin{table}[!htbp]
\centering
{
\caption{\textbf{Information accounting for \method{}.} The origins and roles of information used in construction and interaction.}
\label{tab:accounting}
\renewcommand{\arraystretch}{1.12}
\begin{tabularx}{\linewidth}{@{}>{\raggedright\arraybackslash}p{0.21\linewidth} >{\raggedright\arraybackslash}X >{\raggedright\arraybackslash}X@{}}
\toprule
Origin & Content & Role \\
\midrule
Task inputs
 & Reference corpus, task description, observations, action interface
 & Use and any observation controls follow the experimental protocol. \\
Given representation
 & Schema $\schema$, observation map $\rho$, template semantics
 & Method-specific representation and interpretation choices; fixed before learning. \\
Induced knowledge
 & $A_W$, $R_W$, source records $\pi\subseteq\corpus$
 & Constructed from the reference corpus and fixed during evaluation. \\
Episode evidence
 & $\episodestate$
 & Grounded in public observations; reset for each episode. \\
LLM proposals
 & Action $a$; structural and diagnostic claims where supported
 & Checked against the current knowledge and evidence. \\
\bottomrule
\end{tabularx}
}
\end{table}

%% file: sections/appendix_method_example.tex
\Needspace{6\baselineskip}
\subsubsection[Example: from corpus records to action]{{Example: from corpus records to action}}
\label{app:worked_example}

{
This example starts with a small reference corpus, shows the knowledge constructed from it, and follows its use in a new episode.
The records and observation branches are condensed, fictional examples.

\textbf{Step 1: read the records and construct world knowledge.}
Suppose the agent receives the following two documents, $\corpus=\{d_1,d_2\}$, before evaluation.
\methodexample[Reference corpus]{%
\par\smallskip
\textbf{$d_1$: Line A handover note.}
``Storage $s_1$ supplies iron plates to machine $m_1$.
Mover $q_1$ delivers plates into $s_1$ and is its only incoming connection.''
\par\smallskip
\textbf{$d_2$: Line A maintenance record.}
``Machine $m_1$ had stopped.
Its supplying storage $s_1$ contained no iron plates.
Inspection of its sole incoming mover $q_1$ found no material arriving.
We added iron plates to $s_1$; $m_1$ then resumed production.''
}
\rev{The ontology schema $\schema$ provides the types \texttt{Storage}, \texttt{Machine}, and \texttt{Mover}, the relation \texttt{feeds}, and the observable attributes.}
The LLM proposes a particular supply relation and a reusable diagnostic pattern.
The relation has compatible endpoint types; the pattern's cited incident matches its stopped-machine symptom, supplying-storage cause, and refill intervention.
After the applicable admission checks, $\worldknowledge=(\schema,A_W,R_W)$ retains
\[
\begin{aligned}
A_W&\supseteq\{\mathrm{feeds}(s_1,m_1)\}
&&\text{supported by }d_1,d_2,\\
R_W&\ni r
&&\text{with source }\pi_r=\{d_2\}.
\end{aligned}
\]
The content of $r$ makes the reusable roles explicit:
\[
\begin{aligned}
\text{symptom: }&m:\mathrm{Machine}\text{ is stopped},\\
\text{candidate cause: }&s:\mathrm{Storage}\text{ is depleted},\\
\text{required relation: }&\sigma_r(s,m)=\{\mathrm{feeds}(s,m)\},\\
\text{expected readings: }&\Phi_r(s)=\{\mathrm{stock}(s)=0,\;\mathrm{inflow}(s)=0\},\\
\text{prescribed repair: }&\texttt{refill}(s).
\end{aligned}
\]
The concrete identities remain in $A_W$, while $r$ replaces them with roles that can match another storage--machine pair.
The empty-stock and no-inflow conditions come from $d_2$; they specify what to check when proposing the same explanation for a new pair.
Here and below, $\mathrm{inflow}=0$ means that no incoming material is observed.

\textbf{Step 2: apply the pattern to a new situation.}
Now consider a different pair, $s_2$ and $m_2$.
The current layout and observations provide the following information.
\methodexample[Current episode]{%
\par\smallskip
\textbf{Layout.} Storage $s_2$ supplies machine $m_2$; mover $q_2$ is the sole incoming mover connected to $s_2$.
\par\smallskip
\textbf{Observed.} Machine $m_2$ is stopped, and inspection of $s_2$ finds zero iron plates.
\par\smallskip
\textbf{Not yet observed.} Mover $q_2$ has not been inspected, so whether material is arriving remains unresolved.
}
The layout establishes $\mathrm{feeds}(s_2,m_2)$ in the episode evidence $A_t$.
The historical observation of $q_1$ says nothing about the current inflow through $q_2$.
Using $r$, the LLM proposes $h$: ``Depletion of $s_2$ may explain the stopped $m_2$; the corresponding repair is to refill $s_2$.''
Substituting $s\mapsto s_2$ and $m\mapsto m_2$ gives
\[
\begin{aligned}
\sigma(h)&=\sigma_r(s_2,m_2)=\{\mathrm{feeds}(s_2,m_2)\},\\
\Phi(h)&=\Phi_r(s_2)=\bigl\{
\underbrace{\mathrm{stock}(s_2)=0}_{\varphi_{\mathrm{stock}}},\;
\underbrace{\mathrm{inflow}(s_2)=0}_{\varphi_{\mathrm{inflow}}}\bigr\}.
\end{aligned}
\]
Each $\varphi$ is a condition to test; for example, $\varphi_{\mathrm{stock}}$ has the tuple form $(s_2,\mathrm{stock},=,0)$ in Equation~\ref{eq:method_atom_evaluation}.
The proposal retains both conditions from $d_2$, including the inflow condition that has not yet been observed in this episode.
Both attributes are observable, so the complete hypothesis can be admitted before every reading is available.

Table~\ref{tab:method_example_admission} contrasts a complete proposal with three incomplete proposals, holding the other admission requirements fixed and satisfied.
For the missing-relation case, assume the layout and records do not establish the supply connection.
A missing reading is compatible with admission; an unsupported relation or omitted required condition is not.
}
\input{tables/tab_method_example_admission}
{
An inadmissible proposal has not entered the hypothesis set.
This differs from a \emph{refuted} hypothesis, which was admitted and then contradicted by evidence.

\textbf{Step 3: inspect the missing condition and assess the hypothesis.}
Equation~\ref{eq:method_atom_evaluation} compares each expected condition with its current reading: $\top$ means supported, $\bot$ means contradicted, and $\unk$ means unresolved.
Equation~\ref{eq:method_hypothesis_status} then combines those two results into the status of $h$.
In the current episode above, zero stock supports $\varphi_{\mathrm{stock}}$, while $\varphi_{\mathrm{inflow}}$ is unresolved.
Thus $h$ is open (case A): the past incident suggests an explanation, but the new evidence does not yet confirm it.
The LLM can request \texttt{inspect}$(q_2)$ to learn whether material is arriving.
With stock still zero and $m_2$ still stopped, the response determines which branch follows.
In each pair below, the first entry evaluates stock and the second inflow:
\[
\begin{aligned}
\text{A: }(\top,\unk),\ \text{open}
&\xrightarrow{\text{no material arriving}}
\text{B: }(\top,\top),\ \text{confirmed},\\[3pt]
\text{A: }(\top,\unk),\ \text{open}
&\xrightarrow{\text{material arriving}}
\text{C: }(\top,\bot),\ \text{refuted}.
\end{aligned}
\]
If inspection still leaves inflow unresolved, the case remains A and $h$ remains open.
The no-inflow branch agrees with $d_2$'s expected conditions; the inflow branch contradicts one of them, even though the storage is empty in both cases.
Table~\ref{tab:method_example_observations} gathers these branches and two further alternatives, with the structure and stopped-machine anomaly held fixed.
}
\input{tables/tab_method_example_observations}
{
In case D, the observed stock of five already contradicts the empty-storage condition; the unread inflow does not prevent refutation.
Case E instead has no resolved condition, so there is no contradiction and no confirmation.

\textbf{Step 4: check the proposed repair against the confirmed explanation.}
Continue along branch B, with both conditions supported and $h$ confirmed.
The LLM proposes \texttt{refill}$(s_2)$, transferring the repair in $d_2$ to the storage occupying the same role in the new episode.
The action matches $r$'s prescribed verb and $h$'s candidate cause; both required readings currently hold, so it is backed by $h$.
For Table~\ref{tab:method_example_actions}, take $H'=\{h\}$ and let $s_3$ be another storage that is neither the cause nor an entity tested by $h$.
We write only each command's verb and target, assuming valid material and amount arguments for a refill.
{The table evaluates the repair-backing requirement in Appendix~\ref{app:diagnostic}; other action checks still apply.}
}
\input{tables/tab_method_example_actions}
{
Rows R1--R3 show that the same confirmed explanation can back one repair and fail to back another.
{R4 considers a later state in which the recorded confirmation remains but a current reading is unresolved: this is why the repair check rechecks all expected conditions.}
O1 returns to case A and shows that the agent can seek evidence while the hypothesis is still open; an observation does not require repair backing.

\textbf{Step 5: use the intervention outcome as new evidence.}
Suppose the refill in R1 is executed.
Table~\ref{tab:method_example_outcomes} separates possible responses from the environment.
The refutation rule for an executed repair that leaves the episode running is the diagnostic outcome rule of \benchmark{} described in Appendix~\ref{app:diagnostic}.
}
\input{tables/tab_method_example_outcomes}
{
For the continuing-episode branch, the transition is
\[
\underbrace{\text{confirmed}}_{\text{expected readings supported}}
\;\xrightarrow{\text{refill executed; episode continues}}\;
\underbrace{\text{refuted}}_{\text{intervention contradicts the explanation}}.
\]
This update uses intervention evidence, whereas Equation~\ref{eq:method_hypothesis_status} assessed the initially open hypothesis from its expected readings.
A refused command supplies no such intervention evidence; new observations may still change the assessment.
}

%% file: tables/tab_method_example_admission.tex
\begin{table}[H]
\centering
{\small
\setlength{\tabcolsep}{3pt}
\renewcommand{\arraystretch}{1.1}
\caption{{\textbf{Admission before observation-based assessment.}}}
\label{tab:method_example_admission}
\begin{tabularx}{\linewidth}{@{}>{\raggedright\arraybackslash}p{0.13\linewidth} >{\raggedright\arraybackslash}p{0.35\linewidth} >{\raggedright\arraybackslash}X@{}}
\toprule
{Case} & {Proposal} & {Admission and reason} \\
\midrule
{Complete} & {Established supply relation; both required conditions retained; inflow still unread.} & {Admitted: Missing readings can be investigated after admission.} \\
{Missing relation} & {The proposal names a supplier relation that has no supporting evidence.} & {Not admitted: Establish the relation before applying this pattern.} \\
{Missing condition} & {The proposal cites the pattern but checks only whether storage is empty.} & {Not admitted: Retain the required no-inflow condition.} \\
{No conditions} & {The explanation supplies no observable conditions.} & {Not admitted: An explanation needs a nonempty set of testable conditions.} \\
\bottomrule
\end{tabularx}
}
\end{table}

%% file: tables/tab_method_example_observations.tex
\begin{table}[H]
\centering
{\small
\setlength{\tabcolsep}{3pt}
\renewcommand{\arraystretch}{1.1}
\caption{{\textbf{Different evidence for the same proposed explanation.} Independent alternatives; each case evaluates both expected conditions.}}
\label{tab:method_example_observations}
\begin{tabularx}{\linewidth}{@{}>{\centering\arraybackslash}p{0.05\linewidth} >{\centering\arraybackslash}p{0.07\linewidth} >{\centering\arraybackslash}p{0.08\linewidth} >{\centering\arraybackslash}p{0.14\linewidth} >{\centering\arraybackslash}p{0.16\linewidth} >{\raggedright\arraybackslash}X@{}}
\toprule
{Case} & {Stock} & {Inflow} & {\shortstack{Values\\(stock, inflow)\\Eq.~\ref{eq:method_atom_evaluation}}} & {\shortstack{Status\\Eq.~\ref{eq:method_hypothesis_status}}} & {Reason and next step} \\
\midrule
{A} & {0} & {Unread} & {$(\top, \unk)$} & {Open} & {Empty stock supports one condition; inspect incoming supply to resolve the other.} \\
{B} & {0} & {None} & {$(\top, \top)$} & {Confirmed} & {Both expected conditions are supported. Check the proposed repair next.} \\
{C} & {0} & {Arriving} & {$(\top, \bot)$} & {Refuted} & {Observed inflow contradicts the no-inflow condition. Revise the explanation.} \\
{D} & {5} & {Unread} & {$(\bot, \unk)$} & {Refuted} & {Nonzero stock already contradicts the claim; the unread inflow cannot rescue it.} \\
{E} & {Unread} & {Unread} & {$(\unk, \unk)$} & {Open} & {Neither condition has been resolved. Obtain an informative observation.} \\
\bottomrule
\end{tabularx}
}
\end{table}

%% file: tables/tab_method_example_actions.tex
\begin{table}[H]
\centering
{\small
\setlength{\tabcolsep}{3pt}
\renewcommand{\arraystretch}{1.1}
\caption{{\textbf{Hypothesis confirmation and action backing.} Independent alternatives evaluated with $H' = \{h\}$; the backing check uses this candidate only.}}
\label{tab:method_example_actions}
\begin{tabularx}{\linewidth}{@{}>{\centering\arraybackslash}p{0.05\linewidth} >{\raggedright\arraybackslash}p{0.16\linewidth} >{\centering\arraybackslash}p{0.14\linewidth} >{\centering\arraybackslash}p{0.14\linewidth} >{\centering\arraybackslash}p{0.13\linewidth} >{\raggedright\arraybackslash}X@{}}
\toprule
{Case} & {Proposal} & {\shortstack{Recorded\\status}} & {\shortstack{Conditions\\(stock, inflow)\\Eq.~\ref{eq:method_atom_evaluation}}} & {Backed} & {Reason} \\
\midrule
{R1} & {\texttt{refill}($s_2$)} & {Confirmed} & {$(\top, \top)$} & {Yes} & {Prescribed verb, matching target, and both conditions supported.} \\
{R2} & {\texttt{replace}($s_2$)} & {Confirmed} & {$(\top, \top)$} & {No} & {The pattern prescribes a refill, not replacement.} \\
{R3} & {\texttt{refill}($s_3$)} & {Confirmed} & {$(\top, \top)$} & {No} & {The target is neither the cause nor an entity tested by this hypothesis.} \\
{R4} & {\texttt{refill}($s_2$)} & {Confirmed} & {$(\top, \unk)$} & {No} & {Stored confirmation is insufficient when a current condition is unresolved.} \\
{O1} & {\texttt{inspect}($q_2$)} & {Open} & {$(\top, \unk)$} & {Not required} & {An observation does not require a confirmed repair hypothesis.} \\
\bottomrule
\end{tabularx}
}
\end{table}

%% file: tables/tab_method_example_outcomes.tex
\begin{table}[H]
\centering
{\small
\setlength{\tabcolsep}{3pt}
\renewcommand{\arraystretch}{1.1}
\caption{{\textbf{Intervention outcomes after a backed refill.}}}
\label{tab:method_example_outcomes}
\begin{tabularx}{\linewidth}{@{}>{\raggedright\arraybackslash}p{0.37\linewidth} >{\raggedright\arraybackslash}p{0.19\linewidth} >{\raggedright\arraybackslash}X@{}}
\toprule
{Event} & {Result} & {Meaning} \\
\midrule
{The refill executes and the episode terminates successfully.} & {Task completed} & {The episode ends; no additional hypothesis status is needed.} \\
{The refill executes, but another episode turn follows.} & {Hypothesis refuted} & {Under the diagnostic outcome rule, the intervention contradicts this candidate-cause claim.} \\
{The environment refuses the refill command.} & {No intervention evidence} & {Refusal alone does not refute the hypothesis; inspect the feedback and current evidence.} \\
\bottomrule
\end{tabularx}
}
\end{table}

%% file: sections/appendix_bindings.tex
\section{Environment Bindings and Baselines}
\label{app:bindings}

\subsection[Instantiation per environment]{{Instantiation per environment}}
\label{app:instantiation}

{
Table~\ref{tab:instantiation} instantiates the construction and interaction procedures in Section~\ref{sec:method}.
The common principle is to induce structured knowledge from the reference corpus and use it to check proposals against current evidence.
The representations and verification procedures specialize this principle to each environment.
\benchmark{} combines component relations, diagnostic patterns, and an episode-specific hypothesis ledger; \mars{} uses typed action and effect templates; \alfworld{} uses conditions over observed scene predicates.
Appendix~\ref{app:admission} specifies their admission checks.
}
\input{tables/tab_instantiation}

{
\textbf{Observation grounding.}
\rev{The function $f_{\mathrm{ground}}$ in Algorithm~\ref{alg:inference} updates episode assertions from observations using an environment-specific map $\rho$, left implicit in the algorithm.}
In \benchmark{}, component types and positions come from the public roster, while layout and readouts supply connection and state evidence.
The mapping from a stock trend to an inflow reading is a fixed interpretation of public observations, rather than an induced axiom.
A stock level that remains at zero does not alone settle whether material is arriving; other readings can be needed.
These observation mappings are supplied by the environment binding.
The \mars{} binding exposes position, inventory, vitals, and visible cells; the \alfworld{} binding maps parsed scene information onto the five predicates used by its learned conditions.
}

{
\subsection{Baseline implementations and prompts}
\label{app:baselines}
\raggedbottom

We implement each baseline's knowledge representation and decision procedure using the published method and available reference code.
Environment bindings supply the task description, observation format, action syntax, and reference records.
The descriptions below specify how each method uses the corpus in our experiments.
Where a released implementation omits a component or departs from the published procedure, we follow the paper; such adjustments are noted with each method.

\paragraph{\react{} and \reactsearch{}.}
\react{} alternates a textual thought with an executable action based on the task description and observations.
\reactsearch{} supplements the same actor with up to five corpus records retrieved once at the start of an episode.
The retrieval query uses the initial observation and, in \benchmark{}, the plant roster.
The retrieved records remain available in the episode context.
Both variants share the actor prompt, action parser, and history handling within an environment.

\paragraph{\reflexion{}.}
We use a corpus-based adaptation of reflective memory.
Before evaluation, the model reads each reference document and writes a memory note describing useful evidence and conditional plans.
At the start of an episode, the initial-observation query retrieves up to three relevant notes for the actor.
The reflection prompt asks the model to distinguish observed outcomes from inferred advice and to preserve the conditions under which an action is useful.
This adaptation constructs the reflective memory from the supplied records.

\paragraph{\ifr{}.}
The induction prompt converts recorded interactions into a natural-language rule library.
On \benchmark{}, repair records additionally populate a skill library describing successful procedures.
The planner uses the task, observations, and retrieved knowledge to propose actions; an explanation prompt interprets unsuccessful execution and supports replanning.
We route all LLM calls and actions through the shared interfaces while retaining the separate induction, planning, explanation, and replanning stages; the reference corpus initializes the rule library and, on \benchmark{}, the skill library; episode history is retained across replanning rounds, and library entries above the similarity threshold are replaced by the newly induced rule.

\paragraph{\walle{}.}
\label{app:implementation_adjustments}
The learning stage compares predicted action outcomes with recorded outcomes.
Prediction discrepancies drive natural-language rule proposals, which are compiled into executable checks and validated against recorded transitions.
During evaluation, a proposer and world model use the learned knowledge, and the executable rules check candidate actions before execution.
Our bindings present corpus records in the input format used by this learning procedure.
The public release does not include the \mars{} learning and evaluation pipeline; we assemble it from the published procedure, the released prompts, and the released \alfworld{} learning and verification components, align observation and action representations across corpus ingestion, rule validation, and execution, and fix interface errors in the released code.
Candidate rejection and reproposal follow the released execution loop.
On \benchmark{} (Table~\ref{tab:main}), disagreement between an action-matched executable check and the LLM prediction triggers one further query about the same action, which includes the code verdict and feedback, following the alignment step in \citet[Eq.~9]{zhou2025walle}; the revised prediction determines rejection, and when no matching check can be evaluated, the initial prediction is retained.
A passing Boolean check records the absence of an encoded failure condition.
The action-proposal loop permits five attempts per environment turn and passes the final proposal to the executor if the limit is reached.
The \mars{} binding supplies spatial context through the current observation and bounded interaction history.

\paragraph{\nesyc{}.}
We reconstruct the induction and rule-application procedures in Algorithms~1--2 of \citet{choi2025nesyc}, using the released implementation as a reference.
The learner proposes answer-set rules from the corpus and refines them using contrasting transition examples.
At evaluation, observations are translated into facts and combined with the learned program for symbolic planning.
We connect the hypothesis score to iterative rule revision, completing the refinement loop described in the paper: solver checks provide feedback on labeled examples, with false positives penalized according to the published objective, and statements that fail to parse or ground are removed before scoring and reported to the learner.
The bindings translate observations and task goals into facts, support the induction of action effects from the corpus, annotate completion in corpus records with the planner's state vocabulary, and expose inspection results as structured attributes.
The \benchmark{} maintenance logs contain no failed-action examples, so the hypothesis score reduces to its positive-example term there.

\paragraph{Prompt organization.}
Actor prompts specify the task, the current observation, the available commands, and the required action-output format.
Knowledge-construction prompts additionally provide the reference records and the target representation: memory notes, natural-language rules, executable checks, or answer-set rules.
Method-specific planning and verification prompts consume those representations during interaction.
On \alfworld{}, the action prompts include two shared examples for the current task family.
Retrieval in \reactsearch{}, \reflexion{}, and \ifr{} uses the same embedding model.
The code release will include the full prompt templates.

\subsubsection{Representative prompts}
\label{app:baseline_prompts}

The following boxes show the core instructions and response formats of the implemented prompts, organized by stage as in the prompt appendices of \citet{tang2024mars} and \citet{zhou2025walle}.
Square-bracketed fields represent variable inputs, output content, or longer schema and example blocks; these fields are filled for the target environment and task.
Instruction wording is condensed for readability.
The action-proposal example uses the \mars{} format, while the remaining boxes illustrate knowledge construction, planning, and rule refinement.
\input{sections/appendix_prompt_examples}
\flushbottom
}

%% file: tables/tab_instantiation.tex
\begin{table}[!htbp]
\centering
{
\caption{\textbf{Environment-specific instantiations of \method{}.} The common construction and interaction procedures use the following representations and checks.}
\label{tab:instantiation}
\renewcommand{\arraystretch}{1.12}
\begin{tabularx}{\linewidth}{@{}>{\raggedright\arraybackslash}p{0.17\linewidth} >{\raggedright\arraybackslash}X >{\raggedright\arraybackslash}X >{\raggedright\arraybackslash}X@{}}
\toprule
Element & \benchmark{} & \mars{} & \alfworld{} \\
\midrule
Corpus $\corpus$
 & One plant's maintenance logs and operating documents
 & Recorded transitions from one world
 & Recorded transitions from training games \\
Vocabulary $\schema$
 & Component types, material-flow relations, statuses, and public readings
 & Materials, items, creatures, action and effect templates
 & Action verbs and five observation predicates \\
Induced knowledge
 & Structural facts, diagnostic patterns, non-evidence axioms, practices, aggregate priors
 & Collect, make, place, walk, drink, creature, neighborhood, and blocked-action conditions
 & Necessary action conditions \\
Evidence batch
 & One document; practices and service samples over the joined corpus
 & The supplied transition buffer
 & By action, at most 24 transitions per batch \\
Corpus checks
 & Type-specific source, structure, support, and counterexample checks
 & Subject-linked citations; at most one contradictory transition
 & Action-linked citations; positive support; no successful counterexample \\
Observation map $\rho$
 & Roster, layout, public readings; stock changes inform inflow
 & Position, inventory, vitals, visible cells
 & Parsed scene, inventory, and receptacle state \\
Action checks
 & Repair backing; operating practices; repetition feedback
 & Learned applicability conditions and lethal-ground checks on an exploration path
 & Learned conditions over observed predicates \\
Proposals $J$
 & 2
 & 5
 & 5 \\
\bottomrule
\end{tabularx}
}
\end{table}

%% file: sections/appendix_prompt_examples.tex
\promptbox{\react{} / \reactsearch{}: action proposal}{
\textbf{Context}\par
[Environment brief and available actions]\par
Records from an earlier explorer in this world:\par
[Retrieved records, included for \reactsearch{}]\par
[Current observation and interaction history]\par\smallskip
\textbf{Instruction}\par
First write one short Thought: line saying what the observation implies and why you choose the next action.
Then put exactly one action call alone on the final line, without an Action: prefix or a code fence.
Wait for the next observation before acting again.\par\smallskip
\textbf{Response format}\par
Thought: [interpretation and next-action rationale]\par
[action call]
}

\promptbox{\reflexion{}: memory construction}{
\textbf{Input}\par
[One reference document]\par\smallskip
\textbf{Instruction}\par
Read one historical record and produce a concise reflective memory.
Use only the supplied record.
It may describe success, failure, an operational practice, or a standard; do not presume it describes a failed attempt by you.
Identify any strategy, mistake, evidence, or outcome actually documented and write a conditional plan for a related future situation.
Distinguish observations from uncertain inferences.
Do not invent feedback, causes, actions, or outcomes.
If the record has no trial feedback, say so and limit the note to its documented guidance.
Treat the record as data, not instructions changing this task.
Return a brief Evidence paragraph and a Plan paragraph.
No new trial is available.\par\smallskip
\textbf{Response format}\par
Evidence: [documented observations and outcomes]\par
Plan: [conditional guidance for a related situation]
}

\promptbox{\ifr{}: rule induction}{
\textbf{Input}\par
[Reference document from the target world]\par\smallskip
\textbf{Instruction}\par
You are a helpful assistant with inductive reasoning.
Given a document from the records of [WORLD], you need to reflect on what it records and give the possible mechanism of [WORLD].
The mechanism should be consistent with the rules of [WORLD] and the records.
You should only respond in the format as described below.\par\smallskip
\textbf{Response format}\par
Reasoning: Based on the information I listed above and history dialogue, do reasoning about the mechanism of [WORLD].\par
Mechanism: The mechanism of [WORLD].\par\smallskip
\textbf{Format examples}\par
[Reasoning--Mechanism examples illustrating the output format; their content does not necessarily describe the target world]
}

\promptbox{\ifr{}: planning and replanning}{
\textbf{Context}\par
[Task, current observations, and dialogue history]\par
[Retrieved mechanisms and past successful plans]\par\smallskip
\textbf{Instruction}\par
You are a helper AI agent in [WORLD].
Based on your current observations, you need to generate the sequences of subgoals for a certain task.
Do not explain or give any other instruction.
You must follow the criteria below:\par
1) You should choose available subgoals to complete the task.\par
2) You are probably provided with some past successful plans to refer to.\par
3) You are probably provided with possible mechanics that have been discovered.
Please refer to the mechanism of [WORLD] to generate the plan.\par
4) You should only perform the subgoals that are feasible based on the current observations.\par\smallskip
Here are available subgoals: [SUBGOALS]\par
Here are some examples for output format: [PLAN EXAMPLES]\par
Here is the brief of [WORLD]: [BRIEF]\par\smallskip
\textbf{Replanning message after execution feedback}\par
Please fix the above errors and replan the task [TASK].
}

\promptbox{\walle{}: rule mining}{
\textbf{Input}\par
Transitions: [\{"state\_0": [state], "action": \{"name": [action name], "args": [arguments]\}, "action\_result": \{"feedback": [feedback], "success": [execution outcome]\}\}, ...]\par
Existing rules: [rule list]\par
The environment's action space: [ACTIONS]\par
The environment's state format: [STATE FORMAT]\par\smallskip
\textbf{Instruction}\par
You are responsible for mining new rules from the given transitions, ensuring that these rules differ from the ones already provided.
Focus on generating general and universal rules that are not tied to any specific item or tool.
Extract and utilize only the features that influence the outcome of the action.
Generate only the rules under what conditions the action will fail.
While generating a rule, you also need to state how to check if a transition satisfies this rule.
Please be specific as to which and how features need to be checked.
Ensure the response can be parsed by Python json.loads.\par\smallskip
\textbf{Response format}\par
\{"new\_rules": [\par
\hspace*{1em}"Rule [number]: For action [action], [failure condition]; Checking Method: [features and checks]"\par
]\}
}

\promptbox{\walle{}: rule compilation}{
\textbf{Input}\par
[Learned rule and state/action format]\par
Current Knowledge Graph: [KNOWLEDGE GRAPH]\par
Current Scene Graph: [SCENE GRAPH]\par\smallskip
\textbf{Instruction}\par
You are responsible for generating code rules by implementing the learned rules in Python.
Write a function that takes the current state and an action as inputs, optionally incorporating the graph to provide additional contextual information.
The function should return feedback, success, and suggestion.\par
feedback: a string giving the action feedback.\par
success: a bool indicating whether the action is executed successfully.
If the action type is not the action type in the rule, count as success.\par
suggestion: a string giving a suggestion based on the rule, state, and action if the action fails.\par
The state and action arguments are dicts; action has the form \{"name": "...", "args": \{...\}\}.
The graph argument has the form \{"knowledge": [knowledge graph], "scene": [scene graph]\}.
Respond only with the function, without example usage or other text.\par\smallskip
\textbf{Response format}\par
 def expected\_rule\_code(state, action, graph):\par
\hspace*{1em}[condition checks]\par
\hspace*{1em}return feedback, success, suggestion
}

\promptbox{\nesyc{}: rule induction}{
\textbf{Input}\par
[ASP grammar, environment vocabulary, and target command]\par
[Batch of reference records]\par
Background Knowledge: [background program]\par
[Rule kind] so far: [current rules, or none yet]\par\smallskip
\textbf{Instruction}\par
Use inductive logic programming in Answer Set Programming.
Learn from interpretations: a positive example should be a model of the background and hypothesis; negative examples should not.
Apply theta-subsumption when generalizing.
Use the supplied vocabulary and records.
Unknown labels are not negative examples.\par
Return the complete set of [rule kind] for the target command after reading this batch: keep earlier rules these records still support, generalize where they allow, and add what they require.
One complete rule per line.
Effects must come from observed changes or the supplied API description; do not treat an unsuccessful task as an unexecutable action.
Return only the requested ASP, without explanations or markdown.\par\smallskip
\textbf{Response format}\par
[Complete ASP rule]\par
[Additional complete ASP rules, one per line]
}

\promptbox{\nesyc{}: contrastive refinement}{
\textbf{Input}\par
[Background program and current rules]\par
[Hypothesis score and contrastive feedback on labeled examples]\par\smallskip
\textbf{Instruction}\par
Revise the learned ASP rules using the contrastive feedback.
Apply theta-subsumption: prefer the more general clause that still covers every positive example and excludes every negative one.
A positive example must be a model of background and rules; a negative example must not be.
Constraint rules are the hypothesis to revise.
Effect rules (update/4 heads) are kept as learned unless the feedback reports two rules deriving different values; in that case return the resolved effect rules as well.
Unknown labels are not negative examples.
Never define goal, action, available, step or changed in a rule head.
Return the complete revised rule set, one complete rule per line, without explanations or markdown.\par\smallskip
\textbf{Response format}\par
[Complete revised ASP rule]\par
[Additional complete ASP rules, one per line]
}

%% file: sections/appendix_results.tex
\section{Full Results and Cost Accounting}
\label{app:results}

\input{sections/appendix_tacit_results}

\input{sections/appendix_mars_results}

\subsection[\mars{} per trial]{{\mars{} per trial}}
\label{app:mars_worlds}

{
Tables~\ref{tab:mars_worlds} and~\ref{tab:mars_worlds_cont} report the recorded reward for each evaluation seed behind Table~\ref{tab:mars}.
The \llmluna{} comparison comprises seven methods, eight worlds, and seeds 0--4, for 280 episodes.
Score is computed from achievement success rates pooled over the five episodes, as defined in Appendix~\ref{app:metric_definitions}.
}
\input{tables/tab_mars_worlds}
\input{tables/tab_mars_worlds_cont}

\input{sections/appendix_mars_knowledge_use}

\input{sections/appendix_alfworld_results}

\subsection{LLM usage}
\label{app:usage}

{
We separate knowledge-construction usage from evaluation usage.
Construction costs are measured per construction run: for a plant in \benchmark{}, a world in \mars{}, or a training corpus in \alfworld{}.
Evaluation costs are summed over the associated evaluation tasks and also reported per episode.

{For \walle{} in Table~\ref{tab:main}, construction usage includes the original construction costs of every knowledge base used in the evaluation protocol of Appendix~\ref{app:corpus_protocol}, including reused knowledge.
Evaluation costs cover only the reported episodes.}

{Table~\ref{tab:tacit_usage} reports the \benchmark{} usage under \llmluna{}.
Construction is a one-time cost per plant.
\method{} processes each document with the typed schema and elicits operating practices over the joined corpus, so its construction cost per plant exceeds that of \walle{}, \ifr{}, and \reflexion{}.
Evaluation dominates the total: on medium, construction for the 15 plants sums to 0.88M tokens against 9.6M evaluation tokens for \method{}, which uses the fewest evaluation tokens per episode at every scale (excluding \nesyc{}).}

}
\input{tables/tab_tacit_usage}
\input{sections/appendix_mars_usage}
{
The \alfworld{} evaluation costs appear alongside success in Table~\ref{tab:alfworld_summary}, with the full breakdown in Appendix~\ref{app:alfworld_models}.
{Table~\ref{tab:alfworld_learning_usage} reports the separate knowledge-construction costs.}

An LLM call is one completed model response, including internal planning and verification responses; token counts are those reported by the provider.
Input/call is the total input tokens divided by the number of recorded calls; it measures the mean context supplied to the model.
Total tokens is the sum of input and output tokens before rounding, averaged per evaluation episode or per construction run as specified in each table.
Knowledge-construction, planning, explanation, and verification calls are assigned to their corresponding phase.
}
\input{tables/tab_mars_usage}
\input{tables/tab_mars_learning_usage}
\input{tables/tab_alfworld_learning_usage}
\FloatBarrier

\input{sections/appendix_tacit_diagnostics}

%% file: sections/appendix_tacit_results.tex
\begingroup
\input{macros_tacit_names}
\subsection{Repair success by scale and type}
\label{app:full_results}

Table~\ref{tab:full_results} gives the successes and denominators behind Table~\ref{tab:main}.
\input{tables/tab_full_results}
{
{
\paragraph{\nesyc{} evaluation trajectories.}
\label{app:nesyc_tacit_checks}
For \nesyc{}, repair success is 0/28 and 0/36 on the small and medium suites.
Across the 3,000 evaluation turns, every executed action was waiting.
{Of the 64 initial planner outputs, 47 return no plan and 17 return a single waiting action.}
These observations explain its absence from the repair-candidate diagnostics in Appendix~\ref{app:tacit_diagnostics}.
}

\paragraph{Model conditions.}
\label{app:tacit_models}
Table~\ref{tab:tacit_models} retains the full success, spurious-intervention, and turn comparison behind Figure~\ref{fig:tacit_model_conditions} on the same 36 medium tasks.
The model condition applies to both knowledge construction and interaction.
\walle{} and \nesyc{} are compared under \llmluna{} only (Table~\ref{tab:main}).
}
\input{tables/tab_tacit_models}

\endgroup

%% file: tables/tab_full_results.tex
\begin{table}[!htbp]
\centering
\begingroup
\input{macros_tacit_names}
\caption{\textbf{Repair success counts in the \llmluna{} condition.}
Successes over scheduled episodes behind Table~\ref{tab:main}.
The small, medium, and large suites contain 28, 36, and 40 episodes, respectively.
Unsuccessful episodes remain in the denominators. Bold: best per column, ties included.}
\label{tab:full_results}
\small
\setlength{\tabcolsep}{2pt}
\medmuskip=1mu
\begin{tabular*}{\linewidth}{@{\extracolsep{\fill}}l*{9}{c}@{}}
\toprule
& \multicolumn{3}{c}{Small} & \multicolumn{3}{c}{Medium} & \multicolumn{3}{c}{Large} \\
\cmidrule(lr){2-4}\cmidrule(lr){5-7}\cmidrule(l){8-10}
Method & Chain & Conj. & Elim. & Chain & Conj. & Elim. & Chain & Conj. & Elim. \\
\midrule
\react{} & 0/13 & 5/7 & \textbf{6/8} & 4/14 & 9/14 & 6/8 & 1/17 & 8/11 & 9/12 \\
\reactsearch{} & 2/13 & 6/7 & 3/8 & 3/14 & 9/14 & 7/8 & 4/17 & 4/11 & 8/12 \\
\reflexion{} & 1/13 & 5/7 & \textbf{6/8} & 3/14 & 6/14 & 3/8 & 2/17 & 1/11 & 6/12 \\
\ifr{} & 5/13 & \textbf{7/7} & \textbf{6/8} & 4/14 & 12/14 & \textbf{8/8} & 4/17 & 10/11 & \textbf{12/12} \\
\nesyc{} & 0/13 & 0/7 & 0/8 & 0/14 & 0/14 & 0/8 & 0/17 & 0/11 & 0/12 \\
\walle{} & 1/13 & 1/7 & 1/8 & 2/14 & 4/14 & 7/8 & 1/17 & 5/11 & 9/12 \\
\midrule
\method{} (ours) & \textbf{9/13} & \textbf{7/7} & \textbf{6/8} & \textbf{8/14} & \textbf{14/14} & \textbf{8/8} & \textbf{14/17} & \textbf{11/11} & 11/12 \\
\bottomrule
\end{tabular*}
\par
\endgroup
\end{table}

%% file: tables/tab_tacit_models.tex
\begin{table}[!htbp]
\centering
\begingroup
\input{macros_tacit_names}
\caption{\textbf{Model conditions on medium \benchmark{} (36 tasks).}
SR: repair success (\%); SI: mean spurious interventions; T: mean turns.
Bold/underline: best/second-best within each condition.}
\label{tab:tacit_models}
\setlength{\tabcolsep}{3pt}
\begin{tabular}{@{}l*{5}{r}@{}}
\toprule
 & \multicolumn{1}{c}{\react{}} & \multicolumn{1}{c}{\reactsearch{}} & \multicolumn{1}{c}{\reflexion{}} & \multicolumn{1}{c}{\ifr{}} & \multicolumn{1}{c}{\method{}} \\
\midrule
\multicolumn{6}{@{}l}{\textit{\llmluna{}}} \\
SR $\uparrow$ & 52.8 & 52.8 & 33.3 & \underline{66.7} & \textbf{83.3} \\
SI $\downarrow$ & \underline{11.3} & 13.4 & 15.4 & 11.9 & \textbf{2.4} \\
T $\downarrow$ & 41.8 & 38.7 & 46.4 & \underline{31.7} & \textbf{19.4} \\
\midrule
\multicolumn{6}{@{}l}{\textit{\llmlunareasoning{}}} \\
SR $\uparrow$ & 44.4 & 38.9 & 52.8 & \underline{86.1} & \textbf{97.2} \\
SI $\downarrow$ & \underline{9.0} & 11.4 & 9.7 & 9.8 & \textbf{0.8} \\
T $\downarrow$ & 42.8 & 45.3 & 42.1 & \underline{26.7} & \textbf{8.4} \\
\midrule
\multicolumn{6}{@{}l}{\textit{\llmterra{}}} \\
SR $\uparrow$ & 52.8 & 55.6 & 38.9 & \underline{66.7} & \textbf{100.0} \\
SI $\downarrow$ & 13.7 & 12.6 & 10.5 & \underline{7.9} & \textbf{0.4} \\
T $\downarrow$ & 32.7 & 35.3 & 42.2 & \underline{28.2} & \textbf{8.4} \\
\bottomrule
\end{tabular}
\par
\endgroup
\end{table}

%% file: sections/appendix_mars_results.tex
\begingroup
\input{macros_mars_names}
{
\subsection{\mars{} across model conditions}
\label{app:mars_models}

Tables~\ref{tab:mars_luna_reasoning} and~\ref{tab:mars_terra} give the full world-level results behind the additional conditions in Table~\ref{tab:mars_models}.
The \llmluna{} world-level results remain in Table~\ref{tab:mars}.
Each reported world-level result aggregates five evaluation episodes.
{\nesyc{}'s results appear in Tables~\ref{tab:mars}, \ref{tab:mars_luna_reasoning}, and~\ref{tab:mars_terra}.}

\paragraph{\nesyc{} planning behavior.}
{In our \llmluna{} reconstruction, 5,325 of 6,326 turns across the seven modified worlds execute a no-op.
Erroneous or overly restrictive learned rules can prevent executable plans; for example, a Terrain rule imposes a tool-inventory requirement without restricting it to the relevant action, contradicting the initial state.
The runtime falls back to waiting when planning fails, while rule refinement requires a refused executed command.
These observations concern our reconstruction (Appendix~\ref{app:baselines}).
\par}
}
\input{tables/tab_mars_luna_reasoning}
\input{tables/tab_mars_terra}
\par
\endgroup

%% file: tables/tab_mars_luna_reasoning.tex
\begin{table}[!htbp]
\centering
\begingroup
\input{macros_mars_names}
\caption{\textbf{\mars{} performance with \llmlunareasoning{}.}
\llmlunareasoning{}; five evaluation episodes per world.
Reward is mean $\pm$ sample SD; Score pools achievement success rates across episodes.
Average excludes Default; its Reward SD is computed across matched-seed world averages.
Bold/underlined: best/second-best. Metric definitions are in Appendix~\ref{app:metric_definitions}.}
\label{tab:mars_luna_reasoning}
\fontsize{8pt}{9.5pt}\selectfont
\setlength{\tabcolsep}{2pt}
\medmuskip=1mu
\begin{tabular*}{\linewidth}{@{\extracolsep{\fill}}l*{7}{c}@{}}
\toprule
World & \react{} & \reactsearch{} & \reflexion{} & \ifr{} & \nesyc{} & \walle{} & \method{} \\
\midrule
\multicolumn{8}{@{}l}{\textit{Reward $\uparrow$}} \\
Default & \underline{$12.30\pm4.82$} & $\boldsymbol{15.46\pm5.35}$ & $11.30\pm2.28$ & $10.70\pm3.21$ & $5.77\pm0.58$ & $9.90\pm2.77$ & $10.90\pm3.56$ \\
Terrain & $13.10\pm2.55$ & $10.10\pm5.87$ & \underline{$14.50\pm1.14$} & $\boldsymbol{15.98\pm2.06}$ & $-0.57\pm0.58$ & $10.30\pm3.11$ & $14.06\pm5.67$ \\
Survival & $9.70\pm5.18$ & \underline{$13.70\pm5.73$} & $9.10\pm6.78$ & $12.50\pm3.51$ & $4.07\pm3.55$ & $10.50\pm6.54$ & $\boldsymbol{13.88\pm5.60}$ \\
Task dep. & \underline{$5.70\pm2.30$} & $2.10\pm0.71$ & $2.90\pm1.30$ & $\boldsymbol{7.70\pm3.13}$ & $1.77\pm1.53$ & $3.70\pm2.51$ & \underline{$5.70\pm1.14$} \\
Terr.\ surv. & $7.50\pm2.61$ & $5.30\pm3.27$ & $6.90\pm2.59$ & \underline{$8.30\pm3.11$} & $5.10\pm1.00$ & $7.10\pm2.65$ & $\boldsymbol{8.50\pm3.44}$ \\
Terr.\ task & $5.30\pm2.28$ & $6.30\pm2.68$ & $5.70\pm2.51$ & \underline{$6.90\pm2.28$} & $0.10\pm1.00$ & $4.70\pm2.07$ & $\boldsymbol{8.10\pm1.87}$ \\
Surv.\ task & $5.50\pm5.08$ & \underline{$6.10\pm4.30$} & $5.90\pm5.07$ & $5.90\pm5.63$ & $-0.23\pm0.58$ & $2.30\pm1.48$ & $\boldsymbol{7.30\pm6.57}$ \\
All three & $0.70\pm2.19$ & $\boldsymbol{1.70\pm1.67}$ & \underline{$1.10\pm1.87$} & $0.90\pm1.10$ & $0.10\pm1.00$ & $-0.30\pm0.89$ & $0.90\pm1.79$ \\
\cmidrule(lr){1-8}
Average & $6.79\pm1.10$ & $6.47\pm1.21$ & $6.59\pm1.35$ & \underline{$8.31\pm0.68$} & $1.48\pm0.94$ & $5.47\pm0.97$ & $\boldsymbol{8.35\pm1.26}$ \\
\midrule
\multicolumn{8}{@{}l}{\textit{Score (\%) $\uparrow$}} \\
Default & \underline{$46.65$} & $\boldsymbol{68.27}$ & $26.44$ & $29.64$ & $3.86$ & $15.82$ & $30.26$ \\
Terrain & $32.57$ & $22.80$ & $35.52$ & $\boldsymbol{56.87}$ & $0.17$ & $19.66$ & \underline{$52.42$} \\
Survival & $21.70$ & \underline{$56.98$} & $23.66$ & $43.88$ & $3.24$ & $31.41$ & $\boldsymbol{58.48}$ \\
Task dep. & \underline{$7.43$} & $1.51$ & $2.33$ & $\boldsymbol{9.09}$ & $1.42$ & $3.17$ & $3.92$ \\
Terr.\ surv. & $\boldsymbol{12.86}$ & $6.73$ & $7.99$ & $10.67$ & $3.62$ & $8.43$ & \underline{$12.43$} \\
Terr.\ task & $5.20$ & $6.77$ & $6.29$ & $\boldsymbol{8.10}$ & $0.62$ & $4.80$ & \underline{$7.87$} \\
Surv.\ task & $10.45$ & $10.03$ & $12.72$ & \underline{$14.60$} & $0.21$ & $1.50$ & $\boldsymbol{24.30}$ \\
All three & $1.19$ & \underline{$1.77$} & $\boldsymbol{1.89}$ & $0.97$ & $0.42$ & $0.51$ & $1.48$ \\
\cmidrule(lr){1-8}
Average & $13.06$ & $15.23$ & $12.91$ & \underline{$20.60$} & $1.39$ & $9.93$ & $\boldsymbol{22.99}$ \\
\bottomrule
\end{tabular*}
\par
\endgroup
\end{table}

%% file: tables/tab_mars_terra.tex
\begin{table}[!htbp]
\centering
\begingroup
\input{macros_mars_names}
\caption{\textbf{\mars{} performance with \llmterra{}.}
\llmterra{}; five evaluation episodes per world.
Reward is mean $\pm$ sample SD; Score pools achievement success rates across episodes.
Average excludes Default; its Reward SD is computed across matched-seed world averages.
Bold/underlined: best/second-best. Metric definitions are in Appendix~\ref{app:metric_definitions}.}
\label{tab:mars_terra}
\fontsize{8pt}{9.5pt}\selectfont
\setlength{\tabcolsep}{2pt}
\medmuskip=1mu
\begin{tabular*}{\linewidth}{@{\extracolsep{\fill}}l*{7}{c}@{}}
\toprule
World & \react{} & \reactsearch{} & \reflexion{} & \ifr{} & \nesyc{} & \walle{} & \method{} \\
\midrule
\multicolumn{8}{@{}l}{\textit{Reward $\uparrow$}} \\
Default & $10.10\pm2.74$ & $10.70\pm2.97$ & $10.30\pm1.10$ & $\boldsymbol{13.70\pm2.88}$ & $3.50\pm1.95$ & $11.30\pm2.28$ & \underline{$12.50\pm2.19$} \\
Terrain & $11.30\pm4.21$ & $10.48\pm5.82$ & $11.10\pm4.85$ & $\boldsymbol{15.38\pm4.55}$ & $2.28\pm2.16$ & $10.12\pm4.09$ & \underline{$14.04\pm3.43$} \\
Survival & $8.10\pm5.39$ & $10.28\pm6.87$ & $7.10\pm3.39$ & \underline{$12.78\pm8.06$} & $0.70\pm0.55$ & $7.10\pm4.06$ & $\boldsymbol{15.50\pm1.52}$ \\
Task dep. & $6.70\pm3.51$ & $6.50\pm4.34$ & $6.10\pm3.94$ & \underline{$7.10\pm2.24$} & $0.10\pm0.00$ & $4.70\pm1.82$ & $\boldsymbol{8.10\pm3.94}$ \\
Terr.\ surv. & $7.50\pm2.70$ & \underline{$8.10\pm4.85$} & $\boldsymbol{9.90\pm1.64}$ & $5.30\pm1.30$ & $1.90\pm1.10$ & $4.10\pm2.35$ & $6.50\pm3.13$ \\
Terr.\ task & $\boldsymbol{7.86\pm1.87}$ & \underline{$7.68\pm3.75$} & $6.90\pm1.30$ & $5.10\pm1.87$ & $1.10\pm0.71$ & $5.50\pm2.41$ & $6.10\pm2.00$ \\
Surv.\ task & \underline{$7.50\pm5.68$} & $3.30\pm5.07$ & $6.30\pm3.70$ & $3.10\pm2.00$ & $2.70\pm0.55$ & $5.10\pm3.39$ & $\boldsymbol{7.90\pm6.69}$ \\
All three & $0.30\pm1.10$ & $0.70\pm2.07$ & $1.30\pm1.48$ & $\boldsymbol{3.10\pm2.24}$ & $-0.70\pm0.45$ & $0.70\pm2.30$ & \underline{$1.50\pm2.30$} \\
\cmidrule(lr){1-8}
Average & $7.04\pm0.86$ & $6.72\pm1.59$ & $6.96\pm0.86$ & \underline{$7.41\pm1.82$} & $1.15\pm0.51$ & $5.33\pm1.15$ & $\boldsymbol{8.52\pm1.54}$ \\
\midrule
\multicolumn{8}{@{}l}{\textit{Score (\%) $\uparrow$}} \\
Default & $21.42$ & $20.03$ & $18.81$ & $\boldsymbol{42.95}$ & $2.98$ & $23.71$ & \underline{$37.87$} \\
Terrain & $31.84$ & $33.57$ & $30.44$ & $\boldsymbol{57.79}$ & $1.52$ & $19.18$ & \underline{$38.16$} \\
Survival & $20.77$ & $37.49$ & $12.58$ & \underline{$50.36$} & $0.49$ & $12.92$ & $\boldsymbol{57.22}$ \\
Task dep. & $9.43$ & \underline{$12.13$} & $9.74$ & $8.20$ & $0.23$ & $4.92$ & $\boldsymbol{14.43}$ \\
Terr.\ surv. & $10.06$ & $\boldsymbol{14.47}$ & \underline{$14.23$} & $5.08$ & $1.13$ & $4.57$ & $8.66$ \\
Terr.\ task & \underline{$7.49$} & $\boldsymbol{9.97}$ & $6.77$ & $5.08$ & $0.77$ & $5.25$ & $6.38$ \\
Surv.\ task & \underline{$17.17$} & $6.30$ & $9.32$ & $2.43$ & $1.55$ & $6.52$ & $\boldsymbol{23.35}$ \\
All three & $0.64$ & $1.44$ & $1.56$ & $\boldsymbol{3.10}$ & $0.15$ & $1.19$ & \underline{$2.81$} \\
\cmidrule(lr){1-8}
Average & $13.91$ & $16.48$ & $12.09$ & \underline{$18.86$} & $0.84$ & $7.79$ & $\boldsymbol{21.57}$ \\
\bottomrule
\end{tabular*}
\par
\endgroup
\end{table}

%% file: tables/tab_mars_worlds.tex
\begin{table}[!htbp]
\centering
\begingroup
\input{macros_mars_names}
\caption{\textbf{\mars{} reward per evaluation seed (1/2).}
\llmluna{}; all seven methods and seeds 0--4 from Table~\ref{tab:mars}.
Each seed column is one recorded episode reward; the final column reports the mean and sample SD across the five seeds.}
\label{tab:mars_worlds}
\small
\setlength{\tabcolsep}{2pt}
\medmuskip=1mu
\begin{tabular*}{\linewidth}{@{\extracolsep{\fill}}ll*{5}{r}c@{}}
\toprule
World & Method & Seed 0 & Seed 1 & Seed 2 & Seed 3 & Seed 4 & Mean $\pm$ SD \\
\midrule
Default & \react{} & 7.10 & 17.10 & 7.10 & 13.10 & 6.10 & $10.10\pm4.80$ \\
 & \reactsearch{} & 10.10 & 11.10 & 7.10 & 8.10 & 10.10 & $9.30\pm1.64$ \\
 & \reflexion{} & 8.10 & 3.10 & 8.10 & 11.10 & 9.10 & $7.90\pm2.95$ \\
 & \ifr{} & 6.10 & 7.10 & 11.10 & 6.10 & 10.10 & $8.10\pm2.35$ \\
 & \nesyc{} & 1.10 & 0.10 & 1.10 & 1.10 & 0.10 & $0.70\pm0.55$ \\
 & \walle{} & 5.10 & 16.10 & 7.10 & 0.10 & 3.10 & $6.30\pm6.06$ \\
 & \method{} (ours) & 13.10 & 12.10 & 13.10 & 10.10 & 12.10 & $12.10\pm1.22$ \\
\midrule
Terrain & \react{} & 9.10 & 6.10 & 7.10 & 12.10 & 9.10 & $8.70\pm2.30$ \\
 & \reactsearch{} & 13.10 & 12.10 & 12.10 & 17.40 & 6.10 & $12.16\pm4.03$ \\
 & \reflexion{} & 11.10 & 3.10 & 2.10 & 13.10 & 14.00 & $8.68\pm5.66$ \\
 & \ifr{} & 10.10 & 3.10 & 9.10 & 14.10 & 10.10 & $9.30\pm3.96$ \\
 & \nesyc{} & -0.90 & -0.90 & -0.90 & -0.90 & -0.90 & $-0.90\pm0.00$ \\
 & \walle{} & 5.10 & 4.10 & 5.10 & 12.10 & 11.10 & $7.50\pm3.78$ \\
 & \method{} (ours) & 11.10 & 13.10 & 10.10 & 12.10 & 15.10 & $12.30\pm1.92$ \\
\midrule
Survival & \react{} & 2.10 & 4.10 & 8.10 & 9.10 & 12.10 & $7.10\pm4.00$ \\
 & \reactsearch{} & 4.10 & 11.10 & 8.10 & 12.10 & 12.10 & $9.50\pm3.44$ \\
 & \reflexion{} & 1.10 & 3.10 & 8.10 & 14.10 & 5.10 & $6.30\pm5.07$ \\
 & \ifr{} & 1.10 & 9.10 & 5.10 & 7.10 & 14.10 & $7.30\pm4.82$ \\
 & \nesyc{} & 1.10 & 0.10 & 1.10 & 2.50 & 0.10 & $0.98\pm0.99$ \\
 & \walle{} & 1.10 & 1.10 & 9.10 & 12.10 & 8.10 & $6.30\pm4.97$ \\
 & \method{} (ours) & 2.10 & 16.00 & 5.10 & 14.10 & 15.00 & $10.46\pm6.39$ \\
\midrule
Task dep. & \react{} & 3.10 & 4.10 & 5.10 & 3.10 & 5.10 & $4.10\pm1.00$ \\
 & \reactsearch{} & 1.10 & 4.10 & 4.10 & 3.10 & 2.10 & $2.90\pm1.30$ \\
 & \reflexion{} & 9.10 & 3.10 & 2.10 & 4.10 & 2.10 & $4.10\pm2.92$ \\
 & \ifr{} & 7.10 & 6.10 & 9.10 & 7.10 & 7.10 & $7.30\pm1.10$ \\
 & \nesyc{} & 1.10 & 1.10 & 1.10 & 0.10 & 1.10 & $0.90\pm0.45$ \\
 & \walle{} & 3.10 & 5.10 & 4.10 & 3.10 & 4.10 & $3.90\pm0.84$ \\
 & \method{} (ours) & 12.10 & 10.10 & 12.10 & 10.10 & 12.10 & $11.30\pm1.10$ \\
\bottomrule
\end{tabular*}
\par
\endgroup
\end{table}

%% file: tables/tab_mars_worlds_cont.tex
\begin{table}[!htbp]
\centering
\begingroup
\input{macros_mars_names}
\caption{\textbf{\mars{} reward per evaluation seed (2/2).}
\llmluna{}; all seven methods and seeds 0--4 from Table~\ref{tab:mars}.
Each seed column is one recorded episode reward; the final column reports the mean and sample SD across the five seeds.}
\label{tab:mars_worlds_cont}
\small
\setlength{\tabcolsep}{2pt}
\medmuskip=1mu
\begin{tabular*}{\linewidth}{@{\extracolsep{\fill}}ll*{5}{r}c@{}}
\toprule
World & Method & Seed 0 & Seed 1 & Seed 2 & Seed 3 & Seed 4 & Mean $\pm$ SD \\
\midrule
Terr.\ surv. & \react{} & 4.10 & 8.10 & 3.10 & 7.10 & 8.10 & $6.10\pm2.35$ \\
 & \reactsearch{} & 7.10 & 6.10 & 12.10 & 7.10 & 8.10 & $8.10\pm2.35$ \\
 & \reflexion{} & 3.10 & 7.10 & 6.10 & 1.10 & 8.10 & $5.10\pm2.92$ \\
 & \ifr{} & 3.10 & 2.10 & 6.10 & 6.10 & 13.10 & $6.10\pm4.30$ \\
 & \nesyc{} & 3.10 & 5.10 & 0.10 & 2.10 & 3.10 & $2.70\pm1.82$ \\
 & \walle{} & 7.10 & 0.10 & 3.10 & 2.10 & 7.10 & $3.90\pm3.11$ \\
 & \method{} (ours) & 10.10 & 2.10 & 6.10 & 5.10 & 8.10 & $6.30\pm3.03$ \\
\midrule
Terr.\ task & \react{} & 5.10 & 1.10 & 5.10 & 3.10 & 3.10 & $3.50\pm1.67$ \\
 & \reactsearch{} & 8.10 & 7.10 & 3.10 & 8.10 & 8.10 & $6.90\pm2.17$ \\
 & \reflexion{} & 5.10 & 5.10 & 6.10 & 5.10 & 11.00 & $6.48\pm2.56$ \\
 & \ifr{} & 4.10 & 2.10 & 4.10 & 2.10 & 7.10 & $3.90\pm2.05$ \\
 & \nesyc{} & 0.10 & 2.10 & 2.10 & 1.10 & 0.10 & $1.10\pm1.00$ \\
 & \walle{} & 4.10 & 3.10 & 0.10 & 3.10 & 4.10 & $2.90\pm1.64$ \\
 & \method{} (ours) & 10.10 & 4.10 & 5.10 & 10.10 & 9.10 & $7.70\pm2.88$ \\
\midrule
Surv.\ task & \react{} & 2.10 & 3.10 & 12.10 & 14.10 & 1.10 & $6.50\pm6.11$ \\
 & \reactsearch{} & -0.90 & 0.10 & 2.10 & 6.10 & 3.10 & $2.10\pm2.74$ \\
 & \reflexion{} & 2.10 & 1.10 & 3.10 & 7.10 & 5.10 & $3.70\pm2.41$ \\
 & \ifr{} & 2.10 & 2.10 & 10.10 & 9.10 & 9.10 & $6.50\pm4.04$ \\
 & \nesyc{} & 1.10 & 3.10 & 1.10 & 4.10 & 2.10 & $2.30\pm1.30$ \\
 & \walle{} & 2.10 & 0.10 & 4.10 & 10.10 & 12.10 & $5.70\pm5.18$ \\
 & \method{} (ours) & -0.90 & 0.10 & 5.10 & 11.10 & 12.10 & $5.50\pm6.02$ \\
\midrule
All three & \react{} & 2.10 & 5.10 & -0.90 & -0.90 & -0.90 & $0.90\pm2.68$ \\
 & \reactsearch{} & 3.10 & 1.10 & -0.90 & 5.10 & -0.90 & $1.50\pm2.61$ \\
 & \reflexion{} & 1.10 & 2.10 & -0.90 & 0.10 & 0.10 & $0.50\pm1.14$ \\
 & \ifr{} & -0.90 & -0.90 & 2.10 & 0.10 & -0.90 & $-0.10\pm1.30$ \\
 & \nesyc{} & -0.90 & -0.90 & -0.90 & -0.90 & -0.90 & $-0.90\pm0.00$ \\
 & \walle{} & 3.10 & 1.10 & -0.90 & 0.10 & 0.10 & $0.70\pm1.52$ \\
 & \method{} (ours) & 2.10 & 4.10 & -0.90 & -0.90 & 1.10 & $1.10\pm2.12$ \\
\bottomrule
\end{tabular*}
\par
\endgroup
\end{table}

%% file: sections/appendix_mars_knowledge_use.tex
\begingroup
\input{macros_mars_names}
{
\subsection{\mars{}: applying shared learned conditions}
\label{app:mars_knowledge_use}

\paragraph{Shared knowledge and evaluation scope.}
We analyze the existing \llmluna{} trajectories from all seven modified worlds, with five evaluation seeds per world and method (35 episodes each).
Both methods construct their knowledge before evaluation from the same 200 recorded transitions per world and keep their rule libraries fixed during these trajectories.
The analysis distinguishes whether a method possesses a resource condition from whether its current action respects that condition.

For each of the 14 paired knowledge-construction instances, we manually match explicit resource requirements for crafting and placement that both methods store correctly.
We require an explicit material name; a requirement without a quantity supports only a minimum of one unit.
This produces 22 pair-specific matches, including repeated conditions across instances; 13 of the 14 pairs have at least one match.
The metric checks only these shared resource requirements against the inventory at action selection.
Terrain, workstations, tools, and other feasibility conditions are outside this metric.

\paragraph{Candidates, commands, and denominators.}
For \method{}, the initial candidate is the first single-action proposal in the current turn, before verification feedback.
For \ifr{}, it is the next executable step selected from the plan after processing the previous action's outcome.
We do not test future plan steps against the current inventory or count plan steps discarded while processing a preceding failure.
Every candidate and command included in Table~\ref{tab:mars_knowledge_use} has an observed inventory.

The methods follow different trajectories, so their denominators count their own eligible candidates or issued commands rather than matched states.
For \method{}, the 245 initial candidates comprise 70 resource violations and 175 candidates satisfying the shared resource conditions.
All 70 violations are redirected to gathering or exploration before execution.
Another 39 candidates are redirected after terrain or workstation feedback, leaving 136 original crafting or placement commands to execute.
For \ifr{}, all 107 selected candidates are issued unchanged, including all 20 resource violations.
The redirection row conditions on violating candidates only: 70 for \method{} and 20 for \ifr{}.

\paragraph{Timing and outcome of revision.}
The 70 pre-execution redirections by \method{} occur in 19 episodes across six worlds.
The alternatives produce 54 controller successes and 16 failures, where controller success refers to the issued command.
Of the 20 violations issued by \ifr{}, 19 continue an existing plan without a new LLM call and one follows a fresh plan.
All 20 attempts fail and together consume 20 environment steps.
Each is followed immediately by an explanation and replanning turn that selects a different command outside the shared crafting and placement scope; these alternatives yield 16 controller successes and four failures.
Thus, both methods revise their behavior, but the observed revisions occur before a known-invalid attempt for \method{} and after the attempt for \ifr{}.

\paragraph{Coverage beyond shared conditions.}
The zero violations in Table~\ref{tab:mars_knowledge_use} apply to the shared learned conditions.
When the check is expanded to every actual resource requirement of \method{}'s 781 issued crafting or placement commands, all 363 violations occur among the 622 commands whose action lacks an explicit typed resource-cost rule; the 159 commands whose actions have such rules contain none.
}
\endgroup

%% file: sections/appendix_alfworld_results.tex
\begingroup
\input{macros_alfworld_names}
{
\subsection{\alfworld{} full results}
\label{app:alfworld_models}
\label{app:alfworld_types}

{Table~\ref{tab:alfworld_observations} reports the three evaluated methods---\nesyc{}, \walle{}, and \method{}---under standard and disclosed initial observations, with the input and output tokens behind the \llmluna{} condition in Table~\ref{tab:alfworld_summary}.}

\paragraph{Initial-state disclosure.}
Initial object locations specify which receptacle contains each object at episode reset, including contents that the agent has not yet observed through exploration.
Inventory specifies the objects held by the agent, or explicitly states that its hands are empty.
For this control, the initial object locations and inventory are appended as ordinary observation sentences to the first textual observation.
The same disclosure procedure is applied to all three methods.
Subsequent observations use the standard interface; no additional target annotations or action-effect rules are supplied.
The model, learned knowledge, 134 unseen tasks, seed, and 49-step budget remain the same across the two observation conditions.

Table~\ref{tab:alfworld_observations} reports both conditions for every method.
With disclosure, \method{} completes 134/134 tasks rather than 132/134, and \walle{} completes 133/134 rather than 129/134; both also use fewer steps and evaluation tokens.
Our \nesyc{} reconstruction completes 0/134 tasks under standard observations: initially unobserved target objects prevent planning, and all 6,566 executed steps are idle actions.
With disclosure it completes 35/134 tasks.
The $\dagger$ rows in Tables~\ref{tab:alfworld_summary} and~\ref{tab:alfworld_types} use this disclosed-state run for \nesyc{}, while the \method{} and \walle{} rows use standard observations.

\paragraph{Evaluation call roles.}
{Across the 134 standard-observation tasks, all 1,833 evaluation calls by \method{} propose actions; learned conditions are checked symbolically.
The 3,862 calls by \walle{} comprise 2,036 action proposals and 1,826 outcome predictions.
These counts reconcile with Table~\ref{tab:alfworld_observations}: the methods execute 1,805 and 1,985 environment steps, giving 1.02 and 1.95 LLM calls per step, respectively.
The prediction calls account for 90.0\% of the 2,029-call difference and consume 1,919,113 tokens, or 92.2\% of the observed evaluation-token difference.
Knowledge-construction costs are reported separately (Table~\ref{tab:alfworld_learning_usage}).
\par}
}
\input{tables/tab_alfworld_observations}

{
Table~\ref{tab:alfworld_types} reports success by task family; following \citet{shridhar2021alfworld}, the families are Pick \& Place, Clean \& Place, Heat \& Place, Cool \& Place, Examine in Light, and Pick Two \& Place.
}
\input{tables/tab_alfworld_types}
\endgroup

%% file: tables/tab_alfworld_observations.tex
\begin{table}[!htbp]
\centering
\begingroup
\input{macros_alfworld_names}
\caption{\textbf{\alfworld{} with and without initial-state disclosure (\llmluna{}).}
Each method uses the same learned knowledge and 134 unseen tasks in both conditions, with one attempt and a 49-step budget per task.
Success gives completed tasks; Steps, Calls, and input (In) and output (Out) tokens in thousands are means over all 134 episodes, including failures.
The disclosure condition supplies initial object locations and inventory to every method.}
\label{tab:alfworld_observations}
\small
\setlength{\tabcolsep}{2pt}
\medmuskip=1mu
\begin{tabular*}{\linewidth}{@{\extracolsep{\fill}}l*{10}{r}@{}}
\toprule
& \multicolumn{5}{c}{Standard observations} & \multicolumn{5}{c}{Initial-state disclosure} \\
\cmidrule(lr){2-6}\cmidrule(lr){7-11}
Method & Success & Steps & Calls & In (k) & Out (k) & Success & Steps & Calls & In (k) & Out (k) \\
\midrule
\nesyc{} & 0/134 & 49.0 & 1.0 & 5.82 & 0.06 & 35/134 & 38.6 & 9.3 & 269.33 & 11.99 \\
\walle{} & 129/134 & 14.8 & 28.8 & 47.08 & 0.61 & 133/134 & 7.8 & 15.6 & 27.83 & 0.38 \\
\method{} (ours) & 132/134 & 13.5 & 13.7 & 32.00 & 0.15 & 134/134 & 6.7 & 8.3 & 23.28 & 0.15 \\
\bottomrule
\end{tabular*}
\par
\endgroup
\end{table}

%% file: tables/tab_alfworld_types.tex
\begin{table}[!htbp]
\centering
\begingroup
\input{macros_alfworld_names}
\caption{\textbf{\alfworld{} success (\%) by task family under the \llmluna{} condition.}
The six families follow \citet{shridhar2021alfworld}; parentheses give the number of unseen tasks.
All is success over all 134 tasks, weighted by family size.
Best values are bold; second-best All values are underlined. Ties share the same formatting.
$^\dagger$Initial object locations and inventory are additionally provided to \nesyc{}.}
\label{tab:alfworld_types}
\setlength{\tabcolsep}{3pt}
\begin{tabular}{@{}l*{7}{r}@{}}
\toprule
Method & \shortstack{Pick\\(24)} & \shortstack{Clean\\(31)} & \shortstack{Heat\\(23)} & \shortstack{Cool\\(21)} & \shortstack{Examine\\(18)} & \shortstack{Pick two\\(17)} & All \\
\midrule
\nesyc{}$^\dagger$ & 75.0 & 0.0 & 34.8 & 0.0 & 0.0 & 52.9 & 26.1 \\
\walle{} & \textbf{100.0} & \textbf{96.8} & 95.7 & \textbf{100.0} & 88.9 & 94.1 & \underline{96.3} \\
\method{} (ours) & \textbf{100.0} & \textbf{96.8} & \textbf{100.0} & \textbf{100.0} & \textbf{94.4} & \textbf{100.0} & \textbf{98.5} \\
\bottomrule
\end{tabular}
\par
\endgroup
\end{table}

%% file: tables/tab_tacit_usage.tex
\begin{table}[!htbp]
\centering
\begingroup
\input{macros_tacit_names}
\caption{\textbf{\benchmark{} LLM usage (\llmluna{}).}
Construction calls and tokens are per plant (15 plants per scale); evaluation calls and tokens are per episode (28, 36, and 40 episodes).
Tokens sum input and output in thousands (k).
\react{} and \reactsearch{} construct no knowledge; \nesyc{}, which executes no repair action (Appendix~\ref{app:full_results}), is omitted.}
\label{tab:tacit_usage}
\setlength{\tabcolsep}{3pt}
\begin{tabular}{@{}lrr@{\hspace{12pt}}rr@{}}
\toprule
 & \multicolumn{2}{c}{Construction (per plant)} & \multicolumn{2}{c}{Evaluation (per episode)} \\
\cmidrule(lr){2-3}\cmidrule(lr){4-5}
Method & Calls & Tokens (k) & Calls & Tokens (k) \\
\midrule
\multicolumn{5}{@{}l}{\textit{Small}} \\
\react{} & 0 & 0 & 22.2 & 154.6 \\
\reactsearch{} & 0 & 0 & 22.9 & 166.6 \\
\reflexion{} & 6.8 & 2.3 & 23.2 & 170.4 \\
\ifr{} & 6.8 & 3.7 & 13.8 & 135.3 \\
\walle{} & 13.4 & 11.7 & 88.0 & 436.6 \\
\method{} (ours) & 13.1 & 14.8 & 16.2 & 128.0 \\
\midrule
\multicolumn{5}{@{}l}{\textit{Medium}} \\
\react{} & 0 & 0 & 41.8 & 619.0 \\
\reactsearch{} & 0 & 0 & 38.7 & 564.1 \\
\reflexion{} & 32.0 & 10.6 & 46.4 & 734.3 \\
\ifr{} & 32.0 & 17.1 & 21.5 & 422.0 \\
\walle{} & 39.1 & 34.9 & 152.9 & 1574.0 \\
\method{} (ours) & 48.8 & 58.7 & 27.1 & 265.5 \\
\midrule
\multicolumn{5}{@{}l}{\textit{Large}} \\
\react{} & 0 & 0 & 48.8 & 1204.3 \\
\reactsearch{} & 0 & 0 & 56.4 & 1403.6 \\
\reflexion{} & 63.1 & 20.7 & 63.1 & 1606.2 \\
\ifr{} & 63.1 & 33.8 & 32.2 & 1038.9 \\
\walle{} & 71.5 & 62.8 & 170.6 & 2541.4 \\
\method{} (ours) & 95.8 & 119.7 & 24.6 & 309.4 \\
\bottomrule
\end{tabular}
\par
\endgroup
\end{table}

%% file: sections/appendix_mars_usage.tex
\begingroup
\input{macros_mars_names}
For \mars{}, Table~\ref{tab:mars_usage} reports evaluation costs for all model conditions and Table~\ref{tab:mars_learning_usage} reports construction costs.
Each method--model aggregate in these tables covers 40 evaluation episodes across eight worlds.
Figure~\ref{fig:mars_efficiency} uses the 35 \llmluna{} episodes from the seven modified worlds, matching the main performance average.
For each world, we divide the total evaluation input and output tokens by the total executed environment steps across its five episodes.
We then average these seven rates with equal world weights; the figure's Score is the corresponding mean of the seven pooled world Scores.
Environment steps count primitive controller interactions, not LLM calls or high-level command proposals.
This rate normalizes for interaction length, and the figure pairs it with achievement Score.
Knowledge-construction costs are excluded from this rate and reported separately.
\par
{For the same seven-world \llmluna{} population, pooled input tokens per call are 2.36k for \method{} and 4.26k for \reactsearch{}.
Their equally weighted world-average call rates are similar: 0.3655 and 0.3639 per environment step.
Prompt inspection shows that \method{} supplies compact learned conditions, whereas \reactsearch{} retains five retrieved transition documents; both retain ten turns of history.
Across all 35 paired initial observations, which are identical between methods, first-call input averages 1,053 versus 2,982 tokens.
The usage difference therefore lies primarily in input context.
\par}
\endgroup

%% file: tables/tab_mars_usage.tex
\begin{table}[!htbp]
\centering
\begingroup
\input{macros_mars_names}
\caption{\textbf{\mars{} evaluation resource use.}
Means over 40 episodes per method and model (eight worlds, five episodes each).
Steps are environment steps; Calls, In, Out, and Total tokens are per episode.
Input/call is mean input tokens per call; Total tokens sums input and output before rounding.
Token quantities are in thousands (k).
Knowledge-construction costs are reported separately in Appendix~\ref{app:usage}.}
\label{tab:mars_usage}
\setlength{\tabcolsep}{3pt}
\begin{tabular}{@{}lrrrrrr@{}}
\toprule
Method & Steps & Calls & In (k) & Out (k) & \shortstack{Input/call\\(k)} & \shortstack{Total tokens\\(k)} \\
\midrule
\multicolumn{7}{@{}l}{\textit{\llmluna{}}} \\
\react{} & 198.6 & 60.8 & 109.13 & 1.42 & 1.79 & 110.55 \\
\reactsearch{} & 252.5 & 83.0 & 354.22 & 2.12 & 4.27 & 356.35 \\
\reflexion{} & 249.5 & 75.8 & 174.74 & 2.05 & 2.31 & 176.79 \\
\ifr{} & 218.2 & 55.2 & 298.10 & 12.45 & 5.40 & 310.55 \\
\nesyc{} & 177.9 & 183.1 & 2237.98 & 13.65 & 12.22 & 2251.63 \\
\walle{} & 220.0 & 204.1 & 544.71 & 5.26 & 2.67 & 549.97 \\
\method{} (ours) & 304.8 & 103.8 & 244.95 & 1.01 & 2.36 & 245.95 \\
\midrule
\multicolumn{7}{@{}l}{\textit{\llmlunareasoning{}}} \\
\react{} & 220.4 & 69.0 & 127.57 & 7.70 & 1.85 & 135.27 \\
\reactsearch{} & 267.2 & 91.2 & 395.61 & 11.01 & 4.34 & 406.62 \\
\reflexion{} & 258.0 & 83.0 & 198.72 & 9.60 & 2.39 & 208.33 \\
\ifr{} & 291.7 & 60.5 & 320.37 & 32.28 & 5.30 & 352.65 \\
\nesyc{} & 208.6 & 192.3 & 2182.93 & 114.77 & 11.35 & 2297.70 \\
\walle{} & 228.6 & 148.9 & 379.76 & 15.21 & 2.55 & 394.97 \\
\method{} (ours) & 306.6 & 84.5 & 204.36 & 11.29 & 2.42 & 215.64 \\
\midrule
\multicolumn{7}{@{}l}{\textit{\llmterra{}}} \\
\react{} & 252.1 & 69.9 & 126.91 & 2.13 & 1.82 & 129.04 \\
\reactsearch{} & 262.6 & 70.1 & 300.29 & 2.15 & 4.28 & 302.45 \\
\reflexion{} & 211.6 & 62.8 & 148.76 & 1.98 & 2.37 & 150.73 \\
\ifr{} & 256.0 & 42.8 & 207.06 & 7.34 & 4.84 & 214.39 \\
\nesyc{} & 188.1 & 183.6 & 2067.74 & 13.62 & 11.27 & 2081.36 \\
\walle{} & 210.3 & 146.2 & 319.72 & 2.81 & 2.19 & 322.53 \\
\method{} (ours) & 334.1 & 91.9 & 219.66 & 0.91 & 2.39 & 220.57 \\
\bottomrule
\end{tabular}
\par
\endgroup
\end{table}

%% file: tables/tab_mars_learning_usage.tex
\begin{table}[!htbp]
\centering
\begingroup
\input{macros_mars_names}
\caption{\textbf{\mars{} knowledge-construction resource use.}
Calls, In, Out, and Total tokens are per construction run.
\nesyc{} constructs its rules once per world (eight runs); its rule refinements during play are counted as evaluation (Table~\ref{tab:mars_usage}).
\react{} and \reactsearch{} require no LLM-based knowledge construction; their input length is undefined.
Input/call is mean input tokens per call; Total tokens sums input and output before rounding.
Token quantities are in thousands (k).}
\label{tab:mars_learning_usage}
\setlength{\tabcolsep}{3pt}
\begin{tabular}{@{}lrrrrr@{}}
\toprule
Method & Calls & In (k) & Out (k) & \shortstack{Input/call\\(k)} & \shortstack{Total tokens\\(k)} \\
\midrule
\multicolumn{6}{@{}l}{\textit{\llmluna{}}} \\
\react{} & 0 & 0 & 0 & \textemdash & 0 \\
\reactsearch{} & 0 & 0 & 0 & \textemdash & 0 \\
\reflexion{} & 20.0 & 12.76 & 2.61 & 0.64 & 15.38 \\
\ifr{} & 20.0 & 18.28 & 3.88 & 0.91 & 22.16 \\
\nesyc{} & 63.5 & 3768.39 & 33.08 & 59.34 & 3801.47 \\
\walle{} & 255.3 & 1069.25 & 24.59 & 4.19 & 1093.84 \\
\method{} (ours) & 1.0 & 10.91 & 1.34 & 10.91 & 12.25 \\
\midrule
\multicolumn{6}{@{}l}{\textit{\llmlunareasoning{}}} \\
\react{} & 0 & 0 & 0 & \textemdash & 0 \\
\reactsearch{} & 0 & 0 & 0 & \textemdash & 0 \\
\reflexion{} & 20.0 & 12.76 & 5.01 & 0.64 & 17.77 \\
\ifr{} & 20.0 & 18.28 & 7.04 & 0.91 & 25.32 \\
\nesyc{} & 63.5 & 3739.88 & 79.09 & 58.90 & 3818.98 \\
\walle{} & 228.1 & 1001.05 & 51.70 & 4.39 & 1052.75 \\
\method{} (ours) & 1.0 & 11.00 & 3.19 & 11.00 & 14.19 \\
\midrule
\multicolumn{6}{@{}l}{\textit{\llmterra{}}} \\
\react{} & 0 & 0 & 0 & \textemdash & 0 \\
\reactsearch{} & 0 & 0 & 0 & \textemdash & 0 \\
\reflexion{} & 20.0 & 12.76 & 3.25 & 0.64 & 16.02 \\
\ifr{} & 20.0 & 18.28 & 3.14 & 0.91 & 21.42 \\
\nesyc{} & 63.5 & 3729.85 & 12.96 & 58.74 & 3742.80 \\
\walle{} & 225.6 & 996.33 & 11.87 & 4.42 & 1008.20 \\
\method{} (ours) & 1.0 & 11.20 & 1.12 & 11.20 & 12.32 \\
\bottomrule
\end{tabular}
\par
\endgroup
\end{table}

%% file: tables/tab_alfworld_learning_usage.tex
\begin{table}[!htbp]
\centering
\begingroup
\input{macros_alfworld_names}
{
\caption{\textbf{\alfworld{} knowledge-construction LLM usage (\llmluna{}).}
Totals for one construction run per method on the shared 238-transition corpus.
Each resulting knowledge base is reused across 134 evaluation tasks.
Token quantities use thousands of tokens (k); Total tokens sums input and output before rounding.
Costs exclude corpus collection and evaluation.}
\label{tab:alfworld_learning_usage}
\setlength{\tabcolsep}{3pt}
\begin{tabular}{@{}lrrrr@{}}
\toprule
Method & Calls & In (k) & Out (k) & Total tokens (k) \\
\midrule
\nesyc{} & 79 & 1989.08 & 8.62 & 1997.70 \\
\walle{} & 254 & 349.37 & 14.84 & 364.21 \\
\method{} (ours) & 15 & 116.43 & 0.94 & 117.36 \\
\bottomrule
\end{tabular}
\par
}
\endgroup
\end{table}

%% file: sections/appendix_tacit_diagnostics.tex
\begingroup
\input{macros_tacit_names}
{
\subsection{\benchmark{} candidate and verification diagnostics}
\label{app:tacit_diagnostics}

These post hoc analyses use the 104 \llmluna{} episodes in Table~\ref{tab:main}: 28 small, 36 medium, and 40 large episodes from 15 plant worlds per scale.
Anonymized identifiers are decoded only in this analysis, to compare a proposed target with the benchmark's hidden fault target.
Target alignment measures whether a candidate names the fault.
\nesyc{} is excluded from these diagnostics because its inspected trajectories contain no repair executions.

\paragraph{Replacement candidates and coverage.}
For \method{} and \walle{}, we extract replacement proposals before their native action-verification procedures, including rejected proposals.
For \ifr{}, we parse replacement steps in each newly generated plan, including steps that are not subsequently executed.
Quoted dialogue history, inspection targets, and explanation-only text are excluded.
Within each episode, we retain distinct replacement targets in order of first proposal.
If $C_e^{(k)}$ is the first $k$ distinct targets and $c_e^\ast$ the true fault, coverage is
\begin{equation}
\operatorname{Coverage}@k
=\frac{1}{N}\sum_{e=1}^{N}\mathbb{1}\{c_e^\ast\in C_e^{(k)}\}.
\label{eq:coverage}
\end{equation}
Episodes with fewer than $k$ candidates retain their complete observed candidate set and remain in the denominator.
Table~\ref{tab:tacit_candidate_summary} reports $k=3$ and full-sequence counts on medium plants; Table~\ref{tab:tacit_candidate_diagnostics} additionally reports $k=1$ and $k=5$ across all scales.
Candidate precision divides the number of target-aligned episode--target pairs by the number of distinct proposed episode--target pairs.
We use 20,000 paired world-cluster bootstrap draws within each scale and report percentile 95\% intervals.
The coverage-at-three difference between \method{} and \ifr{} has intervals of [26.7, 66.7], [33.3, 81.0], and [62.2, 90.2] percentage points across the three scales.
}
\input{tables/tab_tacit_candidate_diagnostics}

{
\setlength{\intextsep}{4pt}
\input{tables/tab_tacit_hypothesis_diagnostics}
{
\paragraph{Native hypothesis selection.}
A registered \method{} hypothesis is identified by its cause, pattern reference, and ordered atoms.
The registered totals are 149, 287, and 206 across small, medium, and large, corresponding to 5.32, 7.97, and 5.15 per episode.
Different conditions can name the same cause: the medium cohort contains 155 distinct episode--cause pairs, or 4.31 per episode.
For the selection analysis, we follow the 140, 272, and 192 hypotheses that are visible at the start of a later turn.
The remaining 9, 15, and 14 hypotheses are admitted only on the terminal turn and are excluded from both the initial and retained sets.
Retained hypotheses are open or confirmed at the start of the final environment turn.
Table~\ref{tab:tacit_hypothesis_diagnostics} gives target precision, target-aligned retention, and off-target removal on this matched population.
Figure~\ref{fig:tacit_verification} follows this same population at three intermediate checkpoints aligned by each hypothesis's admission.
For a hypothesis admitted at turn $a$ in an episode of $T$ turns, checkpoint $+k$ reads its status at the start of turn $\min(a+k,T)$, for $k\in\{1,2,3\}$; Final reads turn $T$.
An episode that has already ended retains its last recorded status.
{The Init. baseline includes every hypothesis in the matched population before selection.}
At each later checkpoint, target precision is the fraction naming the true fault among open or confirmed hypotheses.
Selection includes updates after unsuccessful interventions.

\par
\WFclear
}
}

{
\setlength{\intextsep}{4pt}
\input{tables/tab_tacit_executed_diagnostics}
{
\paragraph{Explicit refutation and execution.}
We separately link repair proposals, verifier feedback, and the action executed in the same turn.
For explicit refutation, we count one choice per episode, environment turn, repair verb, and target, over both replacement and refill proposals.
In the medium \method{} cohort, 139 choices receive explicit refutation feedback; 137 target a noncausal unit, and all 137 are withheld or revised in that turn.
Deduplicating across turns leaves 31 off-target choices among 33 distinct episode--verb--target choices.
Explicit refutation is a subset of all negative gate decisions.
Across all scales, 233 of 241 explicitly refuted choices are off-target and withheld in the corresponding turn.
Refutations can follow an unsuccessful earlier repair, contradicted observations, or an explained anomaly disappearing.

For broader gate diagnostics, we use the first decision on each distinct replacement target within a turn.
For \method{}, negative decisions include explicit repair, practice, or repetition feedback.
For \walle{}, the decision is the final world-model Boolean after any code-conflict feedback, before execution at the proposal limit.
\ifr{} provides post-action explanations and replanning, without a corresponding pre-action verifier veto; a verifier rejection rate is therefore undefined for this method.
Table~\ref{tab:tacit_executed_diagnostics} counts the replacement commands actually executed, including repeated targets and actions executed at the proposal limit.
\par
\WFclear
}
}

{
Repeated off-target repairs provide a separate outcome diagnostic.
We count an executed repair command to a noncausal target as repeated if an earlier repair command in the same episode targeted that unit, including a change of repair verb.
Across the 104 matched episodes, \method{}, \walle{}, and \ifr{} issue 13, 633, and 280 such repeats, respectively.

\paragraph{Illustrative traces.}
The following cases illustrate distinct mechanisms.
In a medium chain task, \method{} proposes replacing the faulty input storage under an induced diagnostic pattern but omits its required inflow atom.
The claim is not admitted, and the first executed action opens the storage.
The next proposal adds the missing atom and replaces the target, completing the episode in two turns with no spurious intervention.
On the same task, \walle{} initially predicts failure for an off-target locomotive replacement at turn 18.
Executable-rule feedback changes the prediction to success, and the replacement executes; the episode finishes unresolved after 60 turns with 11 spurious interventions.
The contrast illustrates how action-feasibility feedback can differ from evidence that a target explains the fault.

In a medium elimination task, \method{}'s five tracked hypotheses narrow to one naming the true target, \texttt{st-04}.
Three hypotheses on \texttt{st-00} or \texttt{st-01} are refuted after earlier replacements fail to end the episode, and the anomaly explained by the \texttt{st-02} hypothesis clears.
The \texttt{st-04} hypothesis remains confirmed, and the episode succeeds in eight turns with two spurious interventions.
}
\endgroup

%% file: tables/tab_tacit_candidate_diagnostics.tex
\begin{table}[!htbp]
\centering
\begingroup
\input{macros_tacit_names}
\caption{\textbf{Replacement-candidate diagnostics.} Targets is the mean number of distinct proposed replacement targets per episode. C@$k$ and C@all are true-fault coverage (\%) within the first $k$ targets and the complete sequence. Precision is correct-target/total episode--target pairs. Small, medium, and large contain 28, 36, and 40 episodes.}
\label{tab:tacit_candidate_diagnostics}
\small
\setlength{\tabcolsep}{3pt}
\begin{tabular}{@{}llrrrrrr@{}}
\toprule
Scale & Method & Targets & C@1 & C@3 & C@5 & C@all & Precision \\
\midrule
Small & \method{} (ours) & 2.68 & 50.0 & 75.0 & 82.1 & 82.1 & 23/75 \\
 & \walle{} & 4.79 & 3.6 & 35.7 & 50.0 & 50.0 & 14/134 \\
 & \ifr{} & 5.79 & 10.7 & 28.6 & 50.0 & 64.3 & 18/162 \\
\midrule
Medium & \method{} (ours) & 3.36 & 33.3 & 69.4 & 75.0 & 88.9 & 32/121 \\
 & \walle{} & 5.14 & 0.0 & 27.8 & 36.1 & 47.2 & 17/185 \\
 & \ifr{} & 9.00 & 0.0 & 11.1 & 38.9 & 66.7 & 24/324 \\
\midrule
Large & \method{} (ours) & 2.78 & 57.5 & 87.5 & 92.5 & 97.5 & 39/111 \\
 & \walle{} & 5.70 & 5.0 & 22.5 & 37.5 & 52.5 & 21/228 \\
 & \ifr{} & 14.38 & 0.0 & 10.0 & 20.0 & 65.0 & 26/575 \\
\bottomrule
\end{tabular}
\par
\endgroup
\end{table}

%% file: tables/tab_tacit_hypothesis_diagnostics.tex
\begin{table}[!htbp]
\centering
\begingroup
\input{macros_tacit_names}
\caption{\textbf{Selection within the hypothesis ledger.} Target-aligned/all tracked at baseline, then target-aligned/retained at later checkpoints (retained: open or confirmed). The last two rows give final target-aligned retention and off-target refutation; terminal-only hypotheses (9/15/14 by scale) are excluded.}
\label{tab:tacit_hypothesis_diagnostics}
\small
\setlength{\tabcolsep}{3pt}
\begin{tabular}{@{}lrrr@{}}
\toprule
Checkpoint / measure & Small & Medium & Large \\
\midrule
All tracked & 20/140 & 28/272 & 34/192 \\
Checkpoint +1 & 19/68 & 23/112 & 31/98 \\
Checkpoint +2 & 19/48 & 21/61 & 28/68 \\
Checkpoint +3 & 19/32 & 20/36 & 28/42 \\
Final & 18/21 & 20/21 & 25/32 \\
\midrule
True-target retention & 18/20 & 20/28 & 25/34 \\
Off-target removal & 117/120 & 243/244 & 151/158 \\
\bottomrule
\end{tabular}
\par
\endgroup
\end{table}

%% file: tables/tab_tacit_executed_diagnostics.tex
\begin{table}[!htbp]
\centering
\begingroup
\input{macros_tacit_names}
\caption{\textbf{Precision of executed replacement commands.} Each cell gives the percentage and correct-target/total commands. Repeated targets are counted each time. These final executions are distinct from proposed candidates and native gate passes.}
\label{tab:tacit_executed_diagnostics}
\small
\setlength{\tabcolsep}{3pt}
\begin{tabular}{@{}lrrr@{}}
\toprule
Method & Small & Medium & Large \\
\midrule
\method{} (ours) & 29.3\% (22/75) & 25.4\% (29/114) & 36.4\% (36/99) \\
\walle{} & 2.3\% (3/133) & 5.2\% (12/232) & 3.8\% (14/364) \\
\ifr{} & 9.9\% (18/182) & 5.6\% (24/426) & 3.7\% (26/694) \\
\react{} & 8.9\% (11/124) & 4.7\% (15/321) & 3.5\% (17/482) \\
\reactsearch{} & 5.6\% (11/197) & 4.3\% (17/392) & 2.2\% (15/688) \\
\reflexion{} & 8.6\% (12/140) & 3.1\% (12/387) & 1.3\% (8/600) \\
\bottomrule
\end{tabular}
\par
\endgroup
\end{table}

%% file: sections/appendix_reproducibility.tex
{
\section{Experimental Setup and Reproducibility}
\label{app:reproducibility}
\label{app:experimental_setup}

This appendix specifies the reference corpora, evaluation schedules, model configurations, and aggregation rules used in Section~\ref{sec:setup}.
Knowledge construction and prompts are described in Appendices~\ref{app:admission}, \ref{app:prompts}, and~\ref{app:baselines}.

\subsection{Reference corpora and evaluation tasks}
\label{app:corpus_protocol}

{Each evaluation episode starts with fresh interaction history and transient state.}

}
\input{sections/appendix_tacit_setup}
{

{For \walle{} in Table~\ref{tab:main}, each scale uses one corpus-learned knowledge base for each of its 15 worlds.
All knowledge bases remain fixed during evaluation.
Each episode contributes one outcome under the procedure in Appendix~\ref{app:implementation_adjustments}.}

}
\input{sections/appendix_mars_setup}
\input{sections/appendix_alfworld_setup}
{
\subsection{Model and inference settings}
\label{app:model_settings}

The \llmluna{} condition uses the \texttt{gpt-5.6-luna} model with reasoning effort set to \texttt{none}.
The \llmlunareasoning{} condition uses the same model with reasoning effort \texttt{medium}; the \llmterra{} condition uses \texttt{gpt-5.6-terra} with effort \texttt{none}.
Each condition uses its designated model for both knowledge construction and evaluation.
We retain the provider's default temperature settings and use the same model configuration across methods within a comparison.
The \mars{} and \alfworld{} runs cap each completion at 8,192 tokens.
Retrieval uses the OpenAI \texttt{text-embedding-3-small} embedding model.
The environment-specific proposal budgets of \method{} are listed in Table~\ref{tab:instantiation}.

\subsection{Task metrics and aggregation}
\label{app:metric_definitions}

}
\input{sections/appendix_tacit_metrics}
\input{sections/appendix_mars_metrics}
\input{sections/appendix_alfworld_metrics}

%% file: sections/appendix_tacit_setup.tex
\paragraph{\benchmark{}.}
{We use the fixed reference corpora and evaluation population in Appendix~\ref{app:benchmark_population}, with the scale-specific episode counts and turn budgets in Table~\ref{tab:profiles}.
Corpus access follows the \method{} construction procedure in Appendix~\ref{app:admission} and the baseline procedures in Appendix~\ref{app:baselines}.}

%% file: sections/appendix_mars_setup.tex
\begingroup
\input{macros_mars_names}
\paragraph{\mars{}.}
Following the evaluation configuration of \citet{zhou2025walle}, we use Default and seven worlds formed by changing terrain, survival mechanics, task dependencies, or their combinations.
Methods share the same reference corpus of 200 recorded transitions per world.
Each record contains the observed state, action, outcome, feedback, and subsequent state.
For document-based processing, consecutive groups of ten records form 20 documents per world.

The agent receives a textual local observation, including a $9\times9$ view, nearby objects, inventory, and health, food, drink, and energy levels.
The high-level commands are \texttt{mine}, \texttt{attack}, \texttt{explore}, \texttt{make}, \texttt{place}, and \texttt{sleep}.
Each episode is capped at 1,000 environment steps and 1,000 agent turns, with ten turns of interaction history.
A high-level command can consume multiple environment steps.
Each method is evaluated on five seeds per world, giving 40 episodes across eight worlds for each model condition.
\par
\endgroup

%% file: sections/appendix_alfworld_setup.tex
\begingroup
\input{macros_alfworld_names}
\paragraph{\alfworld{}.}
{We use the text interface and all 134 unseen evaluation tasks, with the six task families and their sizes listed in Table~\ref{tab:alfworld_types}.}
The reference corpus contains 238 transitions from six training games, one per task family.
In each training game, 20 exploratory probes precede an expert-policy trajectory; the resulting corpus contains both successful and unsuccessful interactions.
{The same fixed corpus is used to construct each method's knowledge.}
Each unseen task is evaluated once with a budget of 49 environment steps.
\method{} and \walle{} use standard textual observations; the reported \nesyc{} result additionally provides initial object locations and inventory (Appendix~\ref{app:alfworld_models}).
Environment steps include submitted \texttt{think} and idle commands; internal LLM calls are counted separately.
We report the \llmluna{} condition for this benchmark.
\endgroup

%% file: sections/appendix_tacit_metrics.tex
\paragraph{\benchmark{}.}
{We aggregate the metrics defined in Appendix~\ref{app:scoring} over all scheduled episodes at each plant scale and, where specified, by task type.
}

%% file: sections/appendix_mars_metrics.tex
\begingroup
\input{macros_mars_names}
\paragraph{\mars{}.}
Episode reward sums the environment rewards: one unit for the first completion of each achievement, together with a health-change contribution of $0.1$ per health unit.
We report the mean and sample standard deviation of episode reward over five evaluation episodes in each world, using denominator $n-1$ for the sample variance.
For achievement $i$ in evaluation episode $r$, let $z_{ir}=1$ if the achievement is attained at least once and $0$ otherwise.
Following the achievement-success-rate metric of \citet{tang2024mars}, let $p_i = (100/n)\sum_{r=1}^{n} z_{ir}$ be the percentage of episodes attaining achievement $i$, with $n=5$.
The achievement score is
\begin{equation}
  \mathrm{Score}
  = \exp\!\left(\frac{1}{22}\sum_{i=1}^{22}\log(1+p_i)\right)-1.
  \label{eq:mars_score}
\end{equation}
Success rates are computed across the five episodes within each world, model, and method before applying the geometric aggregation.
This yields one Score per cell.
Following the average-row convention of \citet{zhou2025walle}, the cross-world summary excludes the Default world.
The Score Average is the arithmetic mean of the seven modified-world Scores.
For Reward, we first average the seven modified worlds within each matched environment seed; the Average row reports the mean and sample standard deviation across the five resulting seed averages.
\par
\endgroup

%% file: sections/appendix_alfworld_metrics.tex
\begingroup
\input{macros_alfworld_names}
\paragraph{\alfworld{}.}
Task success is the fraction of the 134 unseen tasks for which the environment reports completion within the step budget.
Mean environment steps are the total executed steps divided by the number of evaluation tasks, including unsuccessful episodes.
Environment steps count executed commands; internal planning and verification calls contribute to LLM usage instead.
\endgroup